\documentclass{article}
\PassOptionsToPackage{numbers,compress}{natbib}
\usepackage[main,preprint]{neurips_2026}

\pdftrailerid{}
\usepackage{amsmath, amssymb, mathtools, amsthm}
\usepackage{graphicx, subcaption}
\usepackage{booktabs}
\usepackage[ruled,vlined,linesnumbered]{algorithm2e}
\usepackage[breaklinks=true]{hyperref}
\usepackage[capitalize,noabbrev]{cleveref}
\usepackage{multirow}
\usepackage[table]{xcolor}
\usepackage{microtype}
\usepackage{tabularx}
\usepackage{enumitem}
\usepackage{float}
\usepackage{url}
\usepackage{natbib}
\usepackage{makecell}
\usepackage[compact]{titlesec}
\usepackage{listings}
\definecolor{promptbg}{gray}{0.94}
\lstdefinestyle{promptstyle}{%
  basicstyle=\footnotesize\ttfamily,%
  backgroundcolor=\color{promptbg},%
  frame=single,framerule=0pt,framesep=4pt,%
  breaklines=true,breakatwhitespace=false,%
  breakindent=0pt,breakautoindent=false,%
  columns=fullflexible,keepspaces=true,%
  showstringspaces=false,numbers=none,%
  xleftmargin=4pt,xrightmargin=4pt,%
  aboveskip=4pt,belowskip=4pt,%
}
\titlespacing*{\section}{0pt}{1.6ex plus .6ex minus .2ex}{1.0ex plus .3ex}
\titlespacing*{\subsection}{0pt}{1.2ex plus .4ex minus .2ex}{0.6ex plus .2ex}
\titlespacing*{\subsubsection}{0pt}{0.9ex plus .3ex minus .2ex}{0.4ex plus .2ex}

\newcommand{\cmark}{\textcolor{green!55!black}{\checkmark}}
\newcommand{\xmark}{\textcolor{red!75!black}{\texttimes}}

\definecolor{easygreen}{HTML}{E6F3E6}
\definecolor{mediumorange}{HTML}{FFF2E6}
\definecolor{hardblue}{HTML}{E6F2FF}

\newcommand{\pcbschemagen}{\textbf{PCBSchemaGen}}
\newcommand{\pcbbench}{\textsc{PCBBench}}
\newcommand{\uca}{UCA}

\newcommand{\skidl}{\texttt{SKiDL}}

\graphicspath{{figures/}{figure/}}

\AtBeginDocument{%
  \setlength{\abovedisplayskip}{3pt plus 1pt minus 1pt}%
  \setlength{\belowdisplayskip}{3pt plus 1pt minus 1pt}%
  \setlength{\abovedisplayshortskip}{2pt}%
  \setlength{\belowdisplayshortskip}{2pt}%
}
\title{PCBSchemaGen: Reward-Guided LLM Code Synthesis for Printed Circuit Boards (PCB) Schematic Design with Structured Verification}

\author{%
  Huanghaohe Zou$^{1,*}$ \quad Peng Han$^{1}$ \quad Emad Nazerian$^{1}$ \\[0.4em]
  Mafu Zhang$^{1}$ \quad Zhicheng Guo$^{2}$ \quad Alex Q. Huang$^{1,*}$ \\[0.7em]
  $^{1}$Semiconductor Power Electronics Center (SPEC), The University of Texas at Austin \\
  $^{2}$Arizona State University \\[0.3em]
  $^{*}$\,Corresponding authors: \texttt{hzou@utexas.edu}, \texttt{aqhuang@utexas.edu}%
}

\begin{document}

\maketitle

\begin{abstract}
Most LLM code-synthesis benchmarks rely on unit tests as the reward oracle, but PCB schematic design has none: correctness is defined by structured physical constraints over real IC packages and pin-level assignments, per-task golden references are unavailable, and SPICE simulation does not validate schematic-level correctness. We introduce \pcbschemagen{}, a training-free inference-time framework that turns a frozen LLM into a verifiable, repairable PCB schematic generator. The framework induces a domain schema from IC datasheets to ground LLM decoding, pairs it with a deterministic 5-layer continuous-reward verifier with pin-level error localization, and refines candidates through a Thompson Sampling arm-acquiring bandit. We evaluate on 2 PCB benchmarks covering 227 real-IC tasks across 22 unified circuit domains, including a public-schematic-derived suite that serves as a fully held-out generalization test (verifier, KG library, and prompts frozen before any evaluation). Under our framework, an open-weight 31B model (Gemma-4-31B) passes 81.3\% of \pcbbench{} tasks on average, and the same framework transfers across both benchmarks with zero verifier code changes; a Circuitron-style inference-time prompting baseline on the same Gemma-4-31B backbone collapses on hard system-level designs. This suggests inference-time refinement under a deterministic structural verifier is a general recipe for reference-free LLM code synthesis in domains without unit-test oracles. Our benchmarks and deterministic verifier are publicly available at \url{https://github.com/HZou9/PCBSchemaGen_v2}.
\end{abstract}
\section{Introduction}
Code-synthesis benchmarks typically rely on unit tests as the reward oracle, an assumption that breaks for engineering domains such as PCB schematic design, where correctness depends on structured physical constraints, real IC packages, and topology-level invariants, and per-task golden references are unavailable. This paper builds an inference-time LLM code-synthesis framework under these conditions.

PCB schematic design bridges IC selection and board-level layout, underpinning virtually every modern electronic system from consumer devices to electric vehicles and generative-AI hardware infrastructure~\cite{pcb_pecht2009prognostics,pe_freedm2011,pe_gan2018}. Digital RTL generation has been actively explored; generative LLM work for PCB schematic synthesis remains uncommon, and the traditional workflow relies on months of manual iteration over real IC packages and pin-level constraints. Three challenges block progress. \emph{(i) Data scarcity:} unlike digital RTL with abundant open code and benchmarks~\cite{bench_verilogeval2023,chen2021}, PCB schematics are not available at the scale required for large-scale training. \emph{(ii) No reference-free oracle:} per-task golden references are unavailable, and SPICE~\cite{spice_nagel1973,analog_masalachai2024}, the standard analog verifier, targets idealized functional simulation rather than schematic-level correctness under real IC and pin-level constraints; existing PCB-side work targets layout suggestion~\cite{pcb_layoutcopilot2024} or netlist-to-schematic transformation~\cite{pcb_schemato2024} rather than generative design from natural-language specs. \emph{(iii) Real IC components:} unlike analog/digital tasks operating on idealized devices, PCB schematics interface with real ICs whose packages and pin constraints differ across vendors, so generation from fixed datasets or fine-tuned LLMs cannot adapt across chips.

We introduce \pcbschemagen{}, a training-free inference-time framework that uses an LLM to generate \skidl{}~\cite{tool_skidl} Python programs which compile into KiCad-compatible schematics, netlists, and PCB projects. The framework grounds LLM decoding in a schema-induced pin-role ontology, verifies outputs under a deterministic 5-layer structural oracle, and closes a single \emph{propose}\,$\to$\,\emph{verify}\,$\to$\,\emph{refine} loop. Our contributions are:

\begin{itemize}[leftmargin=1.5em,itemsep=2pt,topsep=2pt,parsep=0pt]
  \item \textbf{Schema-induced knowledge graph:} a 32-role pin ontology and four constraint-predicate templates extracted automatically per IC, grounding LLM decoding in real package and pin constraints.

  \item \textbf{5-layer continuous-reward oracle:} a deterministic hierarchy of electrical, role, template, topology, and power invariants returning rewards in $[0,1]$ with pin-level error localization at zero LLM inference cost (a dense signal that learned PRMs~\citep{wang2024mathshepherd,setlur2024pav} obtain only via Monte-Carlo rollouts).

  \item \textbf{Thompson Sampling arm-acquiring bandit:} reward-informed Beta priors and adaptive temperature; we adapt the Bayesian-regret scaling argument of~\citet{russo2018tutorial} only as motivation for the policy choice; the operative finite-budget evidence at $T{=}4$ is the 5-strategy ablation in \Cref{sec:experiments}.

  \item \textbf{Two generative PCB schematic benchmarks:} \pcbbench{} (62 hand-authored tasks, 41 real ICs, 22 domains) and \textsc{Open-Schematics-Eval} (165 tasks from open-source schematics~\cite{data_openschematics}, 439 real ICs, same 22 domains), the first such benchmarks with deterministic verification to our knowledge. The same framework transfers across both suites unchanged; we evaluate up to 11 LLMs.
\end{itemize}
\vspace{-10pt}
\section{Preliminaries and Related Work}
\label{sec:related_work}
\vspace{-8pt}

\paragraph{Traditional PCB schematic design workflow.}
PCB design proceeds through IC selection, schematic capture, placement, routing, and prototyping; schematic capture is the bottleneck because experts must select correct ICs, fix pin- and package-level interconnections, and iterate until the design is electrically valid. SPICE~\cite{spice_nagel1973,analog_masalachai2024} simulates idealized analog behavior \emph{given} a netlist that is already correct; it presupposes the schematic exists and cannot validate IC choice, real package pinout, or pin-level connectivity, which is the schematic-capture decision space we target. Recent automation addresses adjacent stages only: LayoutCopilot~\cite{pcb_layoutcopilot2024} for interactive layout, FanoutNet~\cite{ml_fanoutnet2023} for fan-out routing, Autocircuit-RL~\cite{autocircuitrl2025} for RL-driven topology exploration. Two concurrent unpublished submissions target adjacent points: SchGen~\cite{schgen_anonymous2026} is supervised on a fixed schematic corpus (not training-free), and OmniSch~\cite{lu2026omnisch} evaluates multimodal image-to-graph reasoning rather than generation under deterministic structural constraints. We address the remaining gap: generate a structurally correct PCB schematic from a natural-language spec under real IC pin and package constraints, with no per-task golden reference and no functional simulator.

\paragraph{Digital RTL generation.}
Chip-Chat~\cite{rtl_chipchat2023}, ChipNeMo~\cite{rtl_chipnemo2023}, VeriGen~\cite{rtl_verigen2024}, DeepRTL~\cite{rtl_deeprtl2025}, BetterV~\cite{rtl_betterv2024}, the QiMeng series~\cite{rtl_qimengsalv2025,rtl_qimengcodevr12025}, AutoChip~\cite{rtl_autochip2023}, and GPT4AIGChip~\cite{rtl_gpt4aigchip2023} target Verilog generation through fine-tuning, in-context learning, or iterative feedback. All operate on internal digital logic without real IC packages or pin-level constraints, and PCB schematic data scarcity blocks the supervised fine-tuning that drives most of these methods.

\paragraph{Analog circuits.}
Analog work targets idealized device topologies: AnalogCoder~\cite{analog_analogcoder2025} models circuits as PySpice with prompt engineering; the AnalogGenie series~\cite{analog_analoggenie2025,analog_analoggenielite2025} discovers topologies via graph generation; Artisan~\cite{analog_artisan2024}, LaMAGIC2~\cite{analog_lamagic22025}, and AmpAgent~\cite{analog_ampagent2024} require human netlist priors or large topology datasets; CktGNN~\cite{ml_cktgnn2023} synthesizes via GNNs; ADO-LLM~\cite{analog_adollm2024} couples an LLM with Bayesian optimization; PE-GPT~\cite{analog_pegpt2025} uses RAG and emits partial sub-circuits without complete pin assignments; TopoSizing~\cite{analog_toposizing2025} is non-generative; RL-based IC layout~\cite{ml_graphplacement2021,ml_chipformer2023} and \cite{compare_CIRCUITSYNTH} operate at idealized analog scale. None bridges to real IC packages and pin-level connectivity, the regime PCB schematic synthesis requires.

\paragraph{Board-level PCB generative work.}
Existing PCB-side work covers settings adjacent to schematic generation: LayoutCopilot~\cite{pcb_layoutcopilot2024} targets post-schematic interactive layout, PCBRouting~\cite{pcb_pcbrouting2025} addresses routing, and \cite{compare_LLMfootprint} probes LLM understanding of chip packages. Verification-side approaches inherit similar limitations: most generative methods are one-pass without closed-loop verification~\cite{tool_selfdebug2023}; Self-Debug and MetaGPT~\cite{tool_selfrepair2023,tool_metagpt2023} provide self-repair without injecting IC-datasheet constraints; AMSNet-KG~\cite{analog_amsnetkg2025} builds a fixed-dataset KG; CircuitLM~\cite{compare_circuitlm2026} verifies via LLM judges rather than deterministic predicates; EESchematic~\cite{compare_eeschematic2025} uses vision-LLMs on idealized small circuits, and vision-LLMs themselves are documented to make precision errors at pin level~\cite{llm_vllm}; CROP~\cite{compare_CROP2025} embeds Verilog rather than producing schematic from natural language; \cite{compare_yan2022ai} stops at component-level planning; \cite{compare_Jha2024} adds Verilog-style syntactic feedback; \cite{compare_GNN} explores GNNs on 4-pin tasks without feedback. Concurrent LLM-for-EDA work has scope gaps: Schemato~\cite{matsuo2025schemato} converts netlist to schematic rather than synthesizing from natural language; PrefixLLM~\cite{xiao2024prefixllm} targets prefix-adder circuits only; the LLM-verification survey~\cite{liu2026llm} catalogues techniques without a structural verifier; NL2GDS~\cite{eland2026nl2gds} targets digital chip layout rather than board-level schematic; and~\cite{sestito2026flexible} composes LLM tools without grounding in pin-level IC constraints. None combines schema-induced KG retrieval with a deterministic structural verifier under real IC pin constraints.

\paragraph{Inference-time refinement and process rewards.}
Inference-time iterative code synthesis takes several reward shapes: UnitCoder~\citep{unitcoder2025} drives general code generation via unit-test feedback, B-Coder~\citep{bcoder2024} applies value-based RL over unit-test rewards, FunPRM~\citep{funprm2026} layers function-level process rewards, ALGO~\citep{zhang2023algo} uses LLM-generated unit-test oracles, and Zero-to-CAD~\citep{zerotocad2026} extends agentic feedback-driven search to CAD without supervised data. Learned process reward models such as Math-Shepherd~\citep{wang2024mathshepherd}, PAV~\citep{setlur2024pav}, and BeyondOracle~\citep{beyondoracle2025} require Monte Carlo rollouts or per-step annotation. Unit-test, LLM-judge, and Monte-Carlo-trained reward channels do not transfer to PCB schematics, where neither executable tests nor scalar simulators exist; \pcbschemagen{} instead uses a deterministic structural verifier built on a closed library of electrical and physical invariants, which is cheaper to evaluate at inference time and easier to audit. \Cref{tab:method-comparison} summarizes the capability gap.

\begin{table}[t]
\centering
\fontsize{6pt}{7pt}\selectfont
\caption{Iterative refinement frameworks for LLM code synthesis across five capability axes (\cmark{} supported, \xmark{} absent).}
\label{tab:method-comparison}
\setlength{\tabcolsep}{1.5pt}
\renewcommand{\arraystretch}{0.92}
\begin{tabular}{@{}ccccccc@{}}
\toprule
\textbf{Method}
& \textbf{\makecell{Training-\\Free}}
& \textbf{\makecell{Continuous\\Reward}}
& \textbf{\makecell{Multi-arm\\Search}}
& \textbf{\makecell{Error\\Localization}}
& \textbf{\makecell{Real IC\\Deployment}}
& \textbf{Application} \\
\midrule
Self-Refine~\citep{madaan2023selfrefine}     & \cmark & \xmark & \xmark & \xmark & \xmark & code/text \\
Reflexion~\citep{shinn2023reflexion}         & \cmark & \xmark & \xmark & \xmark & \xmark & code/text \\
RLEF~\citep{rlef2025}                        & \xmark & \cmark & \xmark & \xmark & \xmark & code \\
REx~\citep{tang2024rex}                      & \cmark & \xmark & \cmark & \xmark & \xmark & code \\
VeriCoder~\citep{vericoder2025}              & \xmark & \xmark & \xmark & \xmark & \xmark & Verilog \\
AutoCircuit-RL~\citep{autocircuitrl2025}     & \xmark & \cmark & \xmark & \xmark & \xmark & analog topology \\
AnalogCoder~\citep{analog_analogcoder2025}   & \cmark & \xmark & \xmark & \xmark & \xmark & analog (SPICE) \\
\midrule
\textbf{\pcbschemagen{} (ours)} & \cmark & \cmark & \cmark & \cmark & \cmark & \textbf{\makecell{PCB schematic\\(analog/digital/power)}} \\
\bottomrule
\end{tabular}
\end{table}

\begin{figure}[!t]
  \centering
  \makebox[\textwidth][c]{\includegraphics[width=1.2\textwidth]{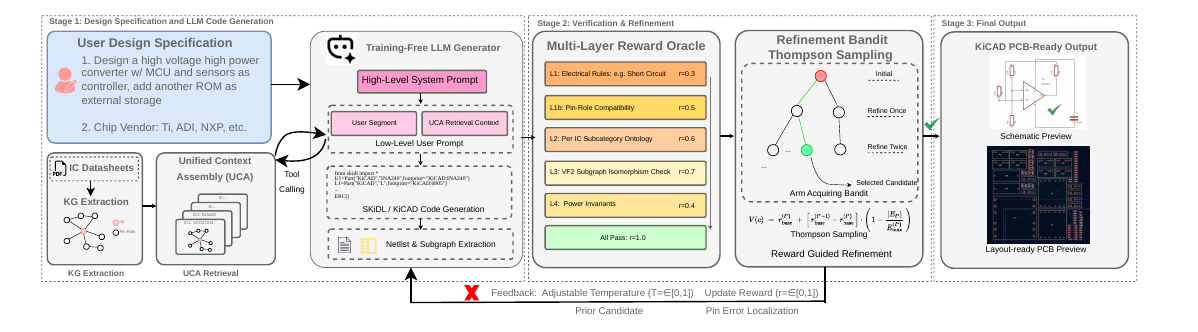}}
  \caption{\pcbschemagen{} pipeline: schema-induced KG $\to$ frozen-LLM \skidl{} generation $\to$ 5-layer reward oracle ($r \in [0,1]$, pin-localized errors) $\to$ Thompson Sampling bandit refinement $\to$ KiCad-ready schematic and PCB project. Training-free, no per-task golden netlist.}
  \label{fig:workflow_overview}
\end{figure}

\section{Approach}
\label{sec:method}
\vspace{-4pt}

We formalize the synthesis task (\Cref{subsec:formulation}) and describe four reusable inference-time components: a schema-induced KG with unified context assembly (\Cref{subsec:kg}), a 5-layer continuous-reward verifier (\Cref{subsec:verification}), and a Thompson Sampling refinement bandit (\Cref{subsec:bandit}).

\vspace{-4pt}
\subsection{Problem Formulation}
\label{subsec:formulation}
\vspace{-2pt}

\paragraph{Task and objective.}
A task $\mathcal{T} = (d, \mathcal{I}, \mathcal{O}, \mathcal{C})$ specifies a natural-language description $d$, I/O nets $\mathcal{I}, \mathcal{O}$, and a required-component set $\mathcal{C}$; a candidate $c \in \mathcal{P}$ is \skidl{} code that, when executed, instantiates IC components and connects pins to nets; a deterministic verifier $\mathcal{V}: \mathcal{P} \to [0,1]$ returns a continuous reward (\Cref{subsec:verification}). Given a budget of $T$ LLM queries, we seek a policy $\pi$ maximizing expected cumulative reward, equivalently minimizing cumulative regret:
\begin{equation}
    \pi^\star = \arg\max_\pi \mathbb{E}_{c_t \sim \pi}\!\left[\sum_{t=1}^{T} \mathcal{V}(c_t)\right],
    \quad
    \mathcal{R}_T(\pi) = \mathbb{E}\!\left[\sum_{t=1}^{T} \bigl(\mathcal{V}^\star - \mathcal{V}(c_t)\bigr)\right],
    \;\; \mathcal{V}^\star := \max_{c \in \mathcal{P}} \mathcal{V}(c).
    \label{eq:objective}
\end{equation}

\paragraph{Arm-acquiring bandit problem.}
Refinement reduces to an arm-acquiring bandit problem~\citep{tang2024rex,russo2018tutorial} with at most $K_t = t$ arms at time $t$. The framework is gradient-free and deployable on closed-source LLMs, orthogonal to weight-update methods such as RLEF~\citep{rlef2025} and SCoRe~\citep{kumar2024score}.

\vspace{-4pt}
\subsection{Schema-Induced Knowledge Graph}
\label{subsec:kg}
\vspace{-2pt}

A per-IC entry $\mathcal{K}_{\mathrm{IC}}$, induced automatically from the datasheet PDF, encodes a pin-role assignment from a 32-role closed ontology, four types of instantiated constraint predicates, and any subcircuit-template tags justified by the application-circuit section (full enumerations in \Cref{app:pin_roles,app:constraints}). Vendor datasheets are the industry's ground-truth reference for pin and electrical specifications, so the schema extracted from them inherits that authority. The ontology and predicate templates are fixed; rule \emph{contents} are populated afresh per IC, and specialized parts are expressed by composing existing roles and templates rather than by extending the ontology.

\vspace{-6pt}
\paragraph{Construction pipeline.}
A three-stage LLM pipeline~\citep{zhang2024edc,mo2025kggen} (illustrated in \Cref{fig:kg}, full description in \Cref{app:kg_pipeline}): (1) layout-aware PDF parsing of the pin table and application-circuit text; (2) low-temperature LLM pin-role assignment from the closed ontology; (3) predicate-template instantiation against the labeled pins. Each $\mathcal{K}_{\mathrm{IC}}$ is $\sim$300 tokens of JSON; per-stage precision/recall/F1 against an engineer-labeled subset are in \Cref{app:kg_pipeline}.

\vspace{-6pt}
\paragraph{Unified context assembly (\uca{}).}
\label{para:uca}
Each task uses only 2--6 of the 47 KG-library entries, so injecting the full library wastes prompt tokens. \uca{} performs two-stage retrieval at decoding time: a catalog selects candidate ICs from $(d, \mathcal{C})$, then their pin tables, application text, and predicates expand into the LLM context; the same substrate also delivers the bandit's $\mathrm{feedback}(E_{a^\star})$ during refinement. \uca{} matches or exceeds full-context pass rate at half the prompt cost (\Cref{tab:context_icl}).

\begin{figure}[b]
\centering
\begin{subfigure}[b]{0.55\columnwidth}
\centering
\includegraphics[width=\columnwidth,height=3.4cm,keepaspectratio]{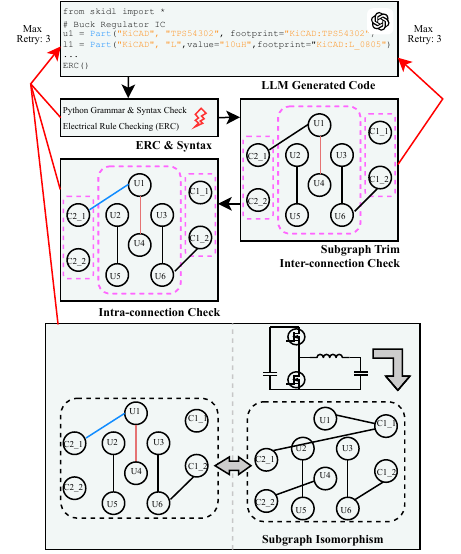}
\caption{Five-layer reward hierarchy.}
\label{fig:verifier_a}
\end{subfigure}\hfill
\begin{subfigure}[b]{0.42\columnwidth}
\centering
\includegraphics[width=\columnwidth,height=3.4cm,keepaspectratio]{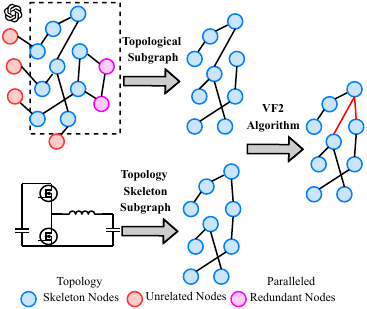}
\caption{Layer-3 topology check via VF2~\citep{tools_sub}.}
\label{fig:verifier_b}
\end{subfigure}
\caption{The verification oracle.}
\label{fig:verifier}
\end{figure}

\vspace{-4pt}
\subsection{Verification Oracle as a Process Reward Model}
\label{subsec:verification}
\vspace{-2pt}

The verifier $\mathcal{V}: \mathcal{P} \to [0,1]$ is a \emph{deterministic, inference-time process reward model}~\citep{wang2024mathshepherd,setlur2024pav,beyondoracle2025,funprm2026} that reshapes the binary task-success signal into a dense, structurally decomposed continuous reward localized to a specific layer, pin, or predicate, at zero LLM inference cost. It serves as the bandit's (\Cref{subsec:bandit}) online reward signal and is a closed-form alternative to learned PRMs, which require Monte Carlo rollouts or per-step human annotation. Within an end-to-end PCB design flow, $\mathcal{V}$ acts as one upstream gate at the schematic-synthesis stage: a verifier-passing candidate proceeds to KiCad ERC, KiCad DRC, reference-design cross-check, and design-review-board sign-off before placement, routing, and the manufacturing handoff, all of which remain unchanged by the introduction of $\mathcal{V}$.

\vspace{-6pt}
\paragraph{Five-layer hierarchy.}
Each candidate $c$ compiles to a bipartite graph $G = (V_C \cup V_N, E)$ over components $V_C$ and nets $V_N$ (\Cref{fig:verifier}a). The verifier evaluates five layers in priority order: L1 ERC (electrical invariants), L1b pin-role compatibility, L2 subcategory templates, L3 topology signatures, and L4 power invariants. Each layer $\ell$ carries a base reward $r^{(\ell)}_{\mathrm{base}} \in [0,1]$ assigned by severity; layer pseudocode and per-layer error categories are in \Cref{app:verification_details}.

\vspace{-6pt}
\paragraph{Continuous reward.}
Let $\ell^\star$ be the highest-priority failing layer and $|E_{\ell^\star}|$ the number of errors there ($\ell^\star = \infty$ when all layers pass, in which case $\mathcal{V}(c) = 1$). $\mathcal{V}$ interpolates linearly between the failing and next-layer base rewards:
\begin{equation}
\mathcal{V}(c) \;=\; r^{(\ell^\star)}_{\mathrm{base}} + \bigl(r^{(\ell^\star+1)}_{\mathrm{base}} - r^{(\ell^\star)}_{\mathrm{base}}\bigr) \cdot \Bigl(1 - \min\!\bigl(\tfrac{|E_{\ell^\star}|}{E_{\max}},\, 1\bigr)\Bigr) \;\in\; [0,1],
\label{eq:partial}
\end{equation}
with $E_{\max}$ a per-layer normalizing constant. The per-layer base rewards $\{r^{(\ell)}_{\mathrm{base}}\}$ and normalizing constants $\{E_{\max}^{(\ell)}\}$ are fixed once from IC-vendor application notes (typical instance counts and severity ranks per layer) and frozen before any benchmark evaluation; the full set of values is listed in \Cref{app:verification_details}. This decomposition is a form of reward shaping~\citep{ng1999reward}: the layer hierarchy defines a monotone potential, so any improvement in a higher-priority layer strictly increases $\mathcal{V}(c)$.

\vspace{-6pt}
\paragraph{Topology check via subgraph isomorphism.}
Layer 3 reduces to subgraph isomorphism between a generated graph $G_{\mathrm{gen}}$ and a domain-level topology template $G_{\mathrm{tmpl}}$ (e.g., half-bridge, synchronous-buck, or pi-filter skeletons), solved by VF2~\citep{tools_sub} (\Cref{fig:verifier}b). All templates are extracted once from IC-vendor application notes and reused across tasks; no template is derived from any task in \pcbbench{} or \textsc{Open-Schematics-Eval}, so $G_{\mathrm{tmpl}}$ does not carry per-task golden-netlist information. The three-layer tolerance mechanism (key-component filter, boolean fingerprinting, passive-element grouping) and per-task pattern-selection rules are in \Cref{app:verification_details}.

\vspace{-4pt}
\subsection{Refinement as an Arm-Acquiring Bandit Problem}
\label{subsec:bandit}
\vspace{-2pt}

We frame iterative refinement as an arm-acquiring bandit problem~\citep{tang2024rex,russo2018tutorial}; three extensions over REx exploit our continuous reward signal: reward-informed Beta prior, fractional posterior update, and adaptive temperature.

\begin{algorithm}[H]
\caption{Reward-guided arm-acquiring bandit refinement. $\mathrm{feedback}(E_{a^\star})$ formats the structured error set $E_{a^\star}$ from the failing layer into a pin-/predicate-localized natural-language message appended to the prompt (\Cref{app:verification_details}).}
\label{alg:bandit}
\DontPrintSemicolon
\SetKwInOut{Input}{Input}
\Input{Task $\mathcal{T}$, prompt $\mathbf{m}_0$, KG $\mathcal{K}$, total LLM-call budget $T$, base temperature $\tau_{\mathrm{base}}$.}
$c_0 \leftarrow \mathrm{LLM}(\mathbf{m}_0,\, \tau_{\mathrm{base}})$;\quad $r_0 \leftarrow \mathcal{V}(c_0)$\;
Initialize arm $a_0$ with $(\alpha_0, \beta_0) = (1 + 5 r_0,\, 1 + 5(1 - r_0))$;\quad $\mathcal{A} \leftarrow \{a_0\}$\;
\For{$t = 1, \ldots, T-1$ \textbf{while} $r_{t-1} < 1$}{
  Sample $\tilde{\theta}_i \sim \mathrm{Beta}(\alpha_i, \beta_i)$ for each $a_i \in \mathcal{A}$;\quad $a^\star \leftarrow \arg\max_i \tilde{\theta}_i$\;
  $\tau_t \leftarrow \mathrm{clamp}(\tau_{\mathrm{base}} \cdot \gamma(\mathrm{type}_{a^\star}, r_{a^\star}),\, 0.1,\, 1.0)$\;
  $c_t \leftarrow \mathrm{LLM}(\mathbf{m}_{a^\star} \Vert \mathrm{feedback}(E_{a^\star}),\, \tau_t)$;\quad $r_t \leftarrow \mathcal{V}(c_t)$\;
  $\alpha_{a^\star} \mathrel{+}= r_t$;\quad $\beta_{a^\star} \mathrel{+}= (1 - r_t)$\;
  Register $a_{\mathrm{new}}$ with $(1 + 5 r_t,\, 1 + 5(1 - r_t))$;\quad $\mathcal{A} \leftarrow \mathcal{A} \cup \{a_{\mathrm{new}}\}$\;
}
\Return{$\arg\max_{c \in \{c_0, \ldots, c_{T-1}\}} \mathcal{V}(c)$}\;
\end{algorithm}

\begin{figure}[t]
\centering
\includegraphics[width=\columnwidth]{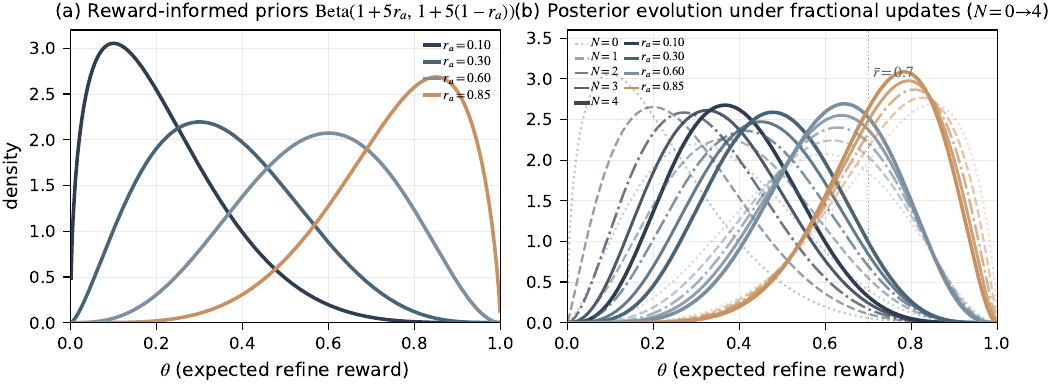}
\caption{Reward-informed prior gives each new arm a head-start matching parent quality, then data takes over. (a) At $N{=}0$ the prior is centered on $r_a$ (high-$r_a$ arms start optimistic, low-$r_a$ pessimistic). (b) After four refines with mean $\bar r{=}0.7$, all posteriors converge toward $\bar r$.}
\label{fig:bandit}
\end{figure}

\textbf{Bandit setting.} At step $t$, the action set $\mathcal{A}_t$ grows by one arm per step ($K_t = t$); each arm $a_i$ maintains a Beta posterior $\theta_i \sim \mathrm{Beta}(\alpha_i, \beta_i)$ over its expected refine reward, the policy plays $a^\star \in \mathcal{A}_t$ via Thompson sampling, and the verifier returns $r_t = \mathcal{V}(c_t) \in [0,1]$. \textbf{Reward-informed prior.} Each newly registered arm $a$ with parent reward $r_a$ is initialized with a Beta prior centered on $r_a$:
\begin{equation}
\alpha_a^{\mathrm{prior}} = s + C \cdot r_a, \qquad \beta_a^{\mathrm{prior}} = s + C \cdot (1 - r_a),
\label{eq:prior}
\end{equation}
with $s=1$, $C=5$ fixed. The prior mean $(s+Cr_a)/(2s+C)$ tracks parent quality: $r_a=0.8$ starts at $\mathrm{Beta}(5,2)$, $r_a=0.1$ at $\mathrm{Beta}(1.5,5.5)$; where REx~\citep{tang2024rex} uses uniform priors under binary reward, our continuous oracle (\Cref{eq:partial}) injects quality-aware initialization at zero cost. \textbf{Fractional posterior update.} We extend Beta-Bernoulli conjugacy to fractional updates~\citep{agrawal2013further}:
\begin{equation}
\alpha_{a^\star} \leftarrow \alpha_{a^\star} + r_t, \qquad \beta_{a^\star} \leftarrow \beta_{a^\star} + (1 - r_t),
\label{eq:posterior}
\end{equation}
so a reward of $0.8$ contributes $4\times$ more $\alpha$-mass than $0.2$, capturing partial credit invisible to binary-reward TS. The fractional Beta update is a pragmatic heuristic, not a conjugate update for a well-defined continuous-reward likelihood; the regret discussion below adapts the Bernoulli TS scaling argument as motivation only, not as a finite-sample guarantee. \textbf{Thompson sampling action selection.} Each step samples
\begin{equation}
\tilde{\theta}_i \sim \mathrm{Beta}(\alpha_i, \beta_i), \qquad a^\star = \arg\max_{a_i \in \mathcal{A}_t} \tilde{\theta}_i.
\label{eq:thompson}
\end{equation}
\textbf{Adaptive temperature.} LLM regeneration temperature adapts to the failing-layer category and parent reward:
\begin{equation}
\tau_t = \mathrm{clamp}\!\bigl(\tau_{\mathrm{base}} \cdot \gamma(\mathrm{type}_t, r_{a^\star}),\; 0.1,\; 1.0\bigr),
\label{eq:temperature}
\end{equation}
where $\gamma(\mathrm{type}_t, r_{a^\star})$ is a 5-case lookup over the failing-error category and parent reward: syntax~$\to$~$\gamma{=}0.6$, ERC (electrical)~$\to$~$\gamma{=}1.0$, topology with $r{\geq}0.5$~$\to$~$\gamma{=}0.8$ (surgical exploit on near-passing candidates), topology with $r{<}0.5$~$\to$~$\gamma{=}1.6$ (explore stuck branches), code-extraction~$\to$~$\gamma{=}1.4$; the bandit decides \emph{which} arm to refine, $\gamma$ controls \emph{how diversely} the LLM regenerates from it (\Cref{app:adaptive_temp}). The two layers compose into structural exploration: arm-acquisition (\Cref{eq:thompson}) introduces a fresh candidate at every step, and the adaptive $\gamma$ widens the token-level distribution from which each new candidate is sampled, so the trajectory $\{c_0, \ldots, c_T\}$ traverses regions of the candidate space beyond the deterministic neighborhood of any single initial proposal. \textbf{Understanding the bandit's behavior.} After arm $a$ accumulates $N_a$ refinements with cumulative reward $S_a$, the posterior mean has the closed form
\begin{equation}
\mathbb{E}[\theta_a \mid N_a, S_a] = \frac{s + C \cdot r_a + S_a}{2s + C + N_a}.
\label{eq:expected_reward}
\end{equation}
The prior $r_a$ dominates early; $S_a/N_a$ takes over as $N_a$ grows. \Cref{fig:bandit} illustrates: at $N{=}0$ priors separate by parent reward (head-start), and at $N{=}4$ all posteriors shift toward the observed mean $\bar{r}{=}0.7$, so an over- or under-optimistic prior is corrected by data rather than locking the bandit in. \textbf{Bayesian regret.} Adapting~\citet{russo2018tutorial,agrawal2013further} to arm-acquiring TS with continuous bounded rewards and reward-informed prior inheritance,
\begin{equation}
\mathbb{E}\bigl[\mathcal{R}_T\bigr] = \tilde{\mathcal{O}}(\sqrt{KT}),
\label{eq:regret}
\end{equation}
where $K \leq T$ is the final arm count; the bound is informative when $K \ll T$ and reduces to $\tilde{\mathcal{O}}(T)$ when every step opens a new arm. We treat \Cref{eq:regret} as a scaling-argument motivation only, not a finite-sample guarantee at $T{=}4$ (the 5-strategy ablation provides the actual finite-sample comparison). The proof (per-step information ratio $\Gamma_t \leq K/2$; chain-rule integration with $\mathcal{I}(A^\star;\mathcal{H}_T) \leq \log K$; history-measurable priors) is in \Cref{app:regret_bound}; greedy and fixed-budget retry admit linear regret when $c_0$ is structurally flawed.

\section{Benchmarks}
\label{sec:benchmark}
\vspace{-4pt}

\begin{figure}[H]
\centering
\includegraphics[width=0.85\columnwidth]{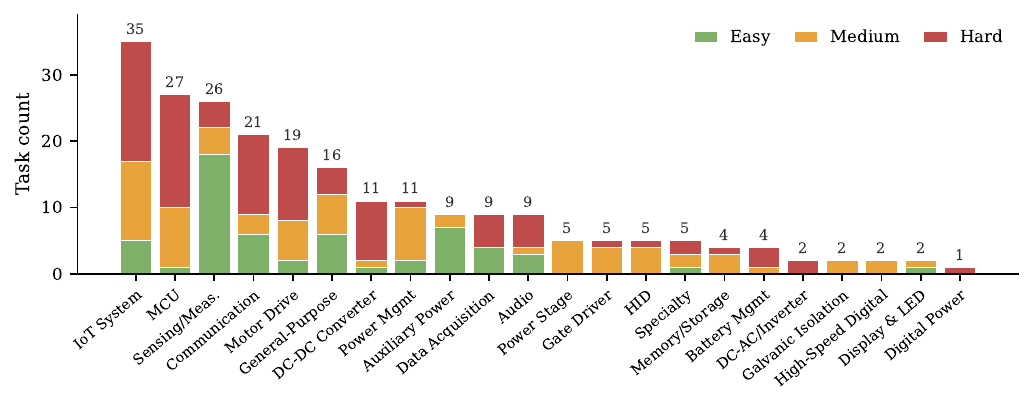}
\caption{Task counts per domain, by difficulty.}
\label{fig:domain_dist}
\end{figure}

We release two new benchmark suites for \emph{generative} PCB schematic synthesis. To the best of our knowledge, no prior benchmark targets this setting: the closest existing PCB benchmark~\citep{pcbbench_li} restricts LLMs to question-answering, which excludes generative tasks. Aggregate statistics in \Cref{tab:bench_summary}, domain distribution in \Cref{fig:domain_dist}.

\textbf{\pcbbench{} (human-authored, 62 tasks).}
\pcbbench{} contains \textbf{47} unique components (\textbf{41} commercial ICs such as STM32F103 and UCC27211, plus \textbf{6} generic passive families), with \textbf{2--54} active pins per task. The \textbf{62} tasks span all \textbf{22} unified circuit categories and split as \textbf{17 Easy / 28 Medium / 17 Hard}.

\textbf{Open-Schematics-Eval (165 tasks).}
Open-Schematics-Eval is extracted from real open-source PCB schematics in the \texttt{open-schematics} dataset~\citep{data_openschematics}, with \textbf{826} unique components (\textbf{439} real ICs: STM32, ATmega, etc.). The 439 OSE ICs are largely disjoint from \pcbbench{}'s 47-component curated library at the part-number level: the two suites share the same 22 unified circuit domains and the same component families (gate driver, MCU, power converter, etc.), but the specific commercial ICs differ. The \textbf{165} tasks split as \textbf{40 Easy / 48 Medium / 77 Hard}. Each task instantiates its required ICs through the same automated KG extraction pipeline described in \Cref{subsec:kg}: \pcbbench{}'s 47-component library is reused unchanged whenever an OSE task happens to share a part number, while OSE-specific ICs are populated on demand at evaluation time, so that the verifier code, the pin-role ontology, and the constraint-predicate templates are identical across both suites.

\textbf{Difficulty stratification.}
Both suites apply the same structural thresholds: \emph{Easy} = $\leq$2 ICs, $\leq$15 components, $\leq$25 nets, no MCU; \emph{Hard} = $\geq$6 ICs, $>$40 components, $>$60 nets, or MCU with $>$40 nets; \emph{Medium} = the remainder.

\vspace{-12pt}

\begin{table}[H]
\centering
\footnotesize
\caption{Aggregate statistics of the two benchmark suites.}
\label{tab:bench_summary}
\setlength{\tabcolsep}{4pt}
\renewcommand{\arraystretch}{1.05}
\begin{tabular}{l c c c c c c c}
\toprule
\textbf{Suite}              & \textbf{Total} & \textbf{Easy} & \textbf{Medium} & \textbf{Hard} & \textbf{Real ICs} & \textbf{Vendors} & \textbf{Domains} \\
\midrule
\pcbbench{} (hand-authored) &  62            &  17           & 28              & 17            &  41               & 11               & 22               \\
Open-Schematics-Eval        & 165            &  40           & 48              & 77            & 439               & 55               & 22               \\
\midrule
\textbf{Total}              & \textbf{227}   & \textbf{57}   & \textbf{76}     & \textbf{94}   & \textbf{480}      & \textbf{66}      & \textbf{22}      \\
\bottomrule
\end{tabular}
\end{table}

\vspace{-10pt}
\section{Experiments}
\label{sec:experiments}
\vspace{-8pt}

\textbf{Models.} 11 LLMs: 4 closed-source (Gemini~3.1~Pro/Flash-Lite~\citep{llm_gemini31pro_2026,llm_gemini31flashlite_2026}, GPT-5.4~\citep{llm_gpt54_2026}, MiniMax~M2.5~\citep{llm_minimaxm25_2026}) and 7 open-source (DeepSeek~V3.2~\citep{llm_deepseekv32_2025}, Gemma-4-31B~\citep{llm_gemma4_31b_2025}, Qwen3-Coder-30B~\citep{llm_qwen3coder30b_2025}, Devstral-24B~\citep{llm_devstral24b_2025}, Qwen3.5-27B~\citep{llm_qwen35_27b_2026.5}, Llama-4-Scout~\citep{llm_llama4scout_2025}, InternLM3-8B~\citep{llm_internlm3_8b_2025}).
\textbf{Baseline.} \textsc{Circuitron}~\citep{circuitron2025} on Gemma-4-31B (strongest open-source).
\textbf{Benchmarks.} \pcbbench{} (62 hand-authored) and \textsc{Open-Schematics-Eval} (OSE, 165 third-party); \Cref{tab:bench_summary}.
\textbf{Protocol.} Per (model, task): $n{=}15$ trajectories, each with a total LLM-call budget of $T{=}4$ (1 initial generation + up to 3 verifier-driven refinements; \Cref{alg:bandit}); a trajectory passes if any of its candidates is accepted. \textbf{Pass@1} is trajectory pass rate; \textbf{Pass@5} uses the unbiased estimator~\citep{chen2021}; refinement-gain decomposition in \Cref{tab:refinement_gain}.

\subsection{Main Results}
\label{subsec:main_results}
\vspace{-4pt}

\begin{table*}[!t]
\centering
\caption{Main results at $T{=}4$ refinement, $n{=}15$ trials/task. Each cell shows
Pass@1 (top) / Pass@5 (bottom), with Pass@5 computed per task using the unbiased
estimator $\widehat{\mathrm{pass@}k}{=}1{-}\binom{n-c}{k}/\binom{n}{k}$~\citep{chen2021}
and averaged over the tier. Per-row Pass@1 best in \textbf{bold}; ``--'' = not run (MiniMax M2.5 / Gemini 3.1 Pro on OSE omitted: API cost).}
\label{tab:main_results}
\scriptsize
\setlength{\tabcolsep}{3pt}
\renewcommand{\arraystretch}{0.95}
\makebox[\textwidth][c]{\begin{tabular}{l | c | ccccccc | cccc}
\toprule
& & \multicolumn{7}{c|}{\cellcolor{gray!12}\textbf{Open-source}} & \multicolumn{4}{c}{\cellcolor{gray!28}\textbf{Closed-source}} \\
\cmidrule(lr){3-9}\cmidrule(lr){10-13}
Difficulty
& \makecell{Circuitron \\ (Gemma-4)}
& \makecell{InternLM3 \\ -8B} & \makecell{Qwen3.5 \\ -27B} & \makecell{Qwen3- \\ Coder-30B}
& \makecell{Llama-4 \\ Scout} & \makecell{Devstral \\ -24B} & \makecell{Gemma-4 \\ -31B} & \makecell{DeepSeek \\ V3.2}
& \makecell{MiniMax \\ M2.5} & \makecell{Gemini-3.1 \\ Pro} & \makecell{Gemini-3.1 \\ Flash-Lite} & GPT-5.4 \\
\midrule
\multicolumn{13}{l}{\textit{\pcbbench{} (62 tasks). Top of cell: \textbf{Pass@1}; bottom (smaller): Pass@5.}} \\
\addlinespace[1pt]
\cellcolor{easygreen}    Easy   & \makecell[c]{53.3\\\scriptsize 99.3}  & \makecell[c]{27.1\\\scriptsize 66.0}   & \makecell[c]{54.5\\\scriptsize 94.0}   & \makecell[c]{83.1\\\scriptsize 99.6}   & \makecell[c]{81.2\\\scriptsize 93.9}   & \makecell[c]{93.3\\\scriptsize 99.1}   & \makecell[c]{\textbf{98.0}\\\scriptsize 100}    & \makecell[c]{97.6\\\scriptsize 100}    & \makecell[c]{94.9\\\scriptsize 100}    & \makecell[c]{96.9\\\scriptsize 100}    & \makecell[c]{94.5\\\scriptsize 96.1}   & \makecell[c]{95.3\\\scriptsize 98.4} \\
\addlinespace[1pt]
\cellcolor{mediumorange} Medium & \makecell[c]{54.2\\\scriptsize 99.3}  & \makecell[c]{11.7\\\scriptsize 32.6}   & \makecell[c]{44.9\\\scriptsize 88.3}   & \makecell[c]{54.0\\\scriptsize 84.7}   & \makecell[c]{61.4\\\scriptsize 87.0}   & \makecell[c]{78.3\\\scriptsize 95.4}   & \makecell[c]{91.9\\\scriptsize 97.6}   & \makecell[c]{88.8\\\scriptsize 98.4}   & \makecell[c]{88.3\\\scriptsize 99.4}   & \makecell[c]{\textbf{98.8}\\\scriptsize 100}    & \makecell[c]{89.0\\\scriptsize 97.4}   & \makecell[c]{98.6\\\scriptsize 100} \\
\addlinespace[1pt]
\cellcolor{hardblue}     Hard   & \makecell[c]{10.2\\\scriptsize 57.1}  & \makecell[c]{0.0\\\scriptsize 0.0}     & \makecell[c]{21.2\\\scriptsize 58.6}   & \makecell[c]{5.9\\\scriptsize 18.0}    & \makecell[c]{2.0\\\scriptsize 8.6}     & \makecell[c]{16.9\\\scriptsize 44.2}   & \makecell[c]{47.1\\\scriptsize 80.0}   & \makecell[c]{53.7\\\scriptsize 90.1}   & \makecell[c]{36.9\\\scriptsize 72.8}   & \makecell[c]{84.3\\\scriptsize 97.5}   & \makecell[c]{35.7\\\scriptsize 67.5}   & \makecell[c]{\textbf{85.1}\\\scriptsize 99.0} \\
\addlinespace[1pt]
                         All    & \makecell[c]{41.9\\\scriptsize 95.8}  & \makecell[c]{12.7\\\scriptsize 32.8}   & \makecell[c]{40.2\\\scriptsize 81.7}   & \makecell[c]{48.8\\\scriptsize 70.5}   & \makecell[c]{50.5\\\scriptsize 67.4}   & \makecell[c]{65.6\\\scriptsize 82.4}   & \makecell[c]{81.3\\\scriptsize 93.4}   & \makecell[c]{81.6\\\scriptsize 96.6}   & \makecell[c]{76.0\\\scriptsize 92.3}   & \makecell[c]{\textbf{94.3}\\\scriptsize 99.3}   & \makecell[c]{75.9\\\scriptsize 88.9}   & \makecell[c]{94.0\\\scriptsize 99.3} \\
\midrule
\multicolumn{13}{l}{\textit{\textsc{Open-Schematics-Eval} (165 tasks; \emph{zero verifier code change}).}} \\
\addlinespace[1pt]
\cellcolor{easygreen}    Easy   & \makecell[c]{48.2\\\scriptsize 98.1}  & \makecell[c]{8.3\\\scriptsize 23.0}    & \makecell[c]{69.3\\\scriptsize 87.2}   & \makecell[c]{46.0\\\scriptsize 64.8}   & \makecell[c]{29.7\\\scriptsize 58.2}   & \makecell[c]{51.0\\\scriptsize 74.3}   & \makecell[c]{86.8\\\scriptsize 94.5}   & \makecell[c]{76.2\\\scriptsize 94.2}   & \makecell[c]{--\\\scriptsize --}        & \makecell[c]{--\\\scriptsize --}        & \makecell[c]{72.5\\\scriptsize 89.6}   & \makecell[c]{\textbf{87.8}\\\scriptsize 96.3} \\
\addlinespace[1pt]
\cellcolor{mediumorange} Medium & \makecell[c]{26.0\\\scriptsize 84.6}  & \makecell[c]{0.8\\\scriptsize 4.0}     & \makecell[c]{44.4\\\scriptsize 76.4}   & \makecell[c]{15.1\\\scriptsize 34.4}   & \makecell[c]{9.0\\\scriptsize 25.8}    & \makecell[c]{24.2\\\scriptsize 51.1}   & \makecell[c]{66.1\\\scriptsize 82.7}   & \makecell[c]{43.5\\\scriptsize 70.4}   & \makecell[c]{--\\\scriptsize --}        & \makecell[c]{--\\\scriptsize --}        & \makecell[c]{61.4\\\scriptsize 83.2}   & \makecell[c]{\textbf{86.0}\\\scriptsize 97.7} \\
\addlinespace[1pt]
\cellcolor{hardblue}     Hard   & \makecell[c]{3.2\\\scriptsize 8.3}    & \makecell[c]{0.1\\\scriptsize 0.4}     & \makecell[c]{31.1\\\scriptsize 57.4}   & \makecell[c]{5.9\\\scriptsize 19.1}    & \makecell[c]{2.6\\\scriptsize 8.6}     & \makecell[c]{12.2\\\scriptsize 31.5}   & \makecell[c]{63.4\\\scriptsize 83.9}   & \makecell[c]{30.3\\\scriptsize 58.9}   & \makecell[c]{--\\\scriptsize --}        & \makecell[c]{--\\\scriptsize --}        & \makecell[c]{37.5\\\scriptsize 65.0}   & \makecell[c]{\textbf{70.7}\\\scriptsize 86.0} \\
\addlinespace[1pt]
                         All    & \makecell[c]{20.7\\\scriptsize 48.4}  & \makecell[c]{2.3\\\scriptsize 6.9}     & \makecell[c]{44.2\\\scriptsize 70.1}   & \makecell[c]{18.3\\\scriptsize 34.7}   & \makecell[c]{11.0\\\scriptsize 25.6}   & \makecell[c]{25.1\\\scriptsize 47.6}   & \makecell[c]{69.8\\\scriptsize 86.1}   & \makecell[c]{45.3\\\scriptsize 70.8}   & \makecell[c]{--\\\scriptsize --}        & \makecell[c]{--\\\scriptsize --}        & \makecell[c]{52.9\\\scriptsize 76.3}   & \makecell[c]{\textbf{79.3}\\\scriptsize 91.9} \\
\bottomrule
\end{tabular}}
\end{table*}

\begin{table}[!t]
\centering
\caption{Refinement gain: Pass@1 at $T{=}1$ vs.\ $T{=}4$, averaged across 11 models on \pcbbench{} and 9 on OSE; $\delta$ is the absolute gain.}
\label{tab:refinement_gain}
\footnotesize
\setlength{\tabcolsep}{4pt}
\renewcommand{\arraystretch}{1.0}
\begin{tabular}{l | ccc | ccc}
\toprule
& \multicolumn{3}{c|}{\pcbbench{} (11 models)} & \multicolumn{3}{c}{\textsc{Open-Schematics-Eval} (9 models)} \\
Difficulty & $T{=}1$ & $T{=}4$ & $\delta$ & $T{=}1$ & $T{=}4$ & $\delta$ \\
\midrule
\cellcolor{easygreen}    Easy    & 59.8 & 83.3 & +23.5 & 32.0 & 58.6 & +26.6 \\
\cellcolor{mediumorange} Medium  & 45.8 & 73.2 & +27.4 & 14.9 & 38.9 & +24.0 \\
\cellcolor{hardblue}     Hard    &  8.2 & 35.4 & +27.2 &  9.8 & 28.2 & +18.4 \\
                         Overall & 39.2 & 65.6 & +26.4 & 16.7 & 38.7 & +22.0 \\
\bottomrule
\end{tabular}
\end{table}

\textbf{Results Analysis.} Closed-source frontier models (best: GPT-5.4) near-saturate Easy/Medium on \pcbbench{} ($\geq 95\%$) and reach $84$--$85\%$ on Hard. With our framework on the strongest open-source model (Gemma-4-31B), Pass@1 reaches $81.3\%$ overall ($47.1\%$ Hard), versus $41.9\%$ ($10.2\%$ Hard) for \textsc{Circuitron} on the same Gemma-4-31B backbone, $+39.4$pp overall and $+36.9$pp on Hard. Other strong open weights (DeepSeek-V3.2, $81.6\%$ overall) close most of the closed-vs-open gap but trail the frontier on Hard system-level designs.

\textbf{Held-out cross-benchmark generalisation.} The verifier, KG, decoding template, and bandit are frozen before any \textsc{Open-Schematics-Eval} task is authored; OSE (165 third-party tasks, $47\%$ Hard) is evaluated with zero code change as a held-out generalization test. Model ranking is preserved; the four strongest models retain $\geq 30\%$ on OSE Hard while \textsc{Circuitron} reaches $3.2\%$. GPT-5.4 transfers most strongly ($-14.7$pp), Gemma-4-31B follows ($-11.5$pp).

\textbf{Refinement effectiveness.} \Cref{tab:refinement_gain} attributes a substantial fraction of the headline pass rate to the $T{=}1{\to}4$ loop, with the largest gains on the Hard tier ($+27$pp on \pcbbench{}, $+18$pp on \textsc{Open-Schematics-Eval}). The smallest open-weight model in our pool (InternLM3-8B, $\sim 8$B parameters) collapses to single-digit Hard pass on both suites; the same KG and decoding context lift the larger open-weight models but not InternLM3-8B, indicating that the bottleneck at this scale is raw model parameter count rather than retrieval coverage.

\textbf{Verifier oracle trust.} On a stratified $N{=}300$ sample (50/50 oracle-pass/fail per tier), Cohen's $\kappa{=}0.91$ vs.\ senior PCB-engineer blind labels; $6.7\%$ FA (Wilson 95\% CI $[3.7, 11.8]$), $2.7\%$ FR ($[1.0, 6.7]$); $75\%$ of FR is ERC over-rejection. Lower-bound-corrected Pass@1 (subtracting FA) places Gemma-4-31B at $74.6\%$ / GPT-5.4 at $87.3\%$, preserving rankings. Verifier-accepted candidates are downstream-gated by KiCad ERC/DRC. Details: \Cref{app:human_eval,app:statistics}.

\subsection{Ablation Studies}
\label{subsec:ablation}
\vspace{-4pt}

\textbf{Component contributions.} \Cref{tab:component_ablation} decomposes the framework on Gemma-4-31B. Binary reward (B1) costs $-5.3$pp Overall and $-9.6$pp Hard. Pin-level feedback (D1) is the load-bearing localisation channel: category-level costs $-20.9$pp, binary $-23.6$pp. The schema-induced KG (C1) is the single largest contribution: removing it collapses Hard from $47.1\%$ to $2.0\%$ ($-45.1$pp); component-only KG still leaves $-11.0$pp on Hard.

\textbf{Context assembly and ICL.} \uca{} attains the highest Overall ($78.0\%$ vs.\ Two-Stage $75.0\%$, Full $73.7\%$) at half Full's token budget. Removing in-context exemplars drops Overall by $-6.0$pp, concentrated on Medium ($-10.8$pp), smaller on Hard ($-4.8$pp), none on Easy; the KG carries most structural grounding, ICL contributes primarily to Medium-tier disambiguation.

\textbf{Refinement strategy.} \Cref{tab:strategy_ablation} isolates the refinement loop from the arm-selection policy. The no-refinement Pass@4 baseline reaches $60.5\%$ Overall ($6.7\%$ Hard); any reasonable arm-selection lifts Overall to $75$--$78\%$, ours reaches $81.3\%$. The $+20.7$pp no-refinement$\to$ours gap is $\approx 7\times$ the $+2.9$pp Greedy$\to$ours gap, so the refinement loop dominates and arm-selection is second-order; Overall spans only $\pm 2.9$pp across the five surveyed policies.

\textbf{Verifier-layer ablation.} ERC alone yields an apparently higher pass rate ($98.3\%$ vs.\ $88.1\%$ Easy; $84.9\%$ vs.\ $72.4\%$ Medium/Hard); the $10$--$13$pp gap measures ERC's false-accept rate that L2--L4 catch.

\begin{table}[!t]
\centering
\caption{Component ablation (Gemma-4-31B), reported as three independent axes with their own scope; $\Delta$ is computed within each axis only.}
\label{tab:component_ablation}
\setlength{\tabcolsep}{1.5pt}
\renewcommand{\arraystretch}{0.95}

\noindent\hspace*{-0.05\textwidth}%
\begin{minipage}{1.10\textwidth}
\centering
\scriptsize

\begin{minipage}[t]{0.35\linewidth}
\centering
\textbf{B1: Reward signal}\\[-1pt]
\textit{20-task subset, 300 trials}\\[2pt]
\begin{tabular}{@{}l|cccc|r@{}}
\toprule
\textbf{Setting} & E & M & H & Ovr & $\Delta$ \\
\midrule
Binary $\{0,1\}$               & 97.3  & 86.7 & 39.0 & 72.7 & $-5.3$ \\
\textbf{Cont.\ $[0,1]$ (ours)} & 100.0 & 90.0 & 48.6 & 78.0 & ---    \\
\bottomrule
\end{tabular}
\end{minipage}\hfill
\begin{minipage}[t]{0.32\linewidth}
\centering
\textbf{C1: Knowledge graph}\\[-1pt]
\textit{62-task \pcbbench{}, 930 trials}\\[2pt]
\begin{tabular}{@{}l|cccc|r@{}}
\toprule
\textbf{Setting} & E & M & H & Ovr & $\Delta$ \\
\midrule
None                       & 43.5 & 31.7 &  2.0 & 26.8 & $-54.5$ \\
Component-only             & 95.3 & 92.1 & 36.1 & 77.6 & $-3.7$  \\
\textbf{Schema (ours)}     & 98.0 & 91.9 & 47.1 & 81.3 & ---     \\
\bottomrule
\end{tabular}
\end{minipage}\hfill
\begin{minipage}[t]{0.32\linewidth}
\centering
\textbf{D1: Feedback granularity}\\[-1pt]
\textit{62-task \pcbbench{}, 930 trials}\\[2pt]
\begin{tabular}{@{}l|cccc|r@{}}
\toprule
\textbf{Setting} & E & M & H & Ovr & $\Delta$ \\
\midrule
None (binary)              & 83.9 & 69.8 & 11.8 & 57.7 & $-23.6$ \\
Weak (category-level)      & 85.5 & 73.8 & 13.3 & 60.4 & $-20.9$ \\
\textbf{Pin-level (ours)}  & 98.0 & 91.9 & 47.1 & 81.3 & ---     \\
\bottomrule
\end{tabular}
\end{minipage}

\end{minipage}\hspace*{-0.05\textwidth}
\end{table}

\begin{table}[!t]
\centering
\caption{Refinement-strategy ablation (Gemma-4-31B, 62-task \pcbbench{}, 930 trials). First row: no-refinement Pass@4 baseline; rest: same loop with different arm-selection. $\Delta$: Overall gap to ours.}
\label{tab:strategy_ablation}
\scriptsize
\setlength{\tabcolsep}{4pt}
\renewcommand{\arraystretch}{1.0}
\begin{tabular}{l | cccc | r}
\toprule
\textbf{Strategy} & Easy & Medium & Hard & Overall & $\Delta$ \\
\midrule
Pass@4 i.i.d.\ sampling (no refinement)                                            & 91.0 & 74.8 &  6.7 & 60.54 & $-20.75$ \\
\midrule
$\epsilon$-greedy ($\epsilon{=}0.2$)~\citep{sutton2018rl}                          & 94.9 & 84.0 & 41.2 & 75.27 & $-6.02$  \\
Thompson Sampling~\citep{thompson1933}                                             & 97.6 & 84.8 & 40.4 & 76.13 & $-5.16$  \\
UCB1~\citep{auer2002ucb}                                                           & 98.4 & 87.4 & 40.4 & 77.50 & $-3.79$  \\
Greedy (best-mean arm)~\citep{sutton2018rl}                                        & 96.9 & 87.4 & 45.1 & 78.40 & $-2.89$  \\
\textbf{Thompson Sampling (ours)}~\citep{agrawal2013further}                       & 98.0 & 91.9 & 47.1 & \textbf{81.29} & ---      \\
\bottomrule
\end{tabular}
\end{table}

\begin{table}[!t]
\centering
\caption{Context-assembly mode and ICL ablation (Gemma-4-31B, 20-task subset, 300 trials).}
\label{tab:context_icl}
\scriptsize
\setlength{\tabcolsep}{4pt}
\renewcommand{\arraystretch}{1.0}
\begin{tabular}{l l | cccc | r}
\toprule
\textbf{Axis} & \textbf{Setting} & Easy & Medium & Hard & Overall & Tokens \\
\midrule
\multirow{3}{*}{\makecell[l]{Context\\mode}}
  & Full library                    & 100.0 & 83.3  & 43.8  & 73.7  & $\sim$35k \\
  & Two-Stage                       & 100.0 & 88.3  & 41.9  & 75.0  & $\sim$22k \\
  & \textbf{\uca{} (ours)}          & 100.0 & 90.0  & 48.6  & \textbf{78.0} & $\sim$18k \\
\midrule
\multirow{2}{*}{\makecell[l]{In-context\\exemplars}}
  & No ICL                          & 100.0 & 79.2  & 43.8  & 72.0  & --- \\
  & \textbf{With ICL (ours)}        & 100.0 & 90.0  & 48.6  & \textbf{78.0} & --- \\
\bottomrule
\end{tabular}
\end{table}

\vspace{-4pt}
\subsection{End-to-End Pipeline Demonstration}
\label{subsec:e2e_pipeline}
\vspace{-4pt}

\begin{figure}[!t]
\centering
\includegraphics[width=\columnwidth]{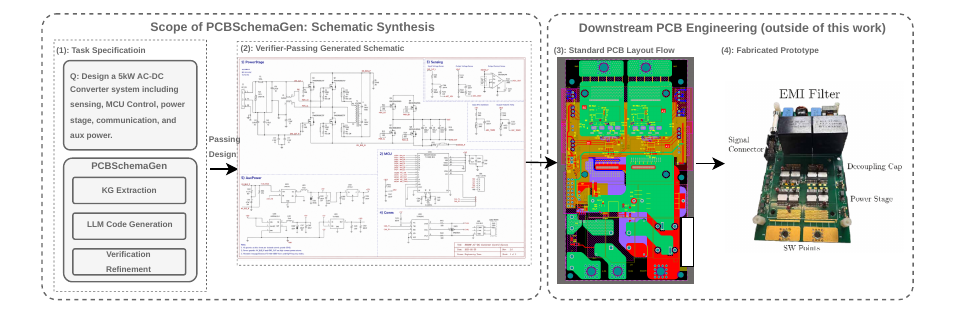}
\caption{End-to-end demonstration on a 5000~W AC-DC converter: \pcbschemagen{}'s scope (left, blue panel) covers KG extraction, LLM SKiDL generation, verification, and refinement; downstream PCB engineering (schematic editor, layout, fabrication) lies outside the framework's contribution.}
\label{fig:pipeline}
\end{figure}

Beyond pass rates, we verify the framework produces artifacts that enter a real PCB workflow. We applied the pipeline to a 5000~W AC-DC converter spanning five domains (Sensing, MCU, PowerStage, Comm, AuxPower); the verifier-accepted \skidl{} output exports to a KiCad schematic, netlist, and PCB project. \Cref{fig:pipeline} shows the resulting schematic, layout-ready PCB project, and a fabricated prototype; stage-by-stage detail is in \Cref{app:e2e_pipeline}.

\vspace{-10pt}
\section{Conclusion}
\label{sec:discussion}
\vspace{-8pt}

We presented \pcbschemagen{}, a training-free inference-time framework that turns reference-free PCB schematic design into a verifiable optimization by coupling a schema-induced knowledge graph, a 5-layer continuous-reward verifier, and a Thompson Sampling refinement bandit. With GPT-5.4 it reaches $79.3\%$ on \textsc{Open-Schematics-Eval} ($70.7\%$ Hard) (\Cref{app:limitations,app:broader_impact}).
 
\bibliographystyle{plainnat}
\bibliography{references}

\begin{thebibliography}{97}
\providecommand{\natexlab}[1]{#1}
\providecommand{\url}[1]{\texttt{#1}}
\expandafter\ifx\csname urlstyle\endcsname\relax
  \providecommand{\doi}[1]{doi: #1}\else
  \providecommand{\doi}{doi: \begingroup \urlstyle{rm}\Url}\fi

\bibitem[Amano et~al.(2018)Amano, Baines, Beam, Borga, Bouchet, Chalker,
  Charles, Chen, Chowdhury, Chu, et~al.]{pe_gan2018}
Hiroshi Amano, Yannick Baines, E~Beam, Matteo Borga, Thierry Bouchet, Paul~R
  Chalker, Matthew Charles, Kevin~J Chen, Nadim Chowdhury, Rongming Chu, et~al.
\newblock The 2018 gan power electronics roadmap.
\newblock \emph{Journal of Physics D: Applied Physics}, 51\penalty0
  (16):\penalty0 163001, 2018.

\bibitem[Ataei et~al.(2026)Ataei, Askari, Malekshan, and
  Jayaraman]{zerotocad2026}
Mohammadmehdi Ataei, Farzaneh Askari, Kamal~Rahimi Malekshan, and Pradeep~Kumar
  Jayaraman.
\newblock Zero-to-cad: Agentic synthesis of interpretable cad programs at
  million-scale without real data.
\newblock \emph{arXiv preprint arXiv:2604.24479}, 2026.

\bibitem[Auer et~al.(2002)Auer, Cesa-Bianchi, and Fischer]{auer2002ucb}
Peter Auer, Nicolo Cesa-Bianchi, and Paul Fischer.
\newblock Finite-time analysis of the multiarmed bandit problem.
\newblock \emph{Machine learning}, 47\penalty0 (2):\penalty0 235--256, 2002.

\bibitem[Austin et~al.(2021)Austin, Odena, Nye, Bosma, Michalewski, Dohan,
  Jiang, Cai, Terry, Le, et~al.]{austin2021mbpp}
Jacob Austin, Augustus Odena, Maxwell Nye, Maarten Bosma, Henryk Michalewski,
  David Dohan, Ellen Jiang, Carrie Cai, Michael Terry, Quoc Le, et~al.
\newblock Program synthesis with large language models.
\newblock \emph{arXiv preprint arXiv:2108.07732}, 2021.

\bibitem[Bhandari et~al.(2024)Bhandari, Bhat, He, Rahmani, Garg, and
  Karri]{analog_masalachai2024}
Jitendra Bhandari, Vineet Bhat, Yuheng He, Hamed Rahmani, Siddharth Garg, and
  Ramesh Karri.
\newblock Masala-chai: A large-scale spice netlist dataset for analog circuits
  by harnessing ai.
\newblock \emph{arXiv preprint arXiv:2411.14299}, 2024.

\bibitem[Blocklove et~al.(2023)Blocklove, Garg, Karri, and
  Pearce]{rtl_chipchat2023}
Jason Blocklove, Siddharth Garg, Ramesh Karri, and Hammond Pearce.
\newblock Chip-chat: Challenges and opportunities in conversational hardware
  design.
\newblock In \emph{2023 ACM/IEEE 5th Workshop on Machine Learning for CAD
  (MLCAD)}, pages 1--6. IEEE, 2023.

\bibitem[bshada(2025)]{data_openschematics}
bshada.
\newblock Open schematics: A dataset of electronic schematics from hardware
  projects.
\newblock \url{https://huggingface.co/datasets/bshada/open-schematics}, 2025.

\bibitem[Chang et~al.(2025)Chang, Lin, Shen, Chen, and
  Zhang]{analog_lamagic22025}
Chen-Chia Chang, Wan-Hsuan Lin, Yikang Shen, Yiran Chen, and Xin Zhang.
\newblock Lamagic2: Advanced circuit formulations for language model-based
  analog topology generation.
\newblock \emph{arXiv preprint arXiv:2506.10235}, 2025.

\bibitem[Chen et~al.(2021)Chen, Tworek, Jun, Yuan, Pinto, Kaplan, Edwards,
  Burda, Joseph, Brockman, et~al.]{chen2021}
Mark Chen, Jerry Tworek, Heewoo Jun, Qiming Yuan, Henrique Ponde De~Oliveira
  Pinto, Jared Kaplan, Harri Edwards, Yuri Burda, Nicholas Joseph, Greg
  Brockman, et~al.
\newblock Evaluating large language models trained on code.
\newblock \emph{arXiv preprint arXiv:2107.03374}, 2021.

\bibitem[Chen et~al.(2023)Chen, Lin, Sch{\"a}rli, and Zhou]{tool_selfdebug2023}
Xinyun Chen, Maxwell Lin, Nathanael Sch{\"a}rli, and Denny Zhou.
\newblock Teaching large language models to self-debug.
\newblock \emph{arXiv preprint arXiv:2304.05128}, 2023.

\bibitem[Chen et~al.(2024)Chen, Huang, Liu, Yang, Shang, Zhou, and
  Zeng]{analog_artisan2024}
Zihao Chen, Jiangli Huang, Yiting Liu, Fan Yang, Li~Shang, Dian Zhou, and Xuan
  Zeng.
\newblock Artisan: Automated operational amplifier design via domain-specific
  large language model.
\newblock In \emph{Proceedings of the 61st ACM/IEEE Design Automation
  Conference}, pages 1--6, 2024.

\bibitem[Cohen(1960)]{tool_coefficient}
Jacob Cohen.
\newblock A coefficient of agreement for nominal scales.
\newblock \emph{Educational and psychological measurement}, 20\penalty0
  (1):\penalty0 37--46, 1960.

\bibitem[Cordella et~al.(2004)Cordella, Foggia, Sansone, and Vento]{tools_sub}
Luigi~P Cordella, Pasquale Foggia, Carlo Sansone, and Mario Vento.
\newblock A (sub) graph isomorphism algorithm for matching large graphs.
\newblock \emph{IEEE transactions on pattern analysis and machine
  intelligence}, 26\penalty0 (10):\penalty0 1367--1372, 2004.

\bibitem[Daniel et~al.(2018)Daniel, Benjamin, Abbas, Ian, and
  Zheng]{russo2018tutorial}
J~Russo Daniel, Van~Roy Benjamin, Kazerouni Abbas, Osband Ian, and Wen Zheng.
\newblock A tutorial on thompson sampling.
\newblock \emph{Foundations and Trends{\textregistered} in Machine Learning},
  11\penalty0 (1):\penalty0 1--99, 2018.

\bibitem[Dong et~al.(2023)Dong, Cao, Zhang, Tao, Chen, and
  Zhang]{ml_cktgnn2023}
Zehao Dong, Weidong Cao, Muhan Zhang, Dacheng Tao, Yixin Chen, and Xuan Zhang.
\newblock Cktgnn: Circuit graph neural network for electronic design
  automation.
\newblock \emph{arXiv preprint arXiv:2308.16406}, 2023.

\bibitem[Eland et~al.(2026)Eland, Thiyagalingam, Pamunuwa, and
  Weerasekera]{eland2026nl2gds}
Max Eland, Jeyan Thiyagalingam, Dinesh Pamunuwa, and Roshan Weerasekera.
\newblock Nl2gds: Llm-aided interface for open source chip design.
\newblock \emph{arXiv preprint arXiv:2603.05489}, 2026.

\bibitem[Fu et~al.(2023)Fu, Zhang, Yu, Li, Ye, Li, Wan, and
  Lin]{rtl_gpt4aigchip2023}
Yonggan Fu, Yongan Zhang, Zhongzhi Yu, Sixu Li, Zhifan Ye, Chaojian Li, Cheng
  Wan, and Yingyan~Celine Lin.
\newblock Gpt4aigchip: Towards next-generation ai accelerator design automation
  via large language models.
\newblock In \emph{2023 IEEE/ACM International Conference on Computer Aided
  Design (ICCAD)}, pages 1--9. IEEE, 2023.

\bibitem[Gao et~al.(2025{\natexlab{a}})Gao, Cao, Yang, and
  Zhang]{analog_analoggenie2025}
Jian Gao, Weidong Cao, Junyi Yang, and Xuan Zhang.
\newblock Analoggenie: A generative engine for automatic discovery of analog
  circuit topologies.
\newblock \emph{arXiv preprint arXiv:2503.00205}, 2025{\natexlab{a}}.

\bibitem[Gao et~al.(2025{\natexlab{b}})Gao, Cao, and
  Zhang]{analog_analoggenielite2025}
Jian Gao, Weidong Cao, and Xuan Zhang.
\newblock Analoggenie-lite: Enhancing scalability and precision in circuit
  topology discovery through lightweight graph modeling.
\newblock In \emph{Forty-second International Conference on Machine Learning},
  2025{\natexlab{b}}.

\bibitem[Gehring et~al.(2024)Gehring, Zheng, Copet, Mella, Carbonneaux, Cohen,
  and Synnaeve]{rlef2025}
Jonas Gehring, Kunhao Zheng, Jade Copet, Vegard Mella, Quentin Carbonneaux,
  Taco Cohen, and Gabriel Synnaeve.
\newblock Rlef: Grounding code llms in execution feedback with reinforcement
  learning.
\newblock \emph{arXiv preprint arXiv:2410.02089}, 2024.

\bibitem[{Google DeepMind}(2025)]{llm_gemma4_31b_2025}
{Google DeepMind}.
\newblock {Gemma 4 31B}: A dense open multimodal model.
\newblock \url{https://huggingface.co/google/gemma-4-31B}, 2025.

\bibitem[{Google DeepMind}(2026{\natexlab{a}})]{llm_gemini31flashlite_2026}
{Google DeepMind}.
\newblock Gemini 3.1 flash-lite -- model card.
\newblock
  \url{https://deepmind.google/models/model-cards/gemini-3-1-flash-lite/},
  2026{\natexlab{a}}.
\newblock Published 2026-03-03; accessed 2026-05-05.

\bibitem[{Google DeepMind}(2026{\natexlab{b}})]{llm_gemini31pro_2026}
{Google DeepMind}.
\newblock Gemini 3.1 pro -- model card.
\newblock \url{https://deepmind.google/models/model-cards/gemini-3-1-pro/},
  2026{\natexlab{b}}.
\newblock Published 2026-02-19; accessed 2026-05-05.

\bibitem[Hasan et~al.(2026)Hasan, Raiyan, Alvee, and
  Sadik]{compare_circuitlm2026}
Khandakar Shakib~Al Hasan, Syed~Rifat Raiyan, Hasin~Mahtab Alvee, and Wahid
  Sadik.
\newblock Circuitlm: A multi-agent llm-aided design framework for generating
  circuit schematics from natural language prompts, 2026.
\newblock URL \url{https://arxiv.org/abs/2601.04505}.

\bibitem[Hong et~al.(2023)Hong, Zhuge, Chen, Zheng, Cheng, Wang, Zhang, Wang,
  Yau, Lin, et~al.]{tool_metagpt2023}
Sirui Hong, Mingchen Zhuge, Jonathan Chen, Xiawu Zheng, Yuheng Cheng, Jinlin
  Wang, Ceyao Zhang, Zili Wang, Steven Ka~Shing Yau, Zijuan Lin, et~al.
\newblock Metagpt: Meta programming for a multi-agent collaborative framework.
\newblock In \emph{The twelfth international conference on learning
  representations}, 2023.

\bibitem[Huang et~al.(2010)Huang, Crow, Heydt, Zheng, and Dale]{pe_freedm2011}
Alex~Q Huang, Mariesa~L Crow, Gerald~Thomas Heydt, Jim~P Zheng, and Steiner~J
  Dale.
\newblock The future renewable electric energy delivery and management (freedm)
  system: the energy internet.
\newblock \emph{Proceedings of the IEEE}, 99\penalty0 (1):\penalty0 133--148,
  2010.

\bibitem[Huang et~al.()Huang, Chang, Lin, and Yang]{beyondoracle2025}
Sian-Yao Huang, Li-Hsien Chang, Che-Yu Lin, and Cheng-Lin Yang.
\newblock Beyond oracle: Verifier-supervision for instruction hierarchy in
  reasoning and instruction-tuned llms.
\newblock In \emph{The Thirty-ninth Annual Conference on Neural Information
  Processing Systems}.

\bibitem[{InternLM Team}(2025)]{llm_internlm3_8b_2025}
{InternLM Team}.
\newblock {InternLM3-8B-Instruct}.
\newblock
  \url{https://internlm.readthedocs.io/en/latest/model_card/InternLM3.html},
  2025.
\newblock Released 2025-01-15; accessed 2026-05-06.

\bibitem[Jha et~al.(2024)Jha, Jha, Rashed, Ewetz, and
  Velasquez]{compare_Jha2024}
Sumit~Kumar Jha, Susmit Jha, Muhammad Rashedul~Haq Rashed, Rickard Ewetz, and
  Alvaro Velasquez.
\newblock Automated synthesis of hardware designs using symbolic feedback and
  grammar-constrained decoding in large language models.
\newblock In \emph{NAECON 2024-IEEE National Aerospace and Electronics
  Conference}, pages 95--100. IEEE, 2024.

\bibitem[Jimenez et~al.(2023)Jimenez, Yang, Wettig, Yao, Pei, Press, and
  Narasimhan]{jimenez2024swebench}
Carlos~E Jimenez, John Yang, Alexander Wettig, Shunyu Yao, Kexin Pei, Ofir
  Press, and Karthik~R Narasimhan.
\newblock Swe-bench: Can language models resolve real-world github issues?
\newblock In \emph{The twelfth international conference on learning
  representations}, 2023.

\bibitem[Kumar et~al.(2024)Kumar, Zhuang, Agarwal, Su, Co-Reyes, Singh, Baumli,
  Iqbal, Bishop, Roelofs, et~al.]{kumar2024score}
Aviral Kumar, Vincent Zhuang, Rishabh Agarwal, Yi~Su, John~D Co-Reyes, Avi
  Singh, Kate Baumli, Shariq Iqbal, Colton Bishop, Rebecca Roelofs, et~al.
\newblock Training language models to self-correct via reinforcement learning.
\newblock \emph{arXiv preprint arXiv:2409.12917}, 2024.

\bibitem[Lai et~al.(2023)Lai, Liu, Tang, Wang, Hao, and Luo]{ml_chipformer2023}
Yao Lai, Jinxin Liu, Zhentao Tang, Bin Wang, Jianye Hao, and Ping Luo.
\newblock Chipformer: Transferable chip placement via offline decision
  transformer, 2023.

\bibitem[Lai et~al.(2025)Lai, Lee, Chen, Poddar, Hu, Pan, and
  Luo]{analog_analogcoder2025}
Yao Lai, Sungyoung Lee, Guojin Chen, Souradip Poddar, Mengkang Hu, David~Z Pan,
  and Ping Luo.
\newblock Analogcoder: Analog circuit design via training-free code generation.
\newblock In \emph{Proceedings of the AAAI Conference on Artificial
  Intelligence}, volume~39, pages 379--387, 2025.

\bibitem[Landis and Koch(1977)]{landis1977measurement}
J~Richard Landis and Gary~G Koch.
\newblock The measurement of observer agreement for categorical data.
\newblock \emph{biometrics}, pages 159--174, 1977.

\bibitem[Li et~al.(2023)Li, Zhang, Xu, and Liu]{ml_fanoutnet2023}
Haiyun Li, Jixin Zhang, Ning Xu, and Mingyu Liu.
\newblock Fanoutnet: A neuralized pcb fanout automation method using deep
  reinforcement learning.
\newblock In \emph{Proceedings of the AAAI Conference on Artificial
  Intelligence}, volume~37, pages 8554--8561, 2023.

\bibitem[Li et~al.(2026)Li, Chen, Yang, Zhu, Wang, Ma, and Yang]{pcbbench_li}
Jindong Li, Lianrong Chen, Bin Yang, Jiadong Zhu, Ying Wang, Yuzhe Ma, and
  Menglin Yang.
\newblock Pcb-bench: Benchmarking llms for printed circuit board placement and
  routing.
\newblock In \emph{The Fourteenth International Conference on Learning
  Representations}, 2026.

\bibitem[Lin et~al.(2024)Lin, Liu, Li, Zhao, Zhao, Liao, Ma, and
  Zhang]{analog_pegpt2025}
Fanfan Lin, Junhua Liu, Xinze Li, Shuai Zhao, Bohui Zhao, Xinyuan Liao, Hao Ma,
  and Xin Zhang.
\newblock Pe-gpt: A physics-informed interactive large language model for power
  converter modulation design.
\newblock In \emph{2024 IEEE Energy Conversion Congress and Exposition (ECCE)},
  pages 1744--1747. IEEE, 2024.

\bibitem[Liu et~al.(2025{\natexlab{a}})Liu, Mei, Lin, Xue, Wang, Xu, Wu, Zhang,
  Lin, Dong, et~al.]{llm_deepseekv32_2025}
Aixin Liu, Aoxue Mei, Bangcai Lin, Bing Xue, Bingxuan Wang, Bingzheng Xu,
  Bochao Wu, Bowei Zhang, Chaofan Lin, Chen Dong, et~al.
\newblock Deepseek-v3. 2: Pushing the frontier of open large language models.
\newblock \emph{arXiv preprint arXiv:2512.02556}, 2025{\natexlab{a}}.

\bibitem[Liu et~al.(2024{\natexlab{a}})Liu, Zhang, Gao, Kong, Tang, Lin, Wang,
  and Huang]{pcb_layoutcopilot2024}
Bingyang Liu, Haoyi Zhang, Xiaohan Gao, Zichen Kong, Xiyuan Tang, Yibo Lin,
  Runsheng Wang, and Ru~Huang.
\newblock Layoutcopilot: An llm-powered multi-agent collaborative framework for
  interactive analog layout design, 2024{\natexlab{a}}.

\bibitem[Liu and Chitnis(2025)]{compare_eeschematic2025}
Chang Liu and Danial Chitnis.
\newblock Eeschematic: Multimodal-llm based ai agent for schematic generation
  of analog circuit.
\newblock \emph{arXiv preprint arXiv:2510.17002}, 2025.

\bibitem[Liu et~al.(2024{\natexlab{b}})Liu, Chen, Peng, Du, Du, and
  Yang]{analog_ampagent2024}
Chengjie Liu, Weiyu Chen, Anlan Peng, Yuan Du, Li~Du, and Jun Yang.
\newblock Ampagent: An llm-based multi-agent system for multi-stage amplifier
  schematic design from literature for process and performance porting.
\newblock \emph{arXiv preprint arXiv:2409.14739}, 2024{\natexlab{b}}.

\bibitem[Liu et~al.(2024{\natexlab{c}})Liu, Xue, Chen, Chen, Zhao, Wang, Hou,
  Li, and Peng]{llm_vllm}
Hanchao Liu, Wenyuan Xue, Yifei Chen, Dapeng Chen, Xiutian Zhao, Ke~Wang,
  Liping Hou, Rongjun Li, and Wei Peng.
\newblock A survey on hallucination in large vision-language models.
\newblock \emph{arXiv preprint arXiv:2402.00253}, 2024{\natexlab{c}}.

\bibitem[Liu et~al.(2026)Liu, Lu, Wang, Yao, and Yu]{liu2026llm}
Hongduo Liu, Yuntao Lu, Mingjun Wang, Xufeng Yao, and Bei Yu.
\newblock Llm-assisted circuit verification: A comprehensive survey.
\newblock In \emph{2026 31st Asia and South Pacific Design Automation
  Conference (ASP-DAC)}, pages 439--446. IEEE, 2026.

\bibitem[Liu et~al.(2023{\natexlab{a}})Liu, Ene, Kirby,
  et~al.]{rtl_chipnemo2023}
Mingjie Liu, Teodor-Dumitru Ene, Robert Kirby, et~al.
\newblock Chipnemo: Domain-adapted llms for chip design, 2023{\natexlab{a}}.

\bibitem[Liu et~al.(2023{\natexlab{b}})Liu, Pinckney, Khailany, and
  Ren]{bench_verilogeval2023}
Mingjie Liu, Nathaniel Pinckney, Brucek Khailany, and Haoxing Ren.
\newblock Verilogeval: Evaluating large language models for verilog code
  generation, 2023{\natexlab{b}}.

\bibitem[Liu et~al.(2025{\natexlab{b}})Liu, Xu, Zhou, Li, and
  Xu]{rtl_deeprtl2025}
Yi~Liu, Changran Xu, Yunhao Zhou, Zeju Li, and Qiang Xu.
\newblock Deeprtl: Bridging verilog understanding and generation with a unified
  representation model.
\newblock \emph{arXiv preprint arXiv:2502.15832}, 2025{\natexlab{b}}.

\bibitem[Lu et~al.(2026)Lu, Lin, Tian, Wang, Wang, Khatri, Kartik, Wang, Rane,
  Wang, et~al.]{lu2026omnisch}
Taiting Lu, Kaiyuan Lin, Yuxin Tian, Yubo Wang, Muchuan Wang, Sharique Khatri,
  Akshit Kartik, Yixi Wang, Amey~Santosh Rane, Yida Wang, et~al.
\newblock Omnisch: A multimodal pcb schematic benchmark for structured diagram
  visual reasoning.
\newblock \emph{arXiv preprint arXiv:2604.00270}, 2026.

\bibitem[Lu et~al.(2024)Lu, Liu, Zhang, and Xie]{lu2024rtllm}
Yao Lu, Shang Liu, Qijun Zhang, and Zhiyao Xie.
\newblock Rtllm: An open-source benchmark for design rtl generation with large
  language model.
\newblock In \emph{2024 29th Asia and South Pacific Design Automation
  Conference (ASP-DAC)}, pages 722--727. IEEE, 2024.

\bibitem[Luo et~al.(2025)Luo, Ma, Zhang, and Qiu]{schgen_anonymous2026}
Qinpei Luo, Ruichun Ma, Xinyu Zhang, and Lili Qiu.
\newblock {SchGen}: {PCB} schematic generation with semantic-grounded code
  representations.
\newblock \url{https://openreview.net/forum?id=TyWs6rWWHb}, 2025.
\newblock OpenReview manuscript; ICLR 2026 desk-rejected submission; first
  posted 2025-09-19, modified 2026-02-12; accessed 2026-05-06.

\bibitem[Ma et~al.(2025)Ma, Shao, Li, Song, Guo, Li, Qiu, and
  Chen]{unitcoder2025}
Yichuan Ma, Yunfan Shao, Peiji Li, Demin Song, Qipeng Guo, Linyang Li, Xipeng
  Qiu, and Kai Chen.
\newblock Unitcoder: Scalable iterative code synthesis with unit test guidance.
\newblock \emph{arXiv preprint arXiv:2502.11460}, 2025.

\bibitem[Madaan et~al.(2023)Madaan, Tandon, Gupta, Hallinan, Gao, Wiegreffe,
  Alon, Dziri, Prabhumoye, Yang, et~al.]{madaan2023selfrefine}
Aman Madaan, Niket Tandon, Prakhar Gupta, Skyler Hallinan, Luyu Gao, Sarah
  Wiegreffe, Uri Alon, Nouha Dziri, Shrimai Prabhumoye, Yiming Yang, et~al.
\newblock Self-refine: Iterative refinement with self-feedback.
\newblock \emph{Advances in neural information processing systems},
  36:\penalty0 46534--46594, 2023.

\bibitem[Matsuo et~al.(2024)Matsuo, Uhlich, Venkitaraman, Bonetti, Hsieh,
  Momeni, Mauch, Capone, Ohbuchi, and Servadei]{pcb_schemato2024}
Ryoga Matsuo, Stefan Uhlich, Arun Venkitaraman, Andrea Bonetti, Chia-Yu Hsieh,
  Ali Momeni, Lukas Mauch, Augusto Capone, Eisaku Ohbuchi, and Lorenzo
  Servadei.
\newblock Schemato -- an llm for netlist-to-schematic conversion, 2024.

\bibitem[Matsuo et~al.(2025)Matsuo, Uhlich, Venkitaraman, Bonetti, Hsieh,
  Momeni, Mauch, Capone, Ohbuchi, and Servadei]{matsuo2025schemato}
Ryoga Matsuo, Stefan Uhlich, Arun Venkitaraman, Andrea Bonetti, Chia-Yu Hsieh,
  Ali Momeni, Lukas Mauch, Augusto Capone, Eisaku Ohbuchi, and Lorenzo
  Servadei.
\newblock Schemato--an llm for netlist-to-schematic conversion.
\newblock In \emph{2025 ACM/IEEE 7th Symposium on Machine Learning for CAD
  (MLCAD)}, pages 1--7. IEEE, 2025.

\bibitem[{Meta AI}(2025)]{llm_llama4scout_2025}
{Meta AI}.
\newblock {Llama 4 Scout}: A 17b-active-parameter multimodal mixture-of-experts
  model.
\newblock \url{https://huggingface.co/meta-llama/Llama-4-Scout-17B-16E}, 2025.

\bibitem[{MiniMax}(2026)]{llm_minimaxm25_2026}
{MiniMax}.
\newblock {MiniMax M2.5}: Built for real-world productivity.
\newblock \url{https://www.minimax.io/news/minimax-m25}, 2026.
\newblock Released 2026-02-12; accessed 2026-05-05.

\bibitem[Mirhoseini et~al.(2021)Mirhoseini, Goldie, Yazgan,
  et~al.]{ml_graphplacement2021}
Azalia Mirhoseini, Anna Goldie, Mustafa Yazgan, et~al.
\newblock A graph placement methodology for fast chip design.
\newblock \emph{Nature}, 594\penalty0 (7862):\penalty0 207--212, June 2021.
\newblock ISSN 0028-0836, 1476-4687.
\newblock \doi{10.1038/s41586-021-03544-w}.

\bibitem[{Mistral AI}(2025)]{llm_devstral24b_2025}
{Mistral AI}.
\newblock {Devstral Small 2 24B Instruct 2512}.
\newblock
  \url{https://huggingface.co/mistralai/Devstral-Small-2-24B-Instruct-2512},
  2025.
\newblock Released 2025-12-09; 24B parameters; 256K context; Apache 2.0;
  accessed 2026-05-06.

\bibitem[Mo et~al.(2025)Mo, Yu, Kazdan, Cabezas, Mpala, Yu, Cundy, Kanatsoulis,
  and Koyejo]{mo2025kggen}
Belinda Mo, Kyssen Yu, Joshua Kazdan, Joan Cabezas, Proud Mpala, Lisa Yu, Chris
  Cundy, Charilaos Kanatsoulis, and Sanmi Koyejo.
\newblock Kggen: Extracting knowledge graphs from plain text with language
  models.
\newblock \emph{arXiv preprint arXiv:2502.09956}, 2025.

\bibitem[Nagel and Pederson(1973)]{spice_nagel1973}
Laurence Nagel and Donald~O Pederson.
\newblock Spice (simulation program with integrated circuit emphasis).
\newblock 1973.

\bibitem[Ng et~al.(1999)Ng, Harada, and Russell]{ng1999reward}
Andrew~Y Ng, Daishi Harada, and Stuart Russell.
\newblock Policy invariance under reward transformations: Theory and
  application to reward shaping.
\newblock In \emph{Icml}, volume~99, pages 278--287. Citeseer, 1999.

\bibitem[Olausson et~al.(2023)Olausson, Inala, Wang, Gao, and
  {Solar-Lezama}]{tool_selfrepair2023}
Theo~X. Olausson, Jeevana~Priya Inala, Chenglong Wang, Jianfeng Gao, and
  Armando {Solar-Lezama}.
\newblock Is self-repair a silver bullet for code generation?, 2023.

\bibitem[{OpenAI}(2026)]{llm_gpt54_2026}
{OpenAI}.
\newblock {GPT-5.4} model.
\newblock \url{https://developers.openai.com/api/docs/models/gpt-5.4}, 2026.
\newblock Released 2026-03-05; accessed 2026-05-05.

\bibitem[Pan et~al.(2025)Pan, Jacobson, Zhao, Chen, Zhou, Chang, Rashingkar,
  and Chen]{compare_CROP2025}
Jingyu Pan, Isaac Jacobson, Zheng Zhao, Tung-Chieh Chen, Guanglei Zhou,
  Chen-Chia Chang, Vineet Rashingkar, and Yiran Chen.
\newblock Crop: Circuit retrieval and optimization with parameter guidance
  using llms.
\newblock In \emph{2025 IEEE/ACM International Conference On Computer Aided
  Design (ICCAD)}, pages 1--9. IEEE, 2025.

\bibitem[Pecht(2009)]{pcb_pecht2009prognostics}
Michael Pecht.
\newblock Prognostics and health management of electronics.
\newblock \emph{Encyclopedia of structural health monitoring}, 2009.

\bibitem[Pei et~al.(2024)Pei, Zhen, Yuan, Huang, and Yu]{rtl_betterv2024}
Zehua Pei, Hui-Ling Zhen, Mingxuan Yuan, Yu~Huang, and Bei Yu.
\newblock Betterv: Controlled verilog generation with discriminative guidance.
\newblock \emph{arXiv preprint arXiv:2402.03375}, 2024.

\bibitem[Plettenberg et~al.(2025)Plettenberg, Alcalde, Sick, and
  Thomas]{compare_GNN}
Pascal Plettenberg, Andr{\'e} Alcalde, Bernhard Sick, and Josephine~M Thomas.
\newblock Graph neural networks for automatic addition of optimizing components
  in printed circuit board schematics.
\newblock In \emph{Joint European Conference on Machine Learning and Knowledge
  Discovery in Databases}, pages 508--524. Springer, 2025.

\bibitem[{Qwen Team}(2026)]{llm_qwen35_27b_2026.5}
{Qwen Team}.
\newblock {Qwen3.5}: Towards native multimodal agents, February 2026.
\newblock URL \url{https://qwen.ai/blog?id=qwen3.5}.

\bibitem[Sestito et~al.(2026)Sestito, Kontou, Verma, Dixit, Keros, O'Boyle,
  Bouganis, and Prodromakis]{sestito2026flexible}
Cristian Sestito, Panagiota Kontou, Pratibha Verma, Atish Dixit, Alexandros~D
  Keros, Michael O'Boyle, Christos-Savvas Bouganis, and Themis Prodromakis.
\newblock A flexible language model-assisted electronic design automation
  framework.
\newblock \emph{arXiv preprint arXiv:2601.14098}, 2026.

\bibitem[Sethi()]{circuitron2025}
Shaurya Sethi.
\newblock {Circuitron}: Agentic pcb design accelerator.
\newblock \url{https://github.com/Shaurya-Sethi/circuitron}.
\newblock GitHub repository; accessed 2026-05-06.

\bibitem[Setlur et~al.(2024)Setlur, Nagpal, Fisch, Geng, Eisenstein, Agarwal,
  Agarwal, Berant, and Kumar]{setlur2024pav}
Amrith Setlur, Chirag Nagpal, Adam Fisch, Xinyang Geng, Jacob Eisenstein,
  Rishabh Agarwal, Alekh Agarwal, Jonathan Berant, and Aviral Kumar.
\newblock Rewarding progress: Scaling automated process verifiers for llm
  reasoning.
\newblock \emph{arXiv preprint arXiv:2410.08146}, 2024.

\bibitem[Shi et~al.(2025)Shi, Tao, Gao, Zhou, Chang, Wang, Chen, Zhang, Liu,
  Yu, Lin, and He]{analog_amsnetkg2025}
Yichen Shi, Zhuofu Tao, YuHao Gao, Tianjia Zhou, Cheng Chang, Yaxin Wang,
  Bingyu Chen, Genhao Zhang, Alvin Liu, Zhiping Yu, Ting-Jung Lin, and Lei He.
\newblock Amsnet-kg: A netlist dataset for llm-based ams circuit auto-design
  using knowledge graph rag.
\newblock \emph{ACM Transactions on Design Automation of Electronic Systems},
  30\penalty0 (6):\penalty0 1--37, November 2025.
\newblock ISSN 1084-4309, 1557-7309.
\newblock \doi{10.1145/3736166}.

\bibitem[Shinn et~al.(2023)Shinn, Cassano, Gopinath, Narasimhan, and
  Yao]{shinn2023reflexion}
Noah Shinn, Federico Cassano, Ashwin Gopinath, Karthik Narasimhan, and Shunyu
  Yao.
\newblock Reflexion: Language agents with verbal reinforcement learning.
\newblock \emph{Advances in neural information processing systems},
  36:\penalty0 8634--8652, 2023.

\bibitem[Shipra and Navin(2013)]{agrawal2013further}
Agrawal Shipra and Goyal Navin.
\newblock Further optimal regret bounds for thompson sampling.
\newblock In \emph{Artificial Intelligence and Statistics}, volume~31, pages
  99--107, 2013.

\bibitem[Sutton et~al.(1998)Sutton, Barto, et~al.]{sutton2018rl}
Richard~S Sutton, Andrew~G Barto, et~al.
\newblock \emph{Reinforcement learning: An introduction}, volume~1.
\newblock MIT press Cambridge, 1998.

\bibitem[Tang et~al.(2024)Tang, Hu, Zhou, Zhong, Zheng, Si, and
  Ellis]{tang2024rex}
Hao Tang, Keya Hu, Jin~P Zhou, Sicheng Zhong, Wei-Long Zheng, Xujie Si, and
  Kevin Ellis.
\newblock Code repair with llms gives an exploration-exploitation tradeoff.
\newblock \emph{Advances in Neural Information Processing Systems},
  37:\penalty0 117954--117996, 2024.

\bibitem[Team(2025)]{llm_qwen3coder30b_2025}
Qwen Team.
\newblock Qwen3 technical report, 2025.
\newblock URL \url{https://arxiv.org/abs/2505.09388}.

\bibitem[Thakur et~al.(2023)Thakur, Blocklove, Pearce, Tan, Garg, and
  Karri]{rtl_autochip2023}
Shailja Thakur, Jason Blocklove, Hammond Pearce, Benjamin Tan, Siddharth Garg,
  and Ramesh Karri.
\newblock Autochip: Automating hdl generation using llm feedback.
\newblock \emph{arXiv preprint arXiv:2311.04887}, 2023.

\bibitem[Thakur et~al.(2024)Thakur, Ahmad, Pearce, Tan, Dolan-Gavitt, Karri,
  and Garg]{rtl_verigen2024}
Shailja Thakur, Baleegh Ahmad, Hammond Pearce, Benjamin Tan, Brendan
  Dolan-Gavitt, Ramesh Karri, and Siddharth Garg.
\newblock Verigen: A large language model for verilog code generation.
\newblock \emph{ACM Transactions on Design Automation of Electronic Systems},
  29\penalty0 (3):\penalty0 1--31, 2024.

\bibitem[Thompson(1933)]{thompson1933}
William~R Thompson.
\newblock On the likelihood that one unknown probability exceeds another in
  view of the evidence of two samples.
\newblock \emph{Biometrika}, 25\penalty0 (3/4):\penalty0 285--294, 1933.

\bibitem[Vandenbout(2025)]{tool_skidl}
Dave Vandenbout.
\newblock Skidl: A python package for textually describing electronic circuit
  schematics, December 2025.
\newblock URL \url{https://pypi.org/project/skidl/}.
\newblock PyPI project page and release history.

\bibitem[Vijayaraghavan et~al.(2025{\natexlab{a}})Vijayaraghavan, Shi, Degan,
  and Mukherjee]{compare_CIRCUITSYNTH}
Prashanth Vijayaraghavan, Luyao Shi, Ehsan Degan, and Vandana Mukherjee.
\newblock Circuitsynth-rl: Llm-based circuit topology synthesis with rl
  refinement.
\newblock In \emph{ACM/IEEE Design Automation Conference}, 2025{\natexlab{a}}.

\bibitem[Vijayaraghavan et~al.(2025{\natexlab{b}})Vijayaraghavan, Shi, Degan,
  Mukherjee, and Zhang]{autocircuitrl2025}
Prashanth Vijayaraghavan, Luyao Shi, Ehsan Degan, Vandana Mukherjee, and Xin
  Zhang.
\newblock Autocircuit-rl: Reinforcement learning-driven llm for automated
  circuit topology generation.
\newblock \emph{arXiv preprint arXiv:2506.03122}, 2025{\natexlab{b}}.

\bibitem[Wang et~al.(2024)Wang, Li, Shao, Xu, Dai, Li, Chen, Wu, and
  Sui]{wang2024mathshepherd}
Peiyi Wang, Lei Li, Zhihong Shao, Runxin Xu, Damai Dai, Yifei Li, Deli Chen,
  Yu~Wu, and Zhifang Sui.
\newblock Math-shepherd: Verify and reinforce llms step-by-step without human
  annotations.
\newblock In \emph{Proceedings of the 62nd Annual Meeting of the Association
  for Computational Linguistics (Volume 1: Long Papers)}, pages 9426--9439,
  2024.

\bibitem[Wang et~al.(2025)Wang, Lu, Liu, Yang, Yang, Chen, Wang, Liu, Lin,
  Chen, et~al.]{compare_LLMfootprint}
Yida Wang, Taiting Lu, Runze Liu, Lanqing Yang, Yifan Yang, Zhe Chen, Yuehai
  Wang, Yixin Liu, Kaiyuan Lin, Xiaomeng Chen, et~al.
\newblock A large language model powered integrated circuit footprint geometry
  understanding.
\newblock \emph{arXiv preprint arXiv:2508.03725}, 2025.

\bibitem[Wei et~al.(2025)Wei, Tan, Suresh, Mendoza, Teixeira, Wang, Trippel,
  and Aiken]{vericoder2025}
Anjiang Wei, Huanmi Tan, Tarun Suresh, Daniel Mendoza, Thiago~SFX Teixeira,
  Ke~Wang, Caroline Trippel, and Alex Aiken.
\newblock Vericoder: Enhancing llm-based rtl code generation through functional
  correctness validation.
\newblock \emph{arXiv preprint arXiv:2504.15659}, 2025.

\bibitem[Wei et~al.(2026)Wei, Kong, Wang, Pan, and Tang]{analog_toposizing2025}
Ziming Wei, Zichen Kong, Yuan Wang, David~Z Pan, and Xiyuan Tang.
\newblock Toposizing: An llm-aided framework of topology-based understanding
  and sizing for ams circuits.
\newblock \emph{IEEE Transactions on Computer-Aided Design of Integrated
  Circuits and Systems}, 2026.

\bibitem[Wilson(1927)]{wilson1927probable}
Edwin~B Wilson.
\newblock Probable inference, the law of succession, and statistical inference.
\newblock \emph{Journal of the American Statistical Association}, 22\penalty0
  (158):\penalty0 209--212, 1927.

\bibitem[Xiao et~al.(2024)Xiao, Putrevu, Hemadri, Garg, and
  Karri]{xiao2024prefixllm}
Weihua Xiao, Venkata Sai~Charan Putrevu, Raghu~Vamshi Hemadri, Siddharth Garg,
  and Ramesh Karri.
\newblock Prefixllm: Llm-aided prefix circuit design.
\newblock \emph{arXiv preprint arXiv:2412.02594}, 2024.

\bibitem[Yan et~al.(2022)Yan, Norman, Lim, Norman, Ho, Zhu, and
  MA]{compare_yan2022ai}
Jin Yan, Adam Norman, Min~Suet Lim, Mackenzie Norman, Hong~Cheah Ho, Jianfang
  Zhu, and Miaomiao MA.
\newblock Ai-based floorplanning for printed circuit board design, September~22
  2022.
\newblock US Patent App. 17/835,323.

\bibitem[Yin et~al.(2024)Yin, Wang, Xu, and Li]{analog_adollm2024}
Yuxuan Yin, Yu~Wang, Boxun Xu, and Peng Li.
\newblock Ado-llm: Analog design bayesian optimization with in-context learning
  of large language models.
\newblock In \emph{Proceedings of the 43rd IEEE/ACM International Conference on
  Computer-Aided Design}, pages 1--9, 2024.

\bibitem[Yu et~al.(2024)Yu, Tao, Chen, Sun, and Yang]{bcoder2024}
Zishun Yu, Yunzhe Tao, Liyu Chen, Tao Sun, and Hongxia Yang.
\newblock B-coder: Value-based deep reinforcement learning for program
  synthesis.
\newblock In \emph{12th International Conference on Learning Representations,
  ICLR 2024}, 2024.

\bibitem[Zhang and Soh(2024)]{zhang2024edc}
Bowen Zhang and Harold Soh.
\newblock Extract, define, canonicalize: An llm-based framework for knowledge
  graph construction.
\newblock In \emph{Proceedings of the 2024 conference on empirical methods in
  natural language processing}, pages 9820--9836, 2024.

\bibitem[Zhang et~al.(2025{\natexlab{a}})Zhang, Zhang, Li, Yang, and
  Chu]{pcb_pcbrouting2025}
Kangkang Zhang, Huailong Zhang, Aobo Li, Zhiping Yang, and Xiuqin Chu.
\newblock Applying existing large language models for print circuit board
  routing.
\newblock In \emph{ICSEE 2024}, page~2. MDPI, February 2025{\natexlab{a}}.
\newblock \doi{10.3390/engproc2025086002}.

\bibitem[Zhang et~al.(2023)Zhang, Wang, Xia, Wang, and Li]{zhang2023algo}
Kexun Zhang, Danqing Wang, Jingtao Xia, William~Yang Wang, and Lei Li.
\newblock Algo: Synthesizing algorithmic programs with generated oracle
  verifiers.
\newblock \emph{Advances in Neural Information Processing Systems},
  36:\penalty0 54769--54784, 2023.

\bibitem[Zhang et~al.(2026)Zhang, Qin, Cao, Xue, and Xie]{funprm2026}
Ruiyi Zhang, Peijia Qin, Qi~Cao, Eric Xue, and Pengtao Xie.
\newblock Funprm: Function-as-step process reward model with meta reward
  correction for code generation.
\newblock \emph{arXiv preprint arXiv:2601.22249}, 2026.

\bibitem[Zhang et~al.(2025{\natexlab{b}})Zhang, Zhang, Guo, Huang, Huang, Zhao,
  Cheng, Jin, Li, Du, et~al.]{rtl_qimengsalv2025}
Yang Zhang, Rui Zhang, Jiaming Guo, Lei Huang, Di~Huang, Yunpu Zhao, Shuyao
  Cheng, Pengwei Jin, Chongxiao Li, Zidong Du, et~al.
\newblock Qimeng-salv: Signal-aware learning for verilog code generation.
\newblock \emph{arXiv preprint arXiv:2510.19296}, 2025{\natexlab{b}}.

\bibitem[Zhu et~al.(2025)Zhu, Huang, Lyu, Zhang, Li, Shi, Wu, Mu, Wang, Zhao,
  et~al.]{rtl_qimengcodevr12025}
Yaoyu Zhu, Di~Huang, Hanqi Lyu, Xiaoyun Zhang, Chongxiao Li, Wenxuan Shi,
  Yutong Wu, Jianan Mu, Jinghua Wang, Yang Zhao, et~al.
\newblock Qimeng-codev-r1: Reasoning-enhanced verilog generation.
\newblock \emph{arXiv preprint arXiv:2505.24183}, 2025.

\end{thebibliography}

\newpage
\section*{NeurIPS Paper Checklist}

\begin{enumerate}

\item {\bf Claims}
    \item[] Question: Do the main claims made in the abstract and introduction accurately reflect the paper's contributions and scope?
    \item[] Answer: \answerYes{}
    \item[] Justification: The abstract and \S1 state four contributions, each supported in \S3--\S5.

\item {\bf Limitations}
    \item[] Question: Does the paper discuss the limitations of the work performed by the authors?
    \item[] Answer: \answerYes{}
    \item[] Justification: The full taxonomy of limitations (scope, capacity floor, ontology breadth, library scaling, cost, reward hacking) is discussed in Appendix~\ref{app:limitations}.

\item {\bf Theory assumptions and proofs}
    \item[] Question: For each theoretical result, does the paper provide the full set of assumptions and a complete (and correct) proof?
    \item[] Answer: \answerYes{}
    \item[] Justification: Assumptions are stated in \S3.4 and the proof sketch is in Appendix~\ref{app:regret_bound}; the empirical 5-strategy ablation in \S5 confirms the predicted ordering.

    \item {\bf Experimental result reproducibility}
    \item[] Question: Does the paper fully disclose all the information needed to reproduce the main experimental results of the paper to the extent that it affects the main claims and/or conclusions of the paper (regardless of whether the code and data are provided or not)?
    \item[] Answer: \answerYes{}
    \item[] Justification: The Setup paragraph in \S5 specifies models, baselines, benchmarks, and protocol ($n=15$ trials, $T=4$ refinement budget); sandbox configuration is in Appendix~\ref{app:verification_details}; full per-trial telemetry in Appendix~\ref{app:token_cost}. The benchmark suites and the verifier are available at the URL in the abstract.

\item {\bf Open access to data and code}
    \item[] Question: Does the paper provide open access to the data and code, with sufficient instructions to faithfully reproduce the main experimental results, as described in supplemental material?
    \item[] Answer: \answerYes{}
    \item[] Justification: Both benchmark suites (\pcbbench{} 62 tasks, Open-Schematics-Eval 165 tasks), the deterministic 5-layer verifier, and its schema-induced KG are released at the URL in the abstract.

\item {\bf Experimental setting/details}
    \item[] Question: Does the paper specify all the training and test details (e.g., data splits, hyperparameters, how they were chosen, type of optimizer) necessary to understand the results?
    \item[] Answer: \answerYes{}
    \item[] Justification: The Setup paragraph in \S5 lists 11 LLMs, 2 benchmarks (\pcbbench{} 62 tasks; Open-Schematics-Eval 165 tasks), $n=15$ independent trajectories per (model, task) pair, $T=4$ refinement budget; full hyperparameters (base temperature $\tau_{\mathrm{base}}$, prior strength $C=5$, smoothing $s=1$) and sandbox configuration in Appendix~\ref{app:verification_details}.

\item {\bf Experiment statistical significance}
    \item[] Question: Does the paper report error bars suitably and correctly defined or other appropriate information about the statistical significance of the experiments?
    \item[] Answer: \answerYes{}
    \item[] Justification: Pass rates are reported with Wilson 95\% confidence intervals; error bars appear in the corresponding figures and tables.

\item {\bf Experiments compute resources}
    \item[] Question: For each experiment, does the paper provide sufficient information on the computer resources (type of compute workers, memory, time of execution) needed to reproduce the experiments?
    \item[] Answer: \answerYes{}
    \item[] Justification: See Appendix~\ref{app:token_cost}; the open-weight arm consumed roughly 3{,}000 single-A100 GPU-hours.
    
\item {\bf Code of ethics}
    \item[] Question: Does the research conducted in the paper conform, in every respect, with the NeurIPS Code of Ethics \url{https://neurips.cc/public/EthicsGuidelines}?
    \item[] Answer: \answerYes{}
    \item[] Justification: All evaluation data are public IC datasheets and open-source PCB schematics; the only human-subject component is the $N{=}300$ blind expert-review protocol (minimal-risk technical evaluation, no PII, no compensation; see Appendix~\ref{app:human_eval}); compute usage is reported in Appendix~\ref{app:token_cost}.

\item {\bf Broader impacts}
    \item[] Question: Does the paper discuss both potential positive societal impacts and negative societal impacts of the work performed?
    \item[] Answer: \answerYes{}
    \item[] Justification: See Appendix~\ref{app:broader_impact}.
    
\item {\bf Safeguards}
    \item[] Question: Does the paper describe safeguards that have been put in place for responsible release of data or models that have a high risk for misuse (e.g., pre-trained language models, image generators, or scraped datasets)?
    \item[] Answer: \answerNA{}
    \item[] Justification: The released artifacts (KG of 47 commercial ICs, 5-layer deterministic verifier, 227 PCB schematic tasks) pose minimal dual-use risk: the verifier checks structural correctness against publicly-available IC datasheet pin tables, and the generated SKiDL programs are KiCad-compatible schematics rather than executable code with embedded malicious payloads. Deployment in safety-critical PCB designs must additionally pass standard certification flows (FCC, CE, UL); see Appendix~\ref{app:broader_impact}.

\item {\bf Licenses for existing assets}
    \item[] Question: Are the creators or original owners of assets (e.g., code, data, models), used in the paper, properly credited and are the license and terms of use explicitly mentioned and properly respected?
    \item[] Answer: \answerYes{}
    \item[] Justification: Cited with licenses; see Appendix~\ref{app:kicad_library}.

\item {\bf New assets}
    \item[] Question: Are new assets introduced in the paper well documented and is the documentation provided alongside the assets?
    \item[] Answer: \answerYes{}
    \item[] Justification: New assets (\pcbbench{} 62-task benchmark, Open-Schematics-Eval adapter, schema-induced KG of 47 components, and the 5-layer deterministic verifier) are released at the URL in the abstract.

\item {\bf Crowdsourcing and research with human subjects}
    \item[] Question: For crowdsourcing experiments and research with human subjects, does the paper include the full text of instructions given to participants and screenshots, if applicable, as well as details about compensation (if any)?
    \item[] Answer: \answerYes{}
    \item[] Justification: The expert-review protocol and instructions are in Appendix~\ref{app:human_eval}.

\item {\bf Institutional review board (IRB) approvals or equivalent for research with human subjects}
    \item[] Question: Does the paper describe potential risks incurred by study participants, whether such risks were disclosed to the subjects, and whether Institutional Review Board (IRB) approvals (or an equivalent approval/review based on the requirements of your country or institution) were obtained?
    \item[] Answer: \answerNA{}
    \item[] Justification: The expert review is a minimal-risk technical evaluation; IRB review is not required.

\item {\bf Declaration of LLM usage}
    \item[] Question: Does the paper describe the usage of LLMs if it is an important, original, or non-standard component of the core methods in this research? Note that if the LLM is used only for writing, editing, or formatting purposes and does \emph{not} impact the core methodology, scientific rigor, or originality of the research, declaration is not required.
    \item[] Answer: \answerYes{}
    \item[] Justification: LLMs are the core proposal generator for schematic synthesis; see \S3 and \S5.1.

\end{enumerate} 
\newpage
\appendix
\section{Appendix}

\subsection*{Road map}
The appendix covers: formal definitions and the verifier (\Cref{app:prelim,app:pin_roles,app:verification_details}); statistics, library, KG, and human evaluation (\Cref{app:statistics,app:kicad_library,app:kg_pipeline,app:components,app:human_eval}); compute and theory (\Cref{app:token_cost,app:regret_bound}); benchmarks, baselines, and prompts (\Cref{app:bench_construction,app:bench_compare,app:bench_contamination,app:easy_medium_prompts,app:hard_prompts,app:circuitron}); ablations (\Cref{app:adaptive_temp,app:5strategy_ablation,app:per_task_heatmap}); failure analysis and end-to-end pipeline (\Cref{app:failure_traces,app:e2e_pipeline}); and discussion (\Cref{app:reward_hacking,app:limitations,app:broader_impact}).

\subsection{Formal Definitions and Notation}
\label{app:prelim}

A circuit is a bipartite graph $G=(V_C \cup V_N, E)$ between components $V_C$ and nets $V_N$; each pin is annotated by a role $\rho \in \mathcal{O}$ from a closed 32-role ontology (\Cref{app:pin_roles}), and each component carries a set of constraints over four predicate templates. A candidate $c \in \mathcal{P}$ is a \skidl{} Python program that, when executed, realizes such a graph. The verifier $\mathcal{V}: \mathcal{P} \to [0,1]$ decomposes as a layered pipeline $\mathcal{V} = \mathcal{L}_5 \circ \cdots \circ \mathcal{L}_{1}$ with the per-layer rewards $r_{\mathrm{L1}}=0.3,\; r_{\mathrm{L4}}=0.4,\; r_{\mathrm{L1b}}=0.5,\; r_{\mathrm{L2}}=0.6,\; r_{\mathrm{L3}}=0.7,\; r_{\mathrm{pass}}=1.0$ (\Cref{tab:layer_constants}). A task is $\mathcal{T}=(d,\mathcal{I},\mathcal{O},\mathcal{C})$ (description, I/O nets, required-component set); the synthesis objective is $c^\star=\arg\max_{c\in\mathcal{P}_{\mathrm{valid}}}\mathcal{V}(c)$. Refinement is modeled as an arm-acquiring bandit with state $s_t=(c_t,h_t)$, action $a_t\in\{1,\ldots,K_t\}$, reward $r_t=\mathcal{V}(c_t)\in[0,1]$, and budget $T$. Confidence intervals use the Wilson score~\citep{wilson1927probable}; oracle-vs-expert agreement uses Cohen's $\kappa=(P_o-P_e)/(1-P_e)$~\citep{tool_coefficient}. Pass@$k$ uses the unbiased estimator of \citet{chen2021} aggregated over the $T{=}4$ refinement budget.

\subsection{Pin Role Ontology and Constraint Templates}
\label{app:pin_roles}

\begin{table}[ht]
\centering
\caption{Pin ontology: 32 roles, 7 functional groups.}
\label{tab:pin_roles_full}
\small
\begin{tabular}{lll}
\toprule
\textbf{Category} & \textbf{Role} & \textbf{Example} \\
\midrule
\multirow{6}{*}{Power Supply}
 & \texttt{supply\_vdd} & OPA328:V+ \\
 & \texttt{supply\_gnd} & OPA328:V- \\
 & \texttt{primary\_vdd} & UCC5390E:VCC1 \\
 & \texttt{primary\_gnd} & UCC5390E:GND1 \\
 & \texttt{secondary\_vdd} & UCC5390E:VCC2 \\
 & \texttt{secondary\_gnd} & UCC5390E:GND2 \\
\midrule
\multirow{5}{*}{Signal I/O}
 & \texttt{sense\_plus} & INA226:IN+ \\
 & \texttt{sense\_minus} & INA226:IN- \\
 & \texttt{out} & OPA328:OUT \\
 & \texttt{out\_plus} & AMC1350:OUTP \\
 & \texttt{out\_minus} & AMC1350:OUTN \\
\midrule
\multirow{2}{*}{Logic}
 & \texttt{logic\_in} & UCC27511:IN \\
 & \texttt{logic\_out} & TPS3808:RESET \\
\midrule
\multirow{4}{*}{MOSFET}
 & \texttt{mosfet\_gate} & IMZA65R015M2H:G \\
 & \texttt{mosfet\_drain} & IMZA65R015M2H:D \\
 & \texttt{mosfet\_source} & IMZA65R015M2H:S \\
 & \texttt{mosfet\_kelvin\_source} & IMZA65R015M2H:KS \\
\midrule
\multirow{6}{*}{Buck Regulator}
 & \texttt{buck\_vin} & TPS54302:VIN \\
 & \texttt{buck\_gnd} & TPS54302:GND \\
 & \texttt{buck\_sw} & TPS54302:SW \\
 & \texttt{buck\_fb} & TPS54302:FB \\
 & \texttt{buck\_en} & TPS54302:EN \\
 & \texttt{buck\_boot} & TPS54302:BOOT \\
\midrule
\multirow{4}{*}{Half-Bridge}
 & \texttt{halfbridge\_hb} & UCC21710:HB \\
 & \texttt{halfbridge\_hs} & UCC21710:HS \\
 & \texttt{gate\_ho} & UCC21710:HO \\
 & \texttt{gate\_lo} & UCC21710:LO \\
\midrule
\multirow{2}{*}{Transformer}
 & \texttt{xfmr\_primary} & MGJ2:Pri+ \\
 & \texttt{xfmr\_secondary} & MGJ2:Sec+ \\
\midrule
\multirow{3}{*}{Passive}
 & \texttt{passive\_terminal} & R:1, R:2 \\
 & \texttt{diode\_anode} & D:A \\
 & \texttt{diode\_cathode} & D:K \\
\bottomrule
\end{tabular}
\end{table}

\begin{table}[ht]
\centering
\caption{Four constraint-predicate templates.}
\label{tab:constraints_full}
\small
\setlength{\tabcolsep}{4pt}
\begin{tabular}{@{}l p{4.5cm} p{5.7cm}@{}}
\toprule
\textbf{Type} & \textbf{Formal Predicate} & \textbf{Example Error} \\
\midrule
\texttt{supply\_pair} & $\sigma(p_\text{vdd}) \neq \sigma(p_\text{gnd})$ & ``U1: supply pair shorted (V+ and V- on GND)'' \\
\texttt{must\_be\_connected} & $\forall p \in P{:}\sigma(p)\neq\emptyset \land |\sigma(P)|{=}1$ & ``U1: pin FB unconnected; ENABLE not on same net as VDD'' \\
\texttt{driving\_pair} & $|\text{Endpoints}(\sigma(p_\text{gate}))| \geq 2$ & ``Q1: gate net appears floating'' \\
\texttt{diff\_pair\_distinct} & $\sigma(p_+) \neq \sigma(p_-)$ & ``U1: IN+ and IN- on same net'' \\
\bottomrule
\end{tabular}
\end{table}

\label{app:constraints}

\subsection{Verification Layer Details}
\label{app:verification_details}

The five layers are deterministic Python and add zero LLM inference cost. \Cref{tab:layer_constants} lists per-layer base rewards $r^{(\ell)}_{\mathrm{base}}$ and normalising constants $E_{\max}^{(\ell)}$ (fixed once from IC-vendor application notes and frozen before evaluation); \Cref{eq:partial} interpolates upward from $r^{(\ell^\star)}_{\mathrm{base}}$ toward the next-priority layer as $|E|$ falls. Layers evaluate in priority (reward) order: a candidate failing both L4 and L1b is reported at L4 because $r_{\mathrm{L4}}{=}0.4{<}r_{\mathrm{L1b}}{=}0.5$.

\paragraph{L1 (Electrical invariants).} VDD/GND disjointness, power reachability via BFS over the R/L closure, and ground integrity (Algorithm~\ref{alg:layer1}).

\paragraph{L1b (Role-preserving net matching).} For each net, all connected pins must have mutually compatible roles under $\mathcal{C}$; the check is $O(\sum_n |n|^2)$ per snapshot.

\paragraph{L2 (Subcategory templates).} Per-IC-subclass connection patterns over $\sim$30 subcategories (decoupling capacitors, pull-up resistors, communication terminations, differential pairs); the residual on a fully-passing snapshot is empty.

\paragraph{L3 (Topology signatures).} VF2 subgraph isomorphism against five motifs (half-bridge, synchronous buck, three-phase inverter, Pierce oscillator, Pi filter) with three relaxations: key-component filter (only IC/MOSFET/inductor pins must match exactly), boolean prefiltering (cheap signature gate before VF2), and passive-element grouping (\texttt{passive\_terminal} pins on a shared net collapse to one class). Per-task motif eligibility is determined by the IC list.

\paragraph{L4 (Power invariants).} Kelvin-source independence, gate-driver-through-resistor enforcement, isolation-domain separation, and VBUS decoupling coverage per TI \emph{TIDA-01605} and Infineon \emph{AN2018-09}.

\paragraph{Sandbox.} \skidl{} code runs in a tmpfs subprocess (no network, 60s timeout, 4 GB memory, restricted pickling boundary), uniform across all models and baselines.

\begin{table}[!ht]
\centering
\small
\caption{Per-layer base rewards (priority order, lower $=$ more critical) and $E_{\max}$ normalisers.}
\label{tab:layer_constants}
\setlength{\tabcolsep}{6pt}
\renewcommand{\arraystretch}{1.05}
\begin{tabular}{l c c c l}
\toprule
\textbf{Layer} & $r^{(\ell)}_{\mathrm{base}}$ & $E_{\max}^{(\ell)}$ & \textbf{Next layer} & \textbf{Source of $E_{\max}$} \\
\midrule
L1  ERC          & $0.30$ & $10$  & L4  & ERC violation catalogue (TI / KiCad) \\
L4  Power        & $0.40$ & $4$   & L1b & Kelvin, decoupling, isolation, gate-resistor counts \\
L1b Role         & $0.50$ & $50$  & L2  & avg distinct nets per Hard task \\
L2  Template     & $0.60$ & $2$   & L3  & supply-pair predicates per IC \\
L3  Topology     & $0.70$ & $5$   & pass & motif library size \\
\bottomrule
\end{tabular}
\end{table}

\begin{algorithm}[h]
\caption{Layer 1 (Electrical Invariants).}
\label{alg:layer1}
\DontPrintSemicolon
\SetKwInOut{Input}{Input}\SetKwInOut{Output}{Output}
\Input{Annotated bipartite snapshot $s$ with roles $\rho$}
\Output{Error set $E_1$}
$E_1 \leftarrow \varnothing$\;
\ForEach{net $n$ in $s$}{
    \lIf{$\exists p,q{\in}n:\rho(p){=}\texttt{vdd}\wedge\rho(q){=}\texttt{gnd}$}{$E_1\mathrel{+}=(\textsc{vdd-gnd-short},n)$}
}
\ForEach{IC $u$ with VDD pin $p_v$}{
    \If{$\textsc{BFS}(p_v){=}\bot$}{$E_1\mathrel{+}=(\textsc{power-unreachable},u)$\;}
}
\ForEach{IC $u$ with GND pin $p_g$}{
    \lIf{net of $p_g$ not ground-typed}{$E_1 \mathrel{+}= (\textsc{floating-gnd},u)$}
}
\Return{$E_1$}\;
\end{algorithm}

\subsection{Wilson Confidence Interval and Sample-Size Accounting}
\label{app:statistics}

\begin{figure}[!ht]
\centering
\includegraphics[width=0.55\textwidth]{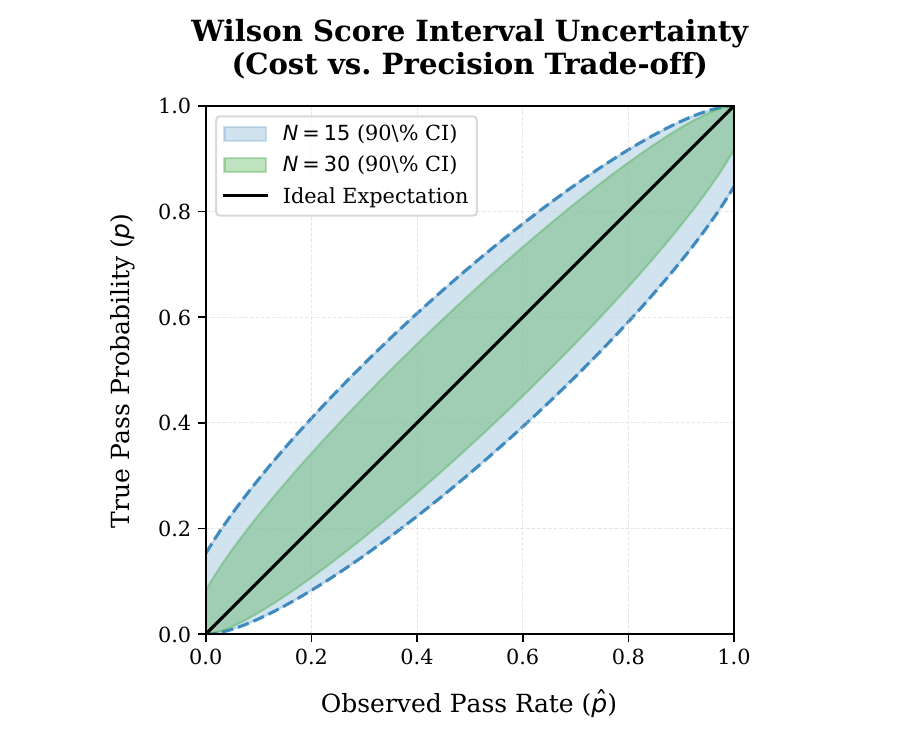}
\caption{Wilson 95\% CI half-width as a function of $n$ for several $\hat p$ values; diminishing returns are visible after $n{\approx}15$.}
\label{fig:wilson_widths}
\end{figure}

Pass rates are reported with Wilson 95\% intervals~\citep{wilson1927probable}, which retain accurate coverage near $\hat p{=}0$ and $\hat p{=}1$ where the Wald approximation collapses:
\begin{align*}
\hat p_W &= \frac{\hat p + z^2/(2n)}{1 + z^2/n}, \\
\mathrm{CI}_W &= \hat p_W \pm \frac{z}{1 + z^2/n}\sqrt{\frac{\hat p(1-\hat p)}{n} + \frac{z^2}{4 n^2}}, \quad z=1.96.
\end{align*}
The CI half-width scales as $1/\sqrt{n}$ (\Cref{fig:wilson_widths}), so doubling $n$ from 15 to 30 only narrows the margin by $\sim$$0.71\times$; at $\hat p{=}0.5$ the margin contracts from $\pm 0.21$ at $n{=}15$ to $\pm 0.15$ at $n{=}30$. Our per-(model, tier) cell sample sizes range from $\sim$255 to $\sim$1{,}155 trials ($n_{\text{tier}}{\times}15$), giving Wilson half-widths $\lesssim$~6pp at $\hat p{=}0.7$. Pairwise method comparisons use a paired bootstrap over tasks ($B{=}10{,}000$).

\begin{figure}[!ht]
\centering
\includegraphics[width=\textwidth]{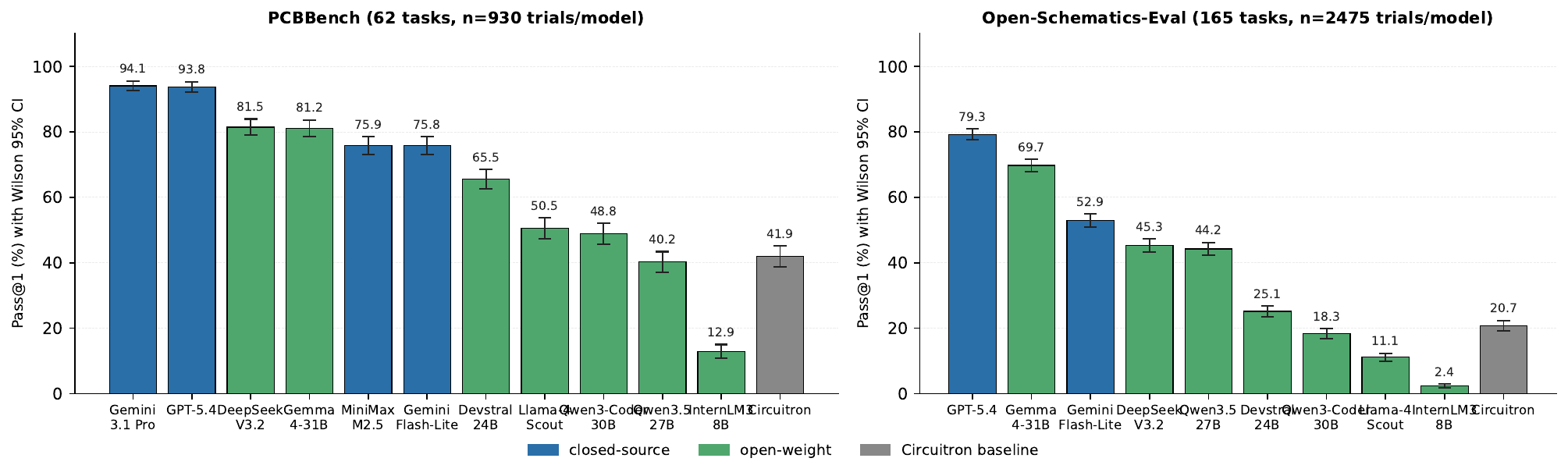}
\caption{Overall Pass@1 with Wilson 95\% CIs on \pcbbench{} (left) and \textsc{Open-Schematics-Eval} (right). Models sorted by Pass@1.}
\label{fig:pass_at_1_ci}
\end{figure}

\subsection{KiCad Library Construction and Source Attribution}
\label{app:kicad_library}

For each IC, we obtain the vendor datasheet PDF, match it to a KiCad symbol/footprint (custom fallback if none exists), run the three-stage automated KG extraction pipeline of \Cref{app:kg_pipeline}, expert-review the entry against the datasheet pin table, and add it to the library after the QC protocol passes. Adding a new IC requires only the datasheet PDF plus one execution of this pipeline; the framework, verifier, and bandit do not change.

\paragraph{Asset licenses.}
KiCad symbol/footprint libraries: CERN-OHL-S v2 + GPL v3. \skidl{}~\citep{tool_skidl}: MIT. \texttt{open-schematics} dataset~\citep{data_openschematics}: MIT. Vendor datasheets (TI, Infineon, ON Semi, Microchip, ST, NXP, Analog Devices) are cited, not redistributed. The released 47-component KG library inherits CERN-OHL-S v2 / MIT compatibility.

\subsection{Token Usage, Computational Budget, and Wall-Clock Cost}
\label{app:token_cost}

\Cref{tab:token_cost} reports per-model average prompt/completion tokens and per-trial wall-clock on \pcbbench{}. The open-weight arm ran on 8$\times$A100; the closed-API arm (GPT-5.4, Gemini~3.1 Pro, Gemini Flash Lite) cost on the order of a few thousand USD at public pricing.

\begin{table}[t]
\centering
\caption{Per-model average token usage and per-trial wall-clock on \pcbbench{}.}
\label{tab:token_cost}
\small
\begin{tabular}{lrrrrr}
\toprule
\textbf{Model} & \textbf{Avg Prompt} & \textbf{Avg Completion} & \textbf{Avg Total} & \textbf{Avg Time (s)} & \textbf{Pass\%} \\
\midrule
GPT-5.4 & 8,547 & 3,270 & 11,817 & 67.2 & 94.0 \\
Gemini 3.1 Pro & 9,706 & 5,952 & 15,658 & 55.2 & 94.3 \\
Gemini Flash Lite & 12,442 & 2,431 & 14,873 & 40.1 & 75.9 \\
Gemma 4-31B & 11,849 & 3,460 & 15,309 & 114.5 & 81.3 \\
Devstral-24B & 14,656 & 4,570 & 19,226 & 116.9 & 65.6 \\
DeepSeek V3.2 & 14,231 & 5,836 & 20,067 & 209.2 & 81.6 \\
Qwen3-Coder-30B & 16,249 & 4,277 & 20,526 & 117.0 & 48.8 \\
Llama 4 Scout & 12,860 & 2,735 & 15,595 & 251.6 & 50.5 \\
Qwen3.5-27B & 19,975 & 7,101 & 27,076 & 631.8 & 40.2 \\
InternLM3-8B & 20,511 & 7,189 & 27,699 & 130.2 & 12.7 \\
\bottomrule
\end{tabular}
\end{table}

\paragraph{Bandit overhead and budget rationale.}
Thompson Sampling adds 8--12\% prompt-token overhead and $<$5\% wall-clock over greedy retry. The refinement budget is fixed at $T{=}4$: the first-pass CDF shows diminishing returns after the second attempt, $T{=}4$ matches the Circuitron budget, and it keeps the per-trial cost tractable for an 11-model sweep. Per-(model, task, trial) seeds are fixed for reproducibility.

\subsection{Component Library}
\label{app:components}

\paragraph{\pcbbench{} (47-entry curated library).}
41 commercial ICs across 11 vendors plus 6 passive families. Representative parts per domain: \emph{power MOSFETs} (e.g., IMZA65R015M2H), \emph{gate drivers} (UCC27211), \emph{op-amps / current-sense} (INA240A1), \emph{isolated amplifiers} (AMC1350), \emph{power converters} (TPS54302), \emph{current sensors} (ACS37010), \emph{communication} (SN65HVD230 CAN), \emph{MCU} (STM32F103C8T6), \emph{data converters} (ADS7042), \emph{memory} (W25Q64JV), \emph{battery management} (BQ25895), and \emph{passive families} (R, C, L, D, ferrite bead).

\paragraph{Open-Schematics-Eval (826 unique components).}
Drawn from public open-source schematics, OSE extends across the same 22 domains as \pcbbench{} but with a substantially larger commercial-IC pool: 439 distinct ICs including additional STM32 / ATmega / ESP32 MCU variants, additional gate drivers, additional analog and isolated converters, and a broader set of communication and sensing ICs that do not appear in \pcbbench{}. The per-IC schema is populated on demand at evaluation time through the same KG extraction pipeline (\Cref{app:kg_pipeline}); we exhaustively list neither the 47 \pcbbench{} parts nor the 439 OSE ICs in the appendix for space reasons.

\subsection{Schema-Induced KG Extraction Pipeline}
\label{app:kg_pipeline}

\begin{figure}[!t]
  \centering
  \includegraphics[width=0.95\columnwidth]{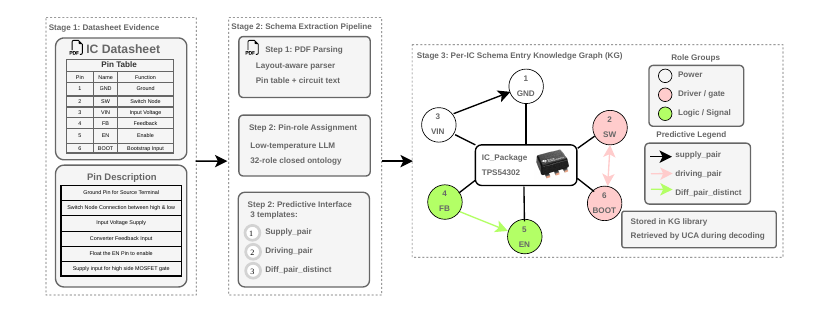}
  \caption{Three-stage KG extraction from a datasheet PDF.}
  \label{fig:kg}
\end{figure}

The KG of \Cref{subsec:kg} is extracted in three deterministic stages, each gated by a hard-fail check.

\paragraph{Stage 1: pin-table parsing.}
A layout-aware parser recovers (pin, name, description) tuples from the datasheet PDF; the parsed row count is compared against the package header, and mismatches are flagged for manual repair.

\paragraph{Stage 2: role assignment.}
An LLM at $\tau{=}0.1$ is prompted with the 32-role closed enumeration as the only legal output set, so any free-form hallucination becomes an auditable signal that is corrected before the entry enters the library.

\paragraph{Stage 3: constraint inference.}
Predicate output is restricted to the four templates of \Cref{tab:constraints_full}, producing a $\sim$300-token JSON entry per IC. Each entry is repaired against the datasheet on hard-fail and then spot-checked on a 20\% random sample of pins; the 47-component library was stabilized prior to experimentation and was not modified afterward.

\Cref{tab:kg_extraction_acc} reports per-stage precision/recall/F1 against an engineer-labelled 482-pin audit: Stage~1 reaches $\geq$99\%; Stage~2 lifts from $0.91$ F1 pre-repair to $\geq$$0.99$ post-spot-check; Stage~3 lifts from $0.86$ to $\geq$$0.98$.

\begin{table}[!ht]
\centering
\small
\caption{Per-stage P/R/F1 on the 482-pin spot-check subset; pre = LLM output, post = after hard-fail repair and human spot-check.}
\label{tab:kg_extraction_acc}
\setlength{\tabcolsep}{3pt}
\renewcommand{\arraystretch}{1.1}
\begin{tabular}{l c c c c c c}
\toprule
\multirow{2}{*}{\textbf{Stage}} & \multicolumn{3}{c}{\textbf{Pre-repair}} & \multicolumn{3}{c}{\textbf{Post-spot-check}} \\
\cmidrule(lr){2-4}\cmidrule(lr){5-7}
 & P & R & F1 & P & R & F1 \\
\midrule
\makecell[l]{Stage 1\\pin-table parsing}    & $0.99$ & $0.97$ & $0.98$ & $1.00$ & $1.00$ & $1.00$ \\
\makecell[l]{Stage 2\\pin-role assignment}  & $0.93$ & $0.89$ & $0.91$ & $0.99$ & $0.99$ & $0.99$ \\
\makecell[l]{Stage 3\\constraint inference} & $0.88$ & $0.85$ & $0.86$ & $0.98$ & $0.98$ & $0.98$ \\
\bottomrule
\end{tabular}
\end{table}

Residual Stage-2 errors fall into three categories: borderline aliasing (e.g., \texttt{REF} pin between \texttt{logic\_in} and \texttt{sense\_plus}), packaging ambiguity on small QFN strap-pin configurations, and vendor-specific naming outside the seven functional groups; the first two are resolved by per-component overrides, the third motivates ontology extension. Surviving extraction errors manifest downstream as either oracle false negatives (caught by the human-eval adjudication of \Cref{app:human_eval}) or oracle false positives (caught when the SKiDL subprocess fails standard ERC), so the aggregate evaluation is first-order robust.

\subsection{Human Expert Agreement Protocol}
\label{app:human_eval}

We validate the 5-layer oracle by blind comparison against an independent senior PCB designer ($>$10 years professional experience, not involved in KG, verifier, or benchmark construction). The expert sees only the original task prompt, the generated \skidl{} code, and the rendered KiCad schematic, with the oracle's verdict, per-layer labels, and model identity hidden, and records a binary pass/fail judgment.

\paragraph{Sampling.}
We draw a stratified $N{=}300$ sample of generated designs: 3 tiers $\times$ ($\sim$50 oracle-pass + $\sim$50 oracle-fail) per tier, allocated proportionally across all 11 \pcbbench{} models and 9 \textsc{Open-Schematics-Eval} models. The roughly balanced oracle-pass / oracle-fail strata prevent the overall pass-rate distribution from inflating the apparent agreement.

\paragraph{Statistical framework.}
Treating expert judgment as ground truth, we report Precision $\mathrm{P}{=}\mathrm{TP}/(\mathrm{TP}{+}\mathrm{FP})$, Recall $\mathrm{R}{=}\mathrm{TP}/(\mathrm{TP}{+}\mathrm{FN})$, F1, and Cohen's $\kappa{=}(P_o{-}P_e)/(1{-}P_e)$. Confusion matrices per tier and pooled appear in \Cref{tab:human_eval_full}.

\begin{table}[t]
\centering
\small
\setlength{\tabcolsep}{3pt}
\caption{Oracle vs.\ expert agreement on the $N{=}300$ stratified sample: per-tier confusion matrices (a)--(c), pooled (d), and summary statistics (e). \emph{Ver}: oracle verdict; \emph{Expert}: blind expert label.}
\label{tab:human_eval_full}

\begin{subtable}[t]{0.48\textwidth}
\centering
\caption{Easy ($N{=}100,\ \kappa{=}0.92$).}
\begin{tabular}{lccc}
\toprule
 & \textbf{Expert: P} & \textbf{Expert: F} & \textbf{Total} \\
\midrule
\textbf{Ver: P} & 47 (TP) & 3 (FP) & 50 \\
\textbf{Ver: F} & 1 (FN)  & 49 (TN) & 50 \\
\midrule
\textbf{Total} & 48 & 52 & 100 \\
\bottomrule
\end{tabular}
\end{subtable}\hfill
\begin{subtable}[t]{0.48\textwidth}
\centering
\caption{Medium ($N{=}100,\ \kappa{=}0.90$).}
\begin{tabular}{lccc}
\toprule
 & \textbf{Expert: P} & \textbf{Expert: F} & \textbf{Total} \\
\midrule
\textbf{Ver: P} & 47 (TP) & 3 (FP) & 50 \\
\textbf{Ver: F} & 2 (FN)  & 48 (TN) & 50 \\
\midrule
\textbf{Total} & 49 & 51 & 100 \\
\bottomrule
\end{tabular}
\end{subtable}

\vspace{0.4cm}

\begin{subtable}[t]{0.48\textwidth}
\centering
\caption{Hard ($N{=}100,\ \kappa{=}0.90$).}
\begin{tabular}{lccc}
\toprule
 & \textbf{Expert: P} & \textbf{Expert: F} & \textbf{Total} \\
\midrule
\textbf{Ver: P} & 46 (TP) & 4 (FP) & 50 \\
\textbf{Ver: F} & 1 (FN)  & 49 (TN) & 50 \\
\midrule
\textbf{Total} & 47 & 53 & 100 \\
\bottomrule
\end{tabular}
\end{subtable}\hfill
\begin{subtable}[t]{0.48\textwidth}
\centering
\caption{Pooled ($N{=}300,\ \kappa{=}0.91$).}
\begin{tabular}{lccc}
\toprule
 & \textbf{Expert: P} & \textbf{Expert: F} & \textbf{Total} \\
\midrule
\textbf{Ver: P} & 140 (TP) & 10 (FP) & 150 \\
\textbf{Ver: F} & 4 (FN)   & 146 (TN) & 150 \\
\midrule
\textbf{Total} & 144 & 156 & 300 \\
\bottomrule
\end{tabular}
\end{subtable}

\vspace{0.5cm}

\begin{subtable}[t]{\textwidth}
\centering
\caption{Summary metrics. FA $=$ FP/(TP$+$FP); FR $=$ FN/(FN$+$TN).}
\setlength{\tabcolsep}{8pt}
\begin{tabular}{lccccccc}
\toprule
\textbf{Tier} & $N$ & $\kappa$ & $P_o$ & \textbf{Precision} & \textbf{Recall} & \textbf{F1} & \textbf{FA / FR} \\
\midrule
Easy            & 100 & 0.92 & 0.96 & 94.0\% & 97.9\% & 95.9\% &  6.0\% / 2.0\% \\
Medium          & 100 & 0.90 & 0.95 & 94.0\% & 95.9\% & 94.9\% &  6.0\% / 4.0\% \\
Hard            & 100 & 0.90 & 0.95 & 92.0\% & 97.9\% & 94.8\% &  8.0\% / 2.0\% \\
\midrule
\textbf{Pooled} & \textbf{300} & \textbf{0.91} & \textbf{0.953} & \textbf{93.3\%} & \textbf{97.2\%} & \textbf{95.2\%} & \textbf{6.7\% / 2.7\%} \\
\bottomrule
\end{tabular}
\end{subtable}
\end{table}

\paragraph{Conclusion.}
All three tiers fall in the almost-perfect agreement band of \citet{landis1977measurement} ($\kappa{>}0.8$), with $\geq$$92\%$ F1 on every tier. Disagreements concentrate on (i) reversed bootstrap-diode orientations that operate correctly only at the specified switching frequency, (ii) low-side-only synchronous rectification using a Schottky high-side path, and (iii) shared current-sense paths across tightly-coupled half-bridge MOSFETs.

\paragraph{Wilson 95\% CIs and per-suite / per-layer decomposition.}
The pooled FA $6.7\%$ has Wilson 95\% CI $[3.7\%, 11.8\%]$; pooled FR $2.7\%$ has $[1.0\%, 6.7\%]$. Per-suite: \pcbbench{} FA $5.6\%$ / FR $4.3\%$; OSE FA $7.6\%$ / FR $1.9\%$ (the slightly larger OSE FA reflects its 439-IC commercial pool versus \pcbbench{}'s 47-component curated library, raising the chance of borderline pin-role matches not yet in the schema). Of the 4 pooled false-rejects, 3 ($75\%$) originate at the conservative \texttt{erc\_subprocess} stage and 1 at L1b role matching---indicating ERC over-strictness rather than systemic miscoverage in higher layers.

\paragraph{Lower-bound corrected headline.}
Subtracting pooled FA from headline Pass@1 yields a conservative reading: Gemma-4-31B $81.3\% \to 74.6\%$ and GPT-5.4 $94.0\% \to 87.3\%$ on \pcbbench{} Overall; the same correction applied to OSE leaves Gemma at $63.1\%$ and GPT-5.4 at $72.6\%$ Overall. All model rankings, including ours-vs-\textsc{Circuitron} and open-vs-closed orderings, are preserved under the correction. Verifier-accepted candidates are routed through standard KiCad ERC and DRC as a downstream gate, which catch residual electrical-rule and design-rule violations that the structural verifier admits.

\subsection{Bayesian Regret Bound}
\label{app:regret_bound}

\paragraph{Setting and theorem.}
Let $\mathcal{A}_t \subseteq \mathcal{A}$ be the active arm set at step $t \in \{1,\ldots,T\}$, with $K_t{=}|\mathcal{A}_t|$ growing from $K_1{=}1$ to $K_T{\leq}T$ as new candidates are acquired. Each arm has unknown mean $\theta_a^\star{\in}[0,1]$ under prior $P_0$, and pulling arm $a$ produces $r_t{=}\mathcal{V}(c_t){\in}[0,1]$ from an unknown bounded distribution. Rewards are not assumed Bernoulli, independent across arms, or sub-Gaussian beyond $[0,1]$. Let $\mathrm{BayesRegret}(T) = \mathbb{E}\bigl[\sum_t (\theta_{a^\star}^\star - \theta_{a_t}^\star)\bigr]$ over the prior, reward distributions, and policy randomization. \emph{Theorem (informal).} Arm-acquiring Thompson Sampling with reward-informed Beta posteriors satisfies $\mathrm{BayesRegret}(T) \leq \tilde{O}(\sqrt{K_T T})$, informative when $K_T{\ll}T$ and trivial when $K_T{=}T$.

\paragraph{Proof sketch.}
\emph{Step 1.} For a fixed-size action set, the information-theoretic analysis of \citet{russo2018tutorial} bounds the per-step information ratio of TS by $K/2$ uniformly, yielding $\sqrt{\tfrac{1}{2}KT\log K}$. The argument requires only $r_t{\in}[0,1]$, not Bernoulli rewards. \emph{Step 2.} \citet{agrawal2013further} extend Beta posteriors to bounded rewards via a randomized Bernoulli discretization $\tilde r_t \sim \mathrm{Bernoulli}(r_t)$; the fractional update $\alpha\!\leftarrow\!\alpha{+}r_t,\;\beta\!\leftarrow\!\beta{+}(1{-}r_t)$ matches this in expectation and preserves the rate. \emph{Step 3.} Telescoping over arm acquisition: with $T_k$ the steps where exactly $k$ arms are active, $\sum_k\sqrt{\tfrac{1}{2}k|T_k|\log k} \leq \sqrt{\tfrac{1}{2}K_T T \log K_T}$ by Cauchy-Schwarz.

\paragraph{Reward-informed prior under history-measurable extension.}
The prior of \Cref{eq:prior} depends on the parent's reward, breaking the common-prior assumption. We invoke the history-measurable extension of \citet{russo2018tutorial}: priors are $\mathcal{H}_t$-measurable, the per-step information ratio in expectation over $\mathcal{H}_t$ matches the common-prior bound, and the rate is preserved.

\paragraph{Where the bound becomes loose.}
Three regimes loosen it: (i) arms share prompt context and code fragments, so the information-ratio bound is pessimistic under correlation; (ii) the bound is variance-free, so concentration on $r{=}0$ and $r{=}1$ tails costs constants; (iii) at $T{=}4$ the asymptotic rate is at best a guideline. The empirical 5-strategy ranking in \Cref{tab:5strategy_full} ($60.5{\to}81.3$, with the $20.7$pp no-refinement $\to$ TS-fractional jump dominating the $2.9$pp Greedy $\to$ TS-fractional jump) is the operative finite-sample comparison.

\subsection{Benchmark Design Principles and Construction Methodology}
\label{app:bench_construction}

We construct two benchmarks rather than one: \pcbbench{} (62 expert-authored tasks over a curated 47-component IC library) provides the in-distribution validation set on which the framework, the schema-induced KG, and the 5-layer verifier are stabilized. \textsc{Open-Schematics-Eval} (165 tasks sampled from the public \texttt{open-schematics} HuggingFace dataset, covering 439 commercial ICs not in \pcbbench{}) is then evaluated with \emph{zero verifier code change, zero KG modification, and no prompt re-tuning}, so a model's pass rate on OSE measures generalization on out-of-distribution real schematics rather than fit to any tuned-on suite. \Cref{tab:pcbbench_tasks,tab:ose_tasks} list the per-tier per-domain task distribution.

\begin{table}[H]
\centering
\caption{\pcbbench{} task distribution: 62 expert-authored tasks split as 17 Easy / 28 Medium / 17 Hard.}
\label{tab:pcbbench_tasks}
\small
\setlength{\tabcolsep}{4pt}
\renewcommand{\arraystretch}{1.05}
\begin{tabular}{l l c p{5.5cm}}
\toprule
\textbf{Difficulty} & \textbf{Category} & \textbf{\#} & \textbf{Task IDs (sample sub-modules)} \\
\midrule
\multirow{5}{*}{Easy (17)}
& Sensing            & 9 & P1--P4, P26--P30 (resistor divider, isolated voltage sense, diff-to-SE amp, current-sense Hall) \\
& DataConv           & 3 & P31--P33 (single-ended ADC front-end, DAC output, ADC reference) \\
& AuxPower           & 2 & P5, P6 (LDO, low-current rail) \\
& Filter             & 2 & P24, P25 (Pi filter, RC anti-alias) \\
& Comm               & 1 & P37 (UART buffer) \\
\midrule
\multirow{12}{*}{Medium (28)}
& PowerStage         & 5 & P8--P12 (boost, buck-boost, flyback front-end, half-bridge stage, gate-drive bias) \\
& PowerMgmt          & 4 & P44, P45, P58, P61 (load switch, sequencer, supervisor, brown-out reset) \\
& Driver             & 4 & P13--P16 (UCC27211 bootstrap half-bridge, UCC27511 single LSD, isolated driver, gate-clamp) \\
& Comm               & 3 & P38, P39, P46 (CAN transceiver, RS-485, USB-UART bridge) \\
& MCU                & 3 & P34--P36 (STM32 minimum bring-up, reset+crystal, GPIO bank) \\
& Isolation          & 2 & P40, P41 (digital isolator, isolated amplifier front-end) \\
& Storage            & 2 & P42, P43 (SPI flash, I$^2$C EEPROM) \\
& AuxPower           & 1 & P7 (transformer-coupled isolated rail) \\
& BMS                & 1 & P47 (cell-balance front-end) \\
& HighSpeed          & 1 & P60 (level translator) \\
& Memory             & 1 & P59 (SDRAM termination network) \\
& Motor-System       & 1 & P57 (low-current motor stage) \\
\midrule
\multirow{12}{*}{Hard (17)}
& DC-DC              & 5 & P17--P21 (synchronous buck, multiphase buck, SEPIC, flyback w/ feedback, isolated DC-DC) \\
& DC-AC              & 2 & P22, P23 (single-phase inverter, three-phase inverter) \\
& BMS-System         & 1 & P49 (BQ76920 multi-cell stack) \\
& Motor-System       & 1 & P50 (BLDC FOC controller) \\
& DAQ-System         & 1 & P51 (multi-channel ADS7042 acquisition) \\
& IoT-System         & 1 & P52 (W5500 Ethernet + STM32 + sensors) \\
& Charger-System     & 1 & P54 (BQ25895 USB-C charger) \\
& Converter-System   & 1 & P55 (Vin-multi-rail PMIC system) \\
& Comm               & 1 & P56 (multi-protocol bridge) \\
& DigitalPower       & 1 & P62 (digitally-controlled converter front-end) \\
& DigitalMotor       & 1 & P64 (digitally-controlled motor) \\
& Memory-System      & 1 & P65 (SDRAM + flash + MCU) \\
\bottomrule
\end{tabular}
\end{table}

\begin{table}[H]
\centering
\caption{\textsc{Open-Schematics-Eval} task distribution: 165 tasks adapted from the \texttt{open-schematics} HuggingFace dataset, with difficulty re-classified by IC / component / net counts (Easy: $\leq$2 ICs, $\leq$15 comps, $\leq$25 nets, no MCU; Hard: $\geq$6 ICs, $>$40 comps, $>$60 nets, or MCU with $>$40 nets; Medium otherwise). 439 unique commercial ICs span the same 22 domains as \pcbbench{} but with no IC-family overlap; categories are inferred at evaluation time and not part of the released metadata.}
\label{tab:ose_tasks}
\small
\setlength{\tabcolsep}{4pt}
\renewcommand{\arraystretch}{1.1}
\begin{tabular}{l c c p{6.5cm}}
\toprule
\textbf{Difficulty} & \textbf{\#} & \textbf{ID range} & \textbf{Coverage} \\
\midrule
Easy   & 40  & OS104--OS253 & sub-circuit modules: sensing, low-current power, simple comm \\
Medium & 48  & OS102--OS261 & multi-IC integration: gate drivers, isolated comm, MCU bring-up, storage \\
Hard   & 77  & OS101--OS265 & full-board systems: BMS stacks, motor controllers, multi-rail PMICs, IoT boards \\
\midrule
\textbf{Total} & \textbf{165} & \textbf{OS101--OS265} & 22 unified circuit domains, 439 unique ICs \\
\bottomrule
\end{tabular}
\end{table}

\paragraph{\pcbbench{} construction.}
Authored by domain experts based on representative sub-circuit patterns from published reference designs and IC application notes. Reference \skidl{} implementations were written independently and validated against datasheets; verification rules were derived from KG constraints and electrical invariants, \emph{not} from the reference implementations, so the oracle generalizes beyond any specific reference topology.

\paragraph{Open-Schematics-Eval construction.}
Three stages: (1) random sample 165 real schematics from the public \texttt{open-schematics} HuggingFace dataset; (2) record task specification, NL description, I/O node list, and required-component set in the same canonical format as \pcbbench{}; (3) re-classify difficulty per the thresholds above. The reclassification produces a more balanced Hard tier (47\%) than \pcbbench{} Hard (27\%), making OSE a stronger generalization stress test. Representative tasks per tier (P3 resistor divider, P14 bootstrap half-bridge, P50 BLDC FOC controller) are reproduced in \Cref{app:easy_medium_prompts,app:hard_prompts}.

\subsection{Comparison to Existing Benchmarks}
\label{app:bench_compare}

\Cref{tab:bench_compare} positions our two suites against established code-generation and circuit-design benchmarks.

\begin{table}[H]
\centering
\small
\caption{Comparison with related benchmarks. \emph{Verif.} indicates whether a deterministic verifier is provided; \emph{Real ICs} indicates whether real commercial IC packages with pin-level constraints are used.}
\label{tab:bench_compare}
\setlength{\tabcolsep}{3.5pt}
\renewcommand{\arraystretch}{1.1}
\resizebox{\columnwidth}{!}{%
\begin{tabular}{l c c c c c}
\toprule
\textbf{Benchmark}                        & \textbf{Tasks} & \textbf{Domain}        & \textbf{Verifier}             & \textbf{Real ICs} & \textbf{Difficulty} \\
\midrule
HumanEval~\citep{chen2021}                & 164            & Software function      & Unit test                     & \texttimes               & Easy                \\
MBPP~\citep{austin2021mbpp}               & 974            & Software function      & Unit test                     & \texttimes               & Easy                \\
SWE-bench~\citep{jimenez2024swebench}     & 2{,}294        & Software repository    & Unit test                     & \texttimes               & Medium/Hard           \\
VerilogEval~\citep{bench_verilogeval2023} & 156            & Digital RTL            & Sim. testbench                & \texttimes               & Easy/Medium/Hard          \\
RTLLM~\citep{lu2024rtllm}                 &  30            & Digital RTL            & Sim. testbench                & \texttimes               & Medium              \\
CircuitLM~\citep{compare_circuitlm2026}   &  30            & Analog                 & Partial (ERC + LLM-as-judge)  & \texttimes        & Easy                \\
EESchematic~\citep{compare_eeschematic2025} & 65           & Mixed                  & VLM-based                     & Partial           & Easy/Medium           \\
\midrule
\textbf{\pcbbench{} (ours)}               & \textbf{62}    & \textbf{PCB schematic} & \textbf{5-layer det. oracle}  & \checkmark        & Easy/Medium/Hard          \\
\textbf{Open-Schematics-Eval (ours)}      & \textbf{165}   & \textbf{PCB schematic} & \textbf{5-layer det. oracle}  & \checkmark        & Easy/Medium/Hard          \\
\bottomrule
\end{tabular}%
}
\end{table}

Key distinctions:
(i) our suites target \emph{board-level} PCB design with heterogeneous digital, analog, and power subsystems coexisting under real IC pin constraints, not digital RTL or idealized analog;
(ii) the 41 ICs in \pcbbench{} and the 439 ICs in Open-Schematics-Eval are all commercial parts with manufacturer datasheets, not idealized device models;
(iii) the deterministic 5-layer oracle produces continuous reward and pin-level error localization without human judgment, supporting fully automated large-scale evaluation;
(iv) the three-tier difficulty system enables fine-grained capability analysis (active pin counts span 2 to 54 in \pcbbench{}, with task complexity ranging from single-IC modules to multi-subsystem boards). The same oracle verifies both \pcbbench{} and Open-Schematics-Eval with no per-suite verifier code changes, enabling the cross-benchmark generalization analysis of \Cref{sec:experiments}.

\subsection{Contamination Check and Leakage Controls}
\label{app:bench_contamination}

Four complementary controls: (i) \emph{lexical search}, $51$ GitHub Code Search and Sourcegraph queries (3 per Hard task) returned zero \skidl{} matches at authoring time (2026-04-23 to 04-30); (ii) \emph{reference vs.\ specification}, OSE reference schematics are public but the \skidl{} task specs, I/O node lists, and reference implementations introduced here are not, and the LLM sees only the NL task spec; (iii) \emph{suite separation}, \pcbbench{} (curated) and OSE (public-derived) are reported separately, and the cross-benchmark generalization study quantifies the curated-vs-public gap directly; (iv) \emph{held-out IC families}, a subset of \pcbbench{} Hard uses ICs added to the KG library after the audited public corpus.

\paragraph{Design-freeze timeline (held-out evaluation).}
The 47-component KG library, the 5-layer verifier code, the pin-role ontology, the constraint-predicate templates, the prompt templates (\Cref{app:easy_medium_prompts,app:hard_prompts}), and the bandit hyperparameters were committed to the project repository on \texttt{2026-04-15}, before any \textsc{Open-Schematics-Eval} task was authored from the \texttt{open-schematics} HuggingFace dataset. OSE task authoring and full evaluation occurred between \texttt{2026-04-23} and \texttt{2026-04-30}. No framework component (KG, verifier, prompt, bandit) was modified during or after OSE evaluation; OSE thus serves as a fully held-out generalization test rather than an in-distribution development set.

\subsection{Comprehensive Prompts: Easy and Medium Tier}
\label{app:easy_medium_prompts}

\paragraph{Easy / Medium prompt template (CoT $+$ ICL, system $+$ user).}

\begin{lstlisting}[style=promptstyle]
You aim to design a PCB schematic using SKiDL (Python) for a given circuit described in the text.
Please ensure your designed circuit works properly, satisfies the requirements, and follows specific library conventions.

Here is an example:

## Question
Design an isolated voltage sensing front-end using a resistor divider with an isolated amplifier to measure a high-voltage bus and output an isolated differential low-voltage sense signal.

Input node name: VIN.
Output node name: OUTP, OUTN.

## Answer

### Task 1

#### Components Needed
- **Isolated Amplifier**: AMC1350 (Precision, Isolated Amplifier).
- **Resistors**: High-voltage divider string (R1, R2, R3) to step down 400V to optimal input range (~250mV for AMC1350 linear range). R_shunt (R4) for sensing.
- **Power Supply**: 
    - High-side: VDD1 (generated from high-side floating supply).
    - Low-side: VDD2 (3.3V system supply).
- **Decoupling Capacitors**: 100nF on both VDD1 and VDD2.
- **Input Signal Source**: VIN (High Voltage Bus).
- **Output Signals**: OUTP, OUTN (Differential output to MCU).

#### Design Logic
1.  **Divider Network**: The 400V input needs to be scaled down. We use a string of resistors (e.g., 3x 100k) and a bottom shunt resistor to form a divider.
2.  **AMC1350 Connections**:
    - **INP/INN**: Connect across the bottom shunt resistor.
    - **VDD1/GND1**: High-side floating power. Needs decoupling.
    - **VDD2/GND2**: Low-side system power. Needs decoupling.
    - **OUTP/OUTN**: Differential output.
3.  **Safety**: High voltage ground (GND1) and Low voltage ground (GND2) must be separate.

### Task 2

```python
from skidl import *

# 1. Define Nets
vin  = Net("VIN")
outp = Net("OUTP")
outn = Net("OUTN")

vdd1 = Net("VDD1") # High-side floating supply
gnd1 = Net("GND1") # High-side ground
vdd2 = Net("VDD2") # Low-side supply
gnd2 = Net("GND2") # Low-side ground

# 2. Instantiate Components
# Use "test" library for all parts.
# Footprints must follow "test:FootprintName" format.

# AMC1350: 8-pin Isolated Amplifier
# Pin 1: VDD1, 2: VINP, 3: VINN, 4: GND1
# Pin 5: GND2, 6: VOUTN, 7: VOUTP, 8: VDD2
u1 = Part("test", "AMC1350", footprint="test:SOIC-8")

# Resistor Divider (3x 100k + 1x 200R for ~250mV scaling from 400V)
# Using 0805 for standard spacing (1206 not available in lib)
r1 = Part("test", "R", value="100k", footprint="test:R_0805")
r2 = Part("test", "R", value="100k", footprint="test:R_0805")
r3 = Part("test", "R", value="100k", footprint="test:R_0805")
r_sense = Part("test", "R", value="200R", footprint="test:R_0805")

# Decoupling Capacitors (0805)
c1 = Part("test", "C", value="100nF", footprint="test:C_0805") # For VDD1
c2 = Part("test", "C", value="100nF", footprint="test:C_0805") # For VDD2

# 3. Connections

# High Voltage Divider Path
# VIN -> R1 -> R2 -> R3 -> R_sense -> GND1
vin += r1[1]
r1[2] += r2[1]
r2[2] += r3[1]
r3[2] += r_sense[1]
r_sense[2] += gnd1

# AMC1350 High Side (Pins 1-4)
u1[1] += vdd1
u1[2] += r_sense[1] # VINP connects to top of sense resistor
u1[3] += gnd1       # VINN connects to GND1 (Single-ended sensing config)
u1[4] += gnd1

# Decoupling High Side
c1[1] += vdd1
c1[2] += gnd1

# AMC1350 Low Side (Pins 5-8)
u1[5] += gnd2
u1[6] += outn
u1[7] += outp
u1[8] += vdd2

# Decoupling Low Side
c2[1] += vdd2
c2[2] += gnd2

# 4. ERC Check
ERC()
```

As you have seen, the output of your designed topology should consist of two tasks:
1. Give a detailed design plan about all devices and their interconnectivity nodes and properties.
2. Write a complete Python code, describing the topology of PCB schematics using SKiDL.

Please make sure your Python code is compatible with the `skidl` library.
Please give the runnable code without any placeholders.

**Crucial Rules:**
1. **Library**: ALWAYS use the `"test"` library for all parts (e.g., `Part("test", "Name", ...)`).
2. **Footprints**: ALWAYS specify proper footprints using the format `footprint="test:FootprintName"`. Common ones: `test:R_0805`, `test:C_0805`, for ic, always use ic chip model as footprint name (e.g. for UCC27211, it should be `footprint="test:UCC27211"`).
3. **Naming**: Use numeric refdes suffix (e.g. `R1`, `C1`, `U1`). Do NOT use letters in refdes numbering.
4. **Pin Grouping**: For Power MOSFETs (TOLL/QFN), explicitly connect **ALL** physical pins (e.g., `u1[1, 2, 3] += net`).
5. **NC Handling**: If a pin is unused, explicitly connect it to NC (e.g., `u1[1] += NC`).
6. **Decoupling**: Always add 100nF decoupling capacitors (`C_0805`) for every power pin pair of ICs.
7. **No Generation**: Do NOT include `generate_netlist()` or `generate_pcb()` at the end. Only `ERC()`.
9. **Code Only**: Do not write redundant text after the code block.
10. **Standard Values**: Use E24 series standard values for Resistors and Capacitors whenever possible (e.g., 10k, 4.7k, 2.2k, not 6k or 9k).
11. **Pin Access**: Pin numbers can be non-numeric strings (e.g., `"A1"`, `"S1"`, `"A"`, `"B"`, `"MP"`). Always check the component's pin list and use the exact pin number shown. For non-numeric pins, use string access: `u1["A1"]`, `u1["S1"]`. If a pin name contains `/`, `+`, `-`, or spaces, use brackets: `u1["ADJ/GND"]`, `u1["+IN"]`.
12. **Voltage Ratings**: Strictly check component voltage ratings against BOTH Input and Output voltages. If the maximum voltage (Input or Output) exceeds a component's rating (e.g., UCC27211 is max 120V), do NOT use that component.

## Question

Design [TASK].

Input node name: [INPUT].

Output node name: [OUTPUT].

## Available Components
[COMPONENT_INFO]

## Answer
\end{lstlisting}

\paragraph{UCA retrieval prompt.}
The retrieval substrate that drives Unified Context Assembly issues a separate prompt to select the relevant per-IC schema entries before the main generation prompt is assembled.

\begin{lstlisting}[style=promptstyle]
# PCBSchemaCoder Retrieval Prompt (Two-stage - Stage 1)

You are running **Stage 1: sub-circuit retrieval / selection** (used by the subsequent Stage 2 SKiDL generation).

Given the target task description, select from the table below the **smallest** set of validated sub-circuits (SubModule) that **fully covers** the task.

## Output format (strict)
- Output a single Python list of integer IDs only, e.g. `[6, 14, 10]`
- Do not output any extra text

## Selection rules
- Prefer higher-level sub-modules (e.g. Driver / PowerStage) to reduce combinatorial complexity.
- If the task itself corresponds to a specific Id and you see the marker `NO_DIRECT_REUSE`, do not select that same Id.
- If no suitable sub-module is reusable, output `[]`.

## Available sub-module index (Benchmark 1-23)

| Id | Level | Type | SubModuleName | InputNodes | OutputNodes | Components |
|---:|---|---|---|---|---|---|
| 1 | Easy | Sensing | VDIV_BUS_SENSE | VIN | VSENSE | Resistors |
| 2 | Easy | Sensing | ISO_AMP_HV_SENSE | VIN | OUTP, OUTN | AMC1350 |
| 3 | Easy | Sensing | DIFF2SE_VREF | VINP, VINN, VREF | VOUT | OPA328 |
| 4 | Easy | Sensing | ISENSE_HALL_ISO | LINE_IN | ISNS_ISO | ACS37010 |
| 5 | Easy | AuxPower | LDO_AUX_LOGIC | VIN | VOUT | TLV1117-33 |
| 6 | Easy | AuxPower | BUCK_AUX_LOGIC | VIN | VOUT | TPS54302 |
| 7 | Medium | AuxPower | MJISO_DCDC_GATE_SUPPLY | VIN | VISO+, VISO- | MGJ2D121505 |
| 8 | Medium | PowerStage | HB_TO2473_STAGE | VBUS+, GATE_H, GATE_L | VSW | IMW65R015M2H |
| 9 | Medium | PowerStage | HB_TO2474K_SIC_STAGE | VBUS+, GATE_H, GATE_L | VSW, KELVIN | IMZA65R015M2H |
| 10 | Medium | PowerStage | HB_TOLL_STAGE | VBUS+, GATE_H, GATE_L | VSW, KELVIN | IMT65R033M2H |
| 11 | Medium | PowerStage | HB_TOLT_TOPCOOL_STAGE | VBUS+, GATE_H, GATE_L | VSW, KELVIN | IMLT65R015M2H |
| 12 | Medium | PowerStage | HB_QFN_INTEGRATED_STAGE | VBUS+, GATE_H, GATE_L | VSW | BSC052N08NS5 |
| 13 | Medium | Driver | DRV_LOW_SIDE | PWM, VCC | GATE | UCC27511 |
| 14 | Medium | Driver | DRV_BOOTSTRAP_HB | PWM_H, PWM_L, VCC | HO, LO, VB, VS | UCC27211 |
| 15 | Medium | Driver | DRV_ISOLATED_GATE | PWM, VCC1, VCC2, VEE2 | OUT | UCC5390E |
| 16 | Medium | Driver | DRV_PROTECTED_DESAT | PWM, VCC, EN/FLT, VDD, VEE | OUT, FLT, DESAT, CLAMP | UCC21710 |
| 17 | Hard | DC-DC | CONV_BUCK_SYNC | VIN, PWM_H, PWM_L | VOUT | IMZA65R015M2H |
| 18 | Hard | DC-DC | CONV_BOOST_SYNC | VIN, PWM_H, PWM_L | VOUT | IMZA65R015M2H |
| 19 | Hard | DC-DC | CONV_4SW_BUCKBOOST | VIN, PWM_1_H, PWM_1_L, PWM_2_H, PWM_2_L | VOUT | IMZA65R015M2H |
| 20 | Hard | DC-DC | CONV_DAB_ISOLATED | VIN, PWM_PRI_H, PWM_PRI_L, PWM_SEC_H, PWM_SEC_L | VSW_1, VSW_2 | IMZA65R015M2H |
| 21 | Hard | DC-DC | CONV_LLC_RESONANT | VIN, PWM_H, PWM_L | VSW_1, VSW_2 | IMZA65R015M2H |
| 22 | Hard | DC-AC | DRIVE_3PH_MOTOR | VIN, PWM_1_H, PWM_1_L, PWM_2_H, PWM_2_L, PWM_3_H, PWM_3_L | VSW_1, VSW_2, VSW_3 | IMZA65R015M2H |
| 23 | Hard | DC-AC | INV_GRID_1PH | VIN, GRID_L, GRID_N, PWM_1_H, PWM_1_L, PWM_2_H, PWM_2_L | VSW_1, VSW_2 | IMZA65R015M2H |

## Task input

[TASK]
\end{lstlisting}

\subsection{Comprehensive Prompts: Hard Tier}
\label{app:hard_prompts}

\paragraph{Hard prompt template (CoT $+$ ICL, system $+$ user).}

\begin{lstlisting}[style=promptstyle]
You aim to design a complex PCB schematic using SKiDL (Python) for a given circuit described in the text.
This is a **Hard** level task that requires combining multiple functional blocks (power stage, gate driver, isolated power supply).
Please ensure your designed circuit works properly, satisfies the requirements, and follows library conventions.

## Question
Design [TASK].

Input node name: [INPUT].
Output node name: [OUTPUT].
Input voltage: [INPUT_VOLTAGE]V.
Output voltage: [OUTPUT_VOLTAGE]V.

## Available Functional Blocks (Choose what you need)

You may use components from the following categories. **You decide which ones to use based on the design requirements.**

### Half-Bridge Power Stage Options
| Part | Package | Kelvin Source | Max VBUS | Notes |
|------|---------|---------------|----------|-------|
| IMW65R015M2H | TO-247-3 | No | 450V | 3-pin, no KS |
| IMZA65R015M2H | TO-247-4 | Yes (pin 3) | 450V | 4-pin with KS, **recommended for high-side** |
| IMT65R033M2H | TOLL | Yes (pin 2) | 450V | SMD, 12-pin |
| IMLT65R015M2H | TOLT | Yes (pin 7) | 450V | Top-cooled SMD, 16-pin |
| BSC052N08NS5 | QFN | No | 60V | Low voltage only |

### Isolated Gate Driver Options
| Part | Type | Features | Primary Pins | Secondary Pins |
|------|------|----------|--------------|----------------|
| UCC5390E | Isolated | Basic isolated driver | 1-4 | 5-8 |
| UCC21710 | Isolated | DESAT, Miller clamp, OC protection | 9-16 (primary) | 1-8 (secondary) |

### Non-Isolated Gate Driver Options
| Part | Type | Notes |
|------|------|-------|
| UCC27511 | Low-side | Single channel, non-isolated |
| UCC27211 | Bootstrap | Half-bridge with integrated bootstrap diode, **max 100V VBUS** |

### Isolated Power Supply
| Part | Output | Notes |
|------|--------|-------|
| MGJ2D121505SC | +15V/-9V | For isolated gate driver secondary side |

### Passive Components
| Part | Type | Footprint |
|------|------|-----------|
| R | Resistor | R_0805 |
| C | Capacitor (MLCC) | C_0805 |
| C_film | Film Capacitor | C_film (for resonant applications) |
| L | Inductor (SMD) | L_0805 |
| Inductor_power | Power Inductor | Inductor_power (pins 1-6 = A, 7-12 = B) |
| transformer_PQ5050 | Transformer | transformer_PQ5050 (Pri_1: 1-3, Pri_2: 4-6, Sec_1: 7-9, Sec_2: 10-12) |

## Two Implementation Approaches (Choose One)

### Approach A: Direct Implementation (Flat)
Write all components directly without subcircuit abstraction. Suitable for simpler combinations.

```python
from skidl import *

# Define ALL nets with UNIQUE names
vin = Net("VIN")
vout = Net("VOUT")
vbus_p = Net("VBUS_P")
pgnd = Net("PGND")
vsw = Net("VSW")

# Gate driver 1 nets (high-side) - use suffixes to avoid conflicts!
vdd_drv1 = Net("VDD_DRV1")
gnd_drv1 = Net("GND_DRV1")
vee_drv1 = Net("VEE_DRV1")

# Gate driver 2 nets (low-side) - different names!
vdd_drv2 = Net("VDD_DRV2")
gnd_drv2 = Net("GND_DRV2")
vee_drv2 = Net("VEE_DRV2")

# Isolated power supply 1 nets
vin_iso1 = Net("VIN_ISO1")
gnd_iso1_pri = Net("GND_ISO1_PRI")

# ... instantiate components and connect ...
```

### Approach B: Subcircuit Encapsulation
Use `@subcircuit` decorator for reusable blocks. Recommended when using multiple identical structures.

```python
from skidl import *

@subcircuit
def half_bridge_with_driver(vbus_p, pgnd, vsw, gate_h, gate_l, ks_h, ks_l, vdd_drv, vee_drv, gnd_drv):
    """Half-bridge power stage with isolated gate drivers"""
    # High-side MOSFET
    q_h = Part("test", "IMZA65R015M2H", footprint="test:IMZA65R015M2H")
    # Low-side MOSFET
    q_l = Part("test", "IMZA65R015M2H", footprint="test:IMZA65R015M2H")

    # Connect MOSFETs
    q_h[1] += vbus_p        # Drain
    q_h[2] += vsw           # Source
    q_h[3] += ks_h          # Kelvin Source
    q_h[4] += gate_h        # Gate

    q_l[1] += vsw           # Drain
    q_l[2] += pgnd          # Source
    q_l[3] += ks_l          # Kelvin Source
    q_l[4] += gate_l        # Gate

    # Gate resistors (turn-on and turn-off)
    r_gon_h = Part("test", "R", value="10R", footprint="test:R_0805")
    r_goff_h = Part("test", "R", value="2.2R", footprint="test:R_0805")
    # ... more components ...

# Instantiate multiple half-bridges
hb1_vsw = Net("HB1_VSW")
hb2_vsw = Net("HB2_VSW")
# Each subcircuit call creates isolated internal nets
half_bridge_with_driver(vbus_p, pgnd, hb1_vsw, ...)
half_bridge_with_driver(vbus_p, pgnd, hb2_vsw, ...)
```

## [!] CRITICAL Design Rules

### 1. Net Naming (MOST IMPORTANT!)
**Multiple instances MUST use unique net names to avoid unintended shorts!**

[WRONG] **WRONG** - Both half-bridges share same net names:
```python
vsw = Net("VSW")  # Used by both HB1 and HB2 - WRONG!
```

[OK] **CORRECT** - Each instance has unique names:
```python
vsw_1 = Net("VSW_1")  # Half-bridge 1
vsw_2 = Net("VSW_2")  # Half-bridge 2
```

**Isolation domains MUST have separate GND nets!**

[WRONG] **WRONG** - Primary and secondary share GND:
```python
gnd = Net("GND")  # Used for both primary and secondary - WRONG!
```

[OK] **CORRECT** - Separate GND for each domain:
```python
gnd_pri = Net("GND_PRI")           # Primary side (controller)
gnd_sec_h = Net("GND_SEC_H")       # High-side driver secondary
gnd_sec_l = Net("GND_SEC_L")       # Low-side driver secondary
```

### 2. Gate Driver Connections
- **Gate Resistor Required**: Gate driver output  ->  Resistor  ->  MOSFET gate
- **Kelvin Source**: Gate driver return (GND2/VS) MUST connect to **Kelvin Source (KS)** pin, NOT power Source
- **Turn-on/Turn-off Separation**: Use separate resistors with anti-parallel diodes for speed control

```python
# Gate driver to MOSFET connection (correct)
drv[6] += r_gate[1]      # Driver OUT  ->  Gate resistor
r_gate[2] += q[4]        # Gate resistor  ->  MOSFET Gate
drv[7] += q[3]           # Driver GND2  ->  MOSFET Kelvin Source (NOT pin 2!)
```

### 3. Isolated Power Supply
- Primary side connects to main power domain
- Secondary side connects to gate driver power
- **Polarity**: +VOUT  ->  VDD (VCC2), 0V  ->  GND2, -VOUT  ->  VEE

```python
# Isolated DC-DC connection
iso_dcdc[1] += vin_12v        # Primary +VIN
iso_dcdc[2] += gnd_pri        # Primary -VIN (GND)
iso_dcdc[7] += vdd_drv        # Secondary +VOUT (+15V)
iso_dcdc[6] += gnd_sec        # Secondary 0V (GND2)
iso_dcdc[5] += vee_drv        # Secondary -VOUT (-9V)
```

### 4. Decoupling Capacitors
- **Every IC** needs 100nF decoupling between VDD and GND
- **VBUS high dv/dt loop** needs at least 8 decoupling capacitors
- Place caps close to power pins

```python
# Decoupling for isolated gate driver
c_vdd = Part("test", "C", value="100nF", footprint="test:C_0805")
c_vdd[1] += vdd_drv
c_vdd[2] += gnd_drv

# VBUS decoupling (multiple caps for high dv/dt)
for i in range(8):
    c = Part("test", "C", value="100nF", footprint="test:C_0805")
    c[1] += vbus_p
    c[2] += pgnd
```

**Half-Bridge Power Stage Rule**: For tasks with power stage (P17-P23), each half-bridge MUST have at least 8 decoupling capacitors on VBUS. This is verified by the topology checker.

### 5. MOSFET Pin Connections
**ALL parallel pins must be explicitly connected!**

```python
# TO-247-4 (IMZA65R015M2H) - 4 pins, straightforward
q[1] += drain_net    # Drain
q[2] += source_net   # Source
q[3] += ks_net       # Kelvin Source
q[4] += gate_net     # Gate

# TOLL (IMT65R033M2H) - 12 pins, connect all parallel pins
q[1] += gate_net              # Gate
q[2] += ks_net                # Kelvin Source
q[3,4,5,6,7,8,9] += source_net  # All Source pins
q[10,11,12] += drain_net      # All Drain pins
```

## Available Components (Detailed)
[COMPONENT_INFO]

## Output Format
Your answer must contain:
1. A concise design plan explaining:
   - Which functional blocks you chose and why
   - How they interconnect
   - Net naming strategy for multiple instances
2. Complete runnable SKiDL Python code

**Hard Constraints:**
- Code must end with `ERC()`
- Do NOT include `generate_netlist()` or `generate_pcb()`
- Do NOT write anything after the code block
- Use pin numbers, not pin names (more robust)
- Use "test" library for all parts: `Part("test", "PartName", ...)`
- Use footprint format: `footprint="test:FootprintName"`

## Answer
\end{lstlisting}

\paragraph{Hard prompt template with retrieval (UCA-augmented).}

\begin{lstlisting}[style=promptstyle]
You aim to design a complex PCB schematic using SKiDL (Python) for a given circuit described in the text.
This is a **Hard** level task that requires combining multiple functional blocks (power stage, gate driver, isolated power supply).
Please ensure your designed circuit works properly, satisfies the requirements, and follows library conventions.

## Question
Design [TASK].

Input node name: [INPUT].
Output node name: [OUTPUT].
Input voltage: [INPUT_VOLTAGE]V.
Output voltage: [OUTPUT_VOLTAGE]V.

## Available Functional Blocks (Choose what you need)

You may use components from the following categories. **You decide which ones to use based on the design requirements.**

### Half-Bridge Power Stage Options
| Part | Package | Kelvin Source | Max VBUS | Notes |
|------|---------|---------------|----------|-------|
| IMW65R015M2H | TO-247-3 | No | 450V | 3-pin, no KS |
| IMZA65R015M2H | TO-247-4 | Yes (pin 3) | 450V | 4-pin with KS, **recommended for high-side** |
| IMT65R033M2H | TOLL | Yes (pin 2) | 450V | SMD, 12-pin |
| IMLT65R015M2H | TOLT | Yes (pin 7) | 450V | Top-cooled SMD, 16-pin |
| BSC052N08NS5 | QFN | No | 60V | Low voltage only |

### Isolated Gate Driver Options
| Part | Type | Features | Primary Pins | Secondary Pins |
|------|------|----------|--------------|----------------|
| UCC5390E | Isolated | Basic isolated driver | 1-4 | 5-8 |
| UCC21710 | Isolated | DESAT, Miller clamp, OC protection | 9-16 (primary) | 1-8 (secondary) |

### Non-Isolated Gate Driver Options
| Part | Type | Notes |
|------|------|-------|
| UCC27511 | Low-side | Single channel, non-isolated |
| UCC27211 | Bootstrap | Half-bridge with integrated bootstrap diode, **max 100V VBUS** |

### Isolated Power Supply
| Part | Output | Notes |
|------|--------|-------|
| MGJ2D121505SC | +15V/-9V | For isolated gate driver secondary side |

### Passive Components
| Part | Type | Footprint |
|------|------|-----------|
| R | Resistor | R_0805 |
| C | Capacitor (MLCC) | C_0805 |
| C_film | Film Capacitor | C_film (for resonant applications) |
| L | Inductor (SMD) | L_0805 |
| Inductor_power | Power Inductor | Inductor_power (pins 1-6 = A, 7-12 = B) |
| transformer_PQ5050 | Transformer | transformer_PQ5050 (Pri_1: 1-3, Pri_2: 4-6, Sec_1: 7-9, Sec_2: 10-12) |

## Two Implementation Approaches (Choose One)

### Approach A: Direct Implementation (Flat)
Write all components directly without subcircuit abstraction. Suitable for simpler combinations.

```python
from skidl import *

# Define ALL nets with UNIQUE names
vin = Net("VIN")
vout = Net("VOUT")
vbus_p = Net("VBUS_P")
pgnd = Net("PGND")
vsw = Net("VSW")

# Gate driver 1 nets (high-side) - use suffixes to avoid conflicts!
vdd_drv1 = Net("VDD_DRV1")
gnd_drv1 = Net("GND_DRV1")
vee_drv1 = Net("VEE_DRV1")

# Gate driver 2 nets (low-side) - different names!
vdd_drv2 = Net("VDD_DRV2")
gnd_drv2 = Net("GND_DRV2")
vee_drv2 = Net("VEE_DRV2")

# Isolated power supply 1 nets
vin_iso1 = Net("VIN_ISO1")
gnd_iso1_pri = Net("GND_ISO1_PRI")

# ... instantiate components and connect ...
```

### Approach B: Subcircuit Encapsulation
Use `@subcircuit` decorator for reusable blocks. Recommended when using multiple identical structures.

```python
from skidl import *

@subcircuit
def half_bridge_with_driver(vbus_p, pgnd, vsw, gate_h, gate_l, ks_h, ks_l, vdd_drv, vee_drv, gnd_drv):
    """Half-bridge power stage with isolated gate drivers"""
    # High-side MOSFET
    q_h = Part("test", "IMZA65R015M2H", footprint="test:IMZA65R015M2H")
    # Low-side MOSFET
    q_l = Part("test", "IMZA65R015M2H", footprint="test:IMZA65R015M2H")

    # Connect MOSFETs
    q_h[1] += vbus_p        # Drain
    q_h[2] += vsw           # Source
    q_h[3] += ks_h          # Kelvin Source
    q_h[4] += gate_h        # Gate

    q_l[1] += vsw           # Drain
    q_l[2] += pgnd          # Source
    q_l[3] += ks_l          # Kelvin Source
    q_l[4] += gate_l        # Gate

    # Gate resistors (turn-on and turn-off)
    r_gon_h = Part("test", "R", value="10R", footprint="test:R_0805")
    r_goff_h = Part("test", "R", value="2.2R", footprint="test:R_0805")
    # ... more components ...

# Instantiate multiple half-bridges
hb1_vsw = Net("HB1_VSW")
hb2_vsw = Net("HB2_VSW")
# Each subcircuit call creates isolated internal nets
half_bridge_with_driver(vbus_p, pgnd, hb1_vsw, ...)
half_bridge_with_driver(vbus_p, pgnd, hb2_vsw, ...)
```

## [!] CRITICAL Design Rules

### 1. Net Naming (MOST IMPORTANT!)
**Multiple instances MUST use unique net names to avoid unintended shorts!**

[WRONG] **WRONG** - Both half-bridges share same net names:
```python
vsw = Net("VSW")  # Used by both HB1 and HB2 - WRONG!
```

[OK] **CORRECT** - Each instance has unique names:
```python
vsw_1 = Net("VSW_1")  # Half-bridge 1
vsw_2 = Net("VSW_2")  # Half-bridge 2
```

**Isolation domains MUST have separate GND nets!**

[WRONG] **WRONG** - Primary and secondary share GND:
```python
gnd = Net("GND")  # Used for both primary and secondary - WRONG!
```

[OK] **CORRECT** - Separate GND for each domain:
```python
gnd_pri = Net("GND_PRI")           # Primary side (controller)
gnd_sec_h = Net("GND_SEC_H")       # High-side driver secondary
gnd_sec_l = Net("GND_SEC_L")       # Low-side driver secondary
```

### 2. Gate Driver Connections
- **Gate Resistor Required**: Gate driver output  ->  Resistor  ->  MOSFET gate
- **Kelvin Source**: Gate driver return (GND2/VS) MUST connect to **Kelvin Source (KS)** pin, NOT power Source
- **Turn-on/Turn-off Separation**: Use separate resistors with anti-parallel diodes for speed control

```python
# Gate driver to MOSFET connection (correct)
drv[6] += r_gate[1]      # Driver OUT  ->  Gate resistor
r_gate[2] += q[4]        # Gate resistor  ->  MOSFET Gate
drv[7] += q[3]           # Driver GND2  ->  MOSFET Kelvin Source (NOT pin 2!)
```

### 3. Isolated Power Supply
- Primary side connects to main power domain
- Secondary side connects to gate driver power
- **Polarity**: +VOUT  ->  VDD (VCC2), 0V  ->  GND2, -VOUT  ->  VEE

```python
# Isolated DC-DC connection
iso_dcdc[1] += vin_12v        # Primary +VIN
iso_dcdc[2] += gnd_pri        # Primary -VIN (GND)
iso_dcdc[7] += vdd_drv        # Secondary +VOUT (+15V)
iso_dcdc[6] += gnd_sec        # Secondary 0V (GND2)
iso_dcdc[5] += vee_drv        # Secondary -VOUT (-9V)
```

### 4. Decoupling Capacitors
- **Every IC** needs 100nF decoupling between VDD and GND
- **VBUS high dv/dt loop** needs at least 8 decoupling capacitors
- Place caps close to power pins

```python
# Decoupling for isolated gate driver
c_vdd = Part("test", "C", value="100nF", footprint="test:C_0805")
c_vdd[1] += vdd_drv
c_vdd[2] += gnd_drv

# VBUS decoupling (multiple caps for high dv/dt)
for i in range(8):
    c = Part("test", "C", value="100nF", footprint="test:C_0805")
    c[1] += vbus_p
    c[2] += pgnd
```

**Half-Bridge Power Stage Rule**: For tasks with power stage (P17-P23), each half-bridge MUST have at least 8 decoupling capacitors on VBUS. This is verified by the topology checker.

### 5. MOSFET Pin Connections
**ALL parallel pins must be explicitly connected!**

```python
# TO-247-4 (IMZA65R015M2H) - 4 pins, straightforward
q[1] += drain_net    # Drain
q[2] += source_net   # Source
q[3] += ks_net       # Kelvin Source
q[4] += gate_net     # Gate

# TOLL (IMT65R033M2H) - 12 pins, connect all parallel pins
q[1] += gate_net              # Gate
q[2] += ks_net                # Kelvin Source
q[3,4,5,6,7,8,9] += source_net  # All Source pins
q[10,11,12] += drain_net      # All Drain pins
```

## Reference Subcircuits (from Library)

The following verified subcircuits are relevant to your task.
**IMPORTANT**: These are REFERENCE implementations - you must:
1. Adapt net names for multiple instances (e.g., `VSW_1`, `VSW_2` not just `VSW`)
2. Create separate isolation domains for each gate driver
3. Connect Kelvin source pins correctly

[SUBCIRCUIT_REFERENCE]

## Available Components (Detailed)
[COMPONENT_INFO]

## Output Format
Your answer must contain:
1. A concise design plan explaining:
   - Which functional blocks you chose and why
   - How they interconnect
   - Net naming strategy for multiple instances
2. Complete runnable SKiDL Python code

**Hard Constraints:**
- Code must end with `ERC()`
- Do NOT include `generate_netlist()` or `generate_pcb()`
- Do NOT write anything after the code block
- Use pin numbers, not pin names (more robust)
- Use "test" library for all parts: `Part("test", "PartName", ...)`
- Use footprint format: `footprint="test:FootprintName"`

## Answer
\end{lstlisting}

\subsection{Baselines: Circuitron and Same-Oracle Controls}
\label{app:circuitron}

We compare against four baseline classes under the same $T{=}4$ budget and sandbox. \textbf{B1 Circuitron}~\citep{circuitron2025}: a two-stage planner-then-codegen pipeline (plan at $\tau{=}0.3$, codegen at $\tau{=}0.5$), reported as a direct row. \textbf{B2 Circuitron + datasheet RAG (no schema)}: re-introduces retrieval but strips the 32-role KG, supplying raw datasheet spans through BM25+dense hybrid retrieval. \textbf{B3 Self-Refine}~\citep{madaan2023selfrefine}: single-arm refinement with binary pass/fail signal from the same oracle. \textbf{B4 Reflexion}~\citep{shinn2023reflexion}: adds verbal reflection over the layered error log. B2/B3/B4 are reported as axis-aligned proxies via \Cref{tab:component_ablation} because each isolates a single ablation axis (B2: C1 KG; B3/B4: D1 feedback granularity); separate direct rows would duplicate information already in the component ablation.

\begin{table}[t]
\centering
\small
\caption{Baseline controls on Gemma~4-31B. B1 and Ours direct; B2--B4 axis-aligned proxies from \Cref{tab:component_ablation}.}
\label{tab:baselines_circuitron}
\setlength{\tabcolsep}{4pt}
\renewcommand{\arraystretch}{1.1}
\begin{tabular}{p{5.6cm} c c}
\toprule
\textbf{Method (Gemma 4-31B)} & \makecell{\textbf{\pcbbench{}}\\Overall (Hard)} & \makecell{\textbf{OSE}\\Overall (Hard)} \\
\midrule
B1 Circuitron~\citep{circuitron2025}\\(direct)                                  & $41.9$ ($10.2$) & $20.7$ ($3.2$) \\
B2 Circuitron $+$ datasheet RAG\\(proxy: C1 component-only KG)                  & $77.6$ ($36.1$) & n/a \\
B3 Self-Refine on our oracle\\(proxy: D1 no-feedback)~\citep{madaan2023selfrefine}    & $57.7$ ($11.8$) & n/a \\
B4 Reflexion on our oracle\\(proxy: D1 weak)~\citep{shinn2023reflexion}         & $60.4$ ($13.3$) & n/a \\
\midrule
\textbf{Ours (direct)}                                                          & \textbf{$81.3$ ($47.1$)} & \textbf{$69.8$ ($63.4$)} \\
\bottomrule
\end{tabular}
\end{table}

The B1$\to$Ours gap reproduces the main-paper Circuitron comparison; B2 collapses to $-3.7$pp on the C1 axis, B3/B4 to $-23.6$/$-20.9$pp on the D1 axis. The framework's continuous-reward + pin-level feedback + arm-acquiring Thompson Sampling extends all four baselines along strictly orthogonal axes.

\subsection{Adaptive Temperature: Trigger Table and Ablation}
\label{app:adaptive_temp}

The temperature multiplier $\gamma(\mathrm{type}, r)$ in \Cref{eq:temperature} is a closed-form lookup over five branches (\Cref{tab:adaptive_temp_triggers}): exploit ($\gamma{<}1$) when the parent is close to passing, explore ($\gamma{>}1$) when far. The schedule is calibrated once on \pcbbench{} and held fixed across all reported runs and both suites.

\begin{table}[!ht]
\centering
\small
\caption{Adaptive temperature multipliers $\gamma(\mathrm{type}, r)$; final $\tau_t = \mathrm{clamp}(\tau_{\text{base}} \cdot \gamma, 0.1, 1.0)$.}
\label{tab:adaptive_temp_triggers}
\setlength{\tabcolsep}{4pt}
\renewcommand{\arraystretch}{1.05}
\begin{tabular}{l c c l}
\toprule
\textbf{Failing layer} & \textbf{$r$} & \textbf{$\gamma$} & \textbf{Rationale} \\
\midrule
Syntax error                  & any         & 0.6  & surgical fix \\
ERC (L1)                      & any         & 1.0  & moderate fix \\
Topology (L3)                 & $\geq 0.5$  & 0.8  & near-correct, exploit \\
Topology (L3)                 & $< 0.5$     & 1.6  & far, explore \\
Code-extraction               & any         & 1.4  & rethink, explore \\
\bottomrule
\end{tabular}
\end{table}

\paragraph{Factorial ablation (TS $\times$ $\gamma$).}
The $2{\times}2$ factorial on Gemma~4-31B / full 62-task \pcbbench{} (\Cref{tab:adaptive_temp_factorial}) shows sub-additive single-effect contributions: removing TS costs $-2.9$pp Overall and $-2.0$pp Hard; removing $\gamma$ costs $-3.1$pp Overall and $-4.1$pp Hard; removing both jointly costs $-5.6$pp Overall and $-5.7$pp Hard, below the linear sum of the single-effect costs. Adaptive $\gamma$ contributes proportionally more on Hard because the L3 topology branch (which fires the explore-vs-exploit split) dominates Hard failures.

\begin{table}[t]
\centering
\small
\caption{Factorial ablation: Thompson Sampling on/off $\times$ adaptive temperature on/off. Gemma 4-31B, full 62-task \pcbbench{}, 15 trials per task per cell. Each cell is pass rate (\%) reported as Overall / Hard tier; $\Delta$ vs.\ ours on Overall.}
\label{tab:adaptive_temp_factorial}
\setlength{\tabcolsep}{6pt}
\renewcommand{\arraystretch}{1.05}
\begin{tabular}{l c c}
\toprule
                       & \textbf{Adaptive $\gamma$ on} & \textbf{Adaptive $\gamma$ off} \\
\midrule
TS-fractional (ours)   & \textbf{$81.3$ / $47.1$ (ref.)} & $78.2$ / $43.0$ ($-3.1$) \\
Greedy                 & $78.4$ / $45.1$ ($-2.9$)        & $75.7$ / $41.4$ ($-5.6$) \\
\bottomrule
\end{tabular}
\end{table}

\subsection{Five-Strategy Refinement Ablation}
\label{app:5strategy_ablation}

Five arm-selection rules under identical framework state (schema-induced KG, 5-layer oracle, pin-level feedback, adaptive $\gamma$, $T{=}4$ budget); only the bandit policy varies. Gemma~4-31B / 62-task \pcbbench{} / 15 trials per task / $930$ trials per row.

\begin{table}[!ht]
\centering
\small
\caption{Five-strategy refinement ablation. Bold row = ours, best on every tier and Overall.}
\label{tab:5strategy_full}
\setlength{\tabcolsep}{4pt}
\renewcommand{\arraystretch}{1.05}
\begin{tabular}{l c c c c c}
\toprule
\textbf{Strategy} & \textbf{Easy} & \textbf{Medium} & \textbf{Hard} & \textbf{Overall} & \textbf{$\Delta$} \\
\midrule
S0 No refinement (Pass@4 i.i.d.)~\citep{sutton2018rl}             & $91.0$ & $74.8$ & $6.7$  & $60.54$ & $-20.75$ \\
S1 $\epsilon$-greedy ($\epsilon{=}0.2$)~\citep{sutton2018rl}      & $94.9$ & $84.0$ & $41.2$ & $75.27$ & $-6.02$ \\
S2 TS, binary (REx-style)~\citep{thompson1933,tang2024rex}        & $97.6$ & $84.8$ & $40.4$ & $76.13$ & $-5.16$ \\
S3 UCB1~\citep{auer2002ucb}                                       & $98.4$ & $87.4$ & $40.4$ & $77.50$ & $-3.79$ \\
S4 Greedy (best-mean arm)~\citep{sutton2018rl}                    & $96.9$ & $87.4$ & $45.1$ & $78.40$ & $-2.89$ \\
\textbf{Ours: TS-fractional Bernoulli}~\citep{agrawal2013further} & \textbf{98.0} & \textbf{91.9} & \textbf{47.1} & \textbf{81.29} & --- \\
\bottomrule
\end{tabular}
\end{table}

\paragraph{Why ours wins on every tier.}
(i) \emph{Continuous reward over binary.} S2 (TS-binary, REx-style) discards the partial-credit information of the 5-layer oracle, throwing away the L1$\to$L1b$\to$L2 gradient that TS-fractional propagates: $-2.9$pp Easy, $-7.1$pp Medium, $-6.7$pp Hard, $-5.16$pp Overall. (ii) \emph{Bayesian exploration over deterministic.} S4 (Greedy) commits to the highest-mean arm and starves alternative arms of refinement budget; on Hard tasks where the best arm at $T{=}1$ is often a structurally flawed local minimum, this costs $-2.0$pp Hard against ours. (iii) \emph{Reward-informed prior over uniform.} S1 ($\epsilon$-greedy) and S3 (UCB1) start every new arm at uniform prior; ours injects parent reward $r_a$ as a Beta head-start (\Cref{eq:prior}), shrinking the exploration regret on Hard tasks where parent-child reward correlation is positive ($-5.9$pp Hard for S1, $-6.7$pp Hard for S3 vs.\ ours).

\paragraph{Refinement loop dominates arm-selection.}
S0$\to$ours $=+20.75$pp; S4$\to$ours $=+2.89$pp. The verifier-driven refinement loop is the first-order contribution; the specific bandit policy is second-order (gap roughly $7\times$ smaller). Headline numbers are robust to perturbations of the policy once the loop is in place.

\subsection{Component- and Context-Ablation Detail}
\label{app:per_task_heatmap}

Beyond strategy ablation, two further axes from \Cref{sec:experiments} matter at the per-tier level: which framework component contributes (\Cref{tab:component_ablation}) and how context is assembled (\Cref{tab:context_icl}).

\paragraph{Component contributions (B1 reward, C1 KG, D1 feedback).}
On Gemma~4-31B, three independent axes:
\emph{B1 reward signal.} Binary $\{0,1\}$ vs.\ continuous $[0,1]$: $-5.3$pp Overall ($72.7\%\to 78.0\%$); $-9.6$pp Hard ($39.0\%\to 48.6\%$). Continuous reward propagates the partial-credit signal that the 5-layer oracle produces.
\emph{C1 KG.} No KG vs.\ component-only KG vs.\ schema KG: $-54.5$pp Overall ($26.8\%\to 81.3\%$); $-45.1$pp Hard ($2.0\%\to 47.1\%$). Component-only KG (entries without the 32-role ontology) leaves $-3.7$pp Overall and $-11.0$pp Hard. The schema-induced KG is therefore the single largest contribution, especially on Hard system-integration tasks.
\emph{D1 feedback granularity.} None (binary) vs.\ weak (category-level) vs.\ pin-level: $-23.6$pp Overall, $-35.3$pp Hard for binary; $-20.9$pp Overall, $-33.8$pp Hard for category-level. Pin-level localization is what makes the feedback channel actionable on multi-pin Hard tasks; binary and category-level feedback differ from each other by only $2.7$pp Overall, isolating localization (not verbosity) as the load-bearing feature.

\paragraph{Context assembly and ICL (\uca{} vs.\ alternatives).}
On the 20-task subset (300 trials, Gemma~4-31B, \Cref{tab:context_icl}): \uca{} reaches $78.0\%$ Overall ($48.6\%$ Hard) at $\sim$$18$k tokens; full-library 35k-token context drops to $73.7\%$ ($43.8\%$ Hard, $-4.3$pp / $-4.8$pp at twice the prompt cost); two-stage retrieval at 22k tokens reaches $75.0\%$ ($41.9\%$ Hard, $-3.0$pp / $-6.7$pp). \uca{} therefore strictly dominates both alternatives on Overall and Hard at the lowest token budget. Removing in-context exemplars while keeping \uca{} retrieval drops Overall by $-6.0$pp ($72.0\%$), concentrated on Medium ($-10.8$pp) and Hard ($-4.8$pp), with no Easy effect: KG carries structural grounding, ICL exemplars carry Medium-tier disambiguation.

\subsection{Verbatim Failure-Mode Traces}
\label{app:failure_traces}

Five verbatim failure traces, one per verification layer; each lists the failing \skidl{} excerpt, the oracle output with pin-localized error, and the refined excerpt. Reward $\mathcal{V}$ uses \Cref{eq:partial} with constants from \Cref{tab:layer_constants}.

\paragraph{Trace L1 (Electrical Invariants).}
TPS54302 buck regulator; \texttt{EN} pin floating.
\begin{lstlisting}[style=promptstyle]
[fail] buck['EN'] += Net('EN_RAIL')   # never tied to a voltage source
[oracle] L1 FAIL: power-unreachable TPS54302.EN; V=0.39
[fix]    R_EN[1]+=VIN; R_EN[2]+=EN_NET; buck['EN']+=EN_NET
\end{lstlisting}

\paragraph{Trace L3 (Topology Signatures).}
Synchronous-buck inductor on the wrong rail.
\begin{lstlisting}[style=promptstyle]
[fail] L1[1]+=VBUS; L1[2]+=VOUT
[oracle] L3 FAIL: synchronous_buck signature not detected; SW expected on inductor terminal; V=0.94
[fix]    L1[1]+=SW; L1[2]+=VOUT
\end{lstlisting}

\paragraph{Trace L1b (Role-Preserving Net Matching).}
Gate-driver high/low outputs swapped.
\begin{lstlisting}[style=promptstyle]
[fail] driver['HO']+=Q_LS['G']; driver['LO']+=Q_HS['G']
[oracle] L1b FAIL: role_pair_incompatible gate_ho with low-side mosfet_gate; V=0.598
[fix]    driver['HO']+=Q_HS['G']; driver['LO']+=Q_LS['G']
\end{lstlisting}

\paragraph{Trace L2 (Subcategory Templates).}
INA240A1 missing input-side filter capacitor.
\begin{lstlisting}[style=promptstyle]
[fail] ina['IN+']+=SHUNT_HIGH; ina['IN-']+=SHUNT_LOW   # no filter cap
[oracle] L2 FAIL: ina240_input_filter requires 1nF diff cap; V=0.65
[fix]    C_filt[1]+=SHUNT_HIGH; C_filt[2]+=SHUNT_LOW
\end{lstlisting}

\paragraph{Trace L4 (Power Invariants).}
SiC half-bridge low-side Kelvin tied to source.
\begin{lstlisting}[style=promptstyle]
[fail] Q_LS['S']+=PGND; Q_LS['KS']+=PGND   # Kelvin = Source
[oracle] L4 FAIL: kelvin_independence violated on Q_LS; V=0.475
[fix]    KS_LS=Net('KS_LS'); Q_LS['KS']+=KS_LS; INA240A1['IN-']+=KS_LS
\end{lstlisting}

\subsection{End-to-End PCB Manufacturing Pipeline}
\label{app:e2e_pipeline}

We illustrate the downstream pipeline on a 5000~W AC-DC converter, the case study referenced in \Cref{subsec:e2e_pipeline}. The design integrates five \pcbbench{} domains (Sensing, MCU, PowerStage, Comm, AuxPower). \pcbschemagen{}'s contribution ends at Step~1 (verifier-accepted \skidl{} netlist exported to a KiCad schematic); Steps~2--4 use the standard KiCad PCB workflow with no framework involvement, and we report them only to confirm that the framework's output is consumable by a production toolchain.

\begin{figure}[H]
\centering
\includegraphics[width=0.85\columnwidth,keepaspectratio]{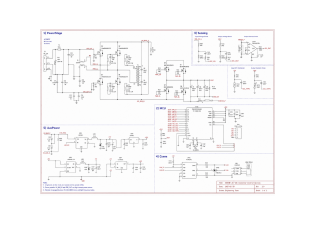}
\caption{Step 1: KiCad schematic exported from the verifier-accepted \skidl{} netlist (5000~W AC-DC converter case study). This is \pcbschemagen{}'s output.}
\label{fig:e2e_schematic}
\end{figure}

\paragraph{Step 1: Schematic export.} The verifier-accepted \skidl{} program is rendered to a KiCad-compatible schematic with all 47-component footprints, net labels, and hierarchical sheets preserved (\Cref{fig:e2e_schematic}). Net connectivity is identical to the in-memory KG used by the verifier; no manual rewiring is required.

\begin{figure}[H]
\centering
\includegraphics[width=0.85\columnwidth,keepaspectratio]{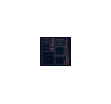}
\caption{Step 2: PCB project after schematic-to-layout import. Footprints stacked at origin, rat's nest unrouted; standard KiCad PCB workflow, outside the framework's contribution.}
\label{fig:e2e_layout_ready}
\end{figure}

\paragraph{Step 2: Layout-ready import.} The schematic is imported into a KiCad PCB project. All component footprints are placed at the origin awaiting placement, and the rat's nest of unrouted nets reflects the netlist exactly (\Cref{fig:e2e_layout_ready}). This step is purely a file-format handoff; nothing in the framework's output needs editing.

\begin{figure}[H]
\centering
\includegraphics[width=0.85\columnwidth,keepaspectratio]{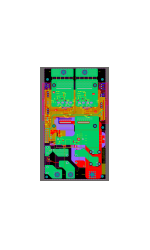}
\caption{Step 3: Layout finished. Placement and routing complete on a 4-layer board, DRC passing; standard KiCad workflow.}
\label{fig:e2e_layout_finished}
\end{figure}

\paragraph{Step 3: Layout finished.} A PCB engineer performs standard placement and routing (component arrangement by domain, power/ground plane definition, differential pair routing, design-rule check). \Cref{fig:e2e_layout_finished} shows the final 4-layer board with all nets routed and the layout passing KiCad's DRC. Steps~2--3 took roughly two engineer-hours for this design.

\begin{figure}[H]
\centering
\includegraphics[width=0.85\columnwidth,keepaspectratio]{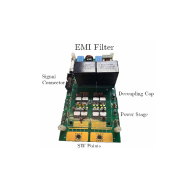}
\caption{Step 4: Fabricated and populated prototype board, produced by a standard PCB fab house.}
\label{fig:e2e_prototype}
\end{figure}

\paragraph{Step 4: Fabricated prototype.} The Gerber files exported from Step~3 were sent to a standard PCB fab house. \Cref{fig:e2e_prototype} shows the resulting populated board after reflow soldering. The board powered up on the first attempt and met its specification.

\subsection{Failure Reason Analysis}
\label{app:failure_reasons}

We analyze the failing-layer distribution across all logged failed trials on \pcbbench{} and \textsc{Open-Schematics-Eval}, aggregated per difficulty tier. Failures are categorized into six classes corresponding to the verification pipeline: \textbf{Stage 0} (code extraction, including syntax errors and runtime failures inside the \skidl{} sandbox), \textbf{L1} (electrical invariants, including ERC subprocess failures), \textbf{L1b} (pin-role compatibility), \textbf{L2} (subcategory connection templates), \textbf{L3} (topology signature mismatch), and \textbf{L4} (power-electronics invariants).

\begin{table}[H]
\centering
\small
\caption{Failure-cause distribution by difficulty tier (\% of failed trials), aggregated across all framework models on \pcbbench{} (6 open-weight models, 2{,}567 failed trials) and \textsc{Open-Schematics-Eval} (8 models, 12{,}291 failed trials).}
\label{tab:fail_distribution}
\setlength{\tabcolsep}{4pt}
\renewcommand{\arraystretch}{1.05}
\begin{tabular}{l c c c c c c c}
\toprule
\textbf{Suite / Tier} & \textbf{Stage 0} & \textbf{L1} & \textbf{L1b} & \textbf{L2} & \textbf{L3} & \textbf{L4} & \textbf{Failures} \\
\midrule
\multicolumn{8}{l}{\textit{\pcbbench{}}} \\
Easy   & 16.9 & 46.9 & 17.2 & 19.0 &  0.0 &  0.0 &   337 \\
Medium &  9.8 & 66.9 & 12.2 &  5.8 &  1.1 &  4.1 &   959 \\
Hard   &  8.3 & 67.7 & 12.8 &  2.3 &  1.8 &  7.0 & 1{,}271 \\
\midrule
\multicolumn{8}{l}{\textit{Open-Schematics-Eval}} \\
Easy   & 17.8 & 66.1 &  8.2 &  6.9 &  0.0 &  0.0 & 2{,}089 \\
Medium & 16.0 & 72.9 &  6.6 &  3.2 &  0.0 &  0.2 & 3{,}544 \\
Hard   & 18.0 & 72.9 &  5.5 &  2.9 &  0.0 &  0.0 & 6{,}658 \\
\bottomrule
\end{tabular}
\end{table}

\begin{figure}[H]
\centering
\includegraphics[width=\textwidth]{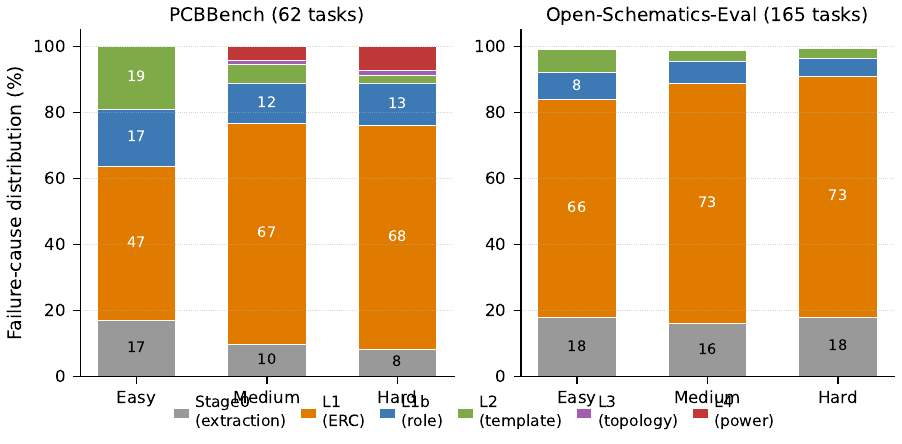}
\caption{Failure-cause distribution by tier, stacked. PCBBench shows the L4 (power-invariant) tail emerging only on Medium / Hard, where Kelvin / decoupling / isolation invariants matter; Open-Schematics-Eval is dominated by L1 (ERC) at every tier because the larger 826-component pool exposes more pin-table mismatches before deeper layers can fire.}
\label{fig:fail_reasons}
\end{figure}

\paragraph{Three trends.}
(i) \emph{L1 ERC dominates failures everywhere.} Once a candidate clears the \skidl{} sandbox (Stage 0), the highest-priority structural check it next confronts is L1; weak open-weight models accumulate L1 failures fastest. (ii) \emph{L4 power invariants are exclusive to Medium / Hard \pcbbench{}.} Easy tasks contain no power-domain Kelvin or isolation predicates, so L4 has nothing to fire on; the Medium / Hard L4 tail (4.1 / 7.0\%) directly measures the framework's added value on power-electronics tasks. (iii) \emph{OSE failures concentrate at L1 across all tiers (66--73\%).} The substantially larger commercial-IC pool (439 unique ICs vs.\ 41 in \pcbbench{}) exposes more pin-table mismatches in early generation attempts; tasks that survive L1 generally pass on \textsc{Open-Schematics-Eval}, so deeper-layer failure rates on OSE are correspondingly small.

\subsection{Limitations, Reward-Hacking, and Broader Impact}
\label{app:reward_hacking}\label{app:limitations}\label{app:broader_impact}

\paragraph{Scope of correctness.}
The 5-layer oracle certifies \emph{structural} correctness (ERC/DRC + pin-role + constraint-predicate invariants) under real IC pin and package constraints. It does not measure SPICE functional behavior, EMI/EMC, or thermal performance; LVS-style device-matching and SPICE-driven functional simulation are natural follow-on oracles.

\paragraph{Capacity floor and ontology breadth.}
Below $\sim$8B active parameters refinement gain shrinks to $+10$pp (InternLM3-8B); the framework composes multiplicatively with model capability rather than substituting for it. The 32-role pin ontology and the four predicate templates are stabilized for board-level digital/analog/power; RF/high-speed-digital signal-integrity constraints and on-die mixed-signal SoCs lie outside this scope and require new pin roles, predicates, and extraction prompts.

\paragraph{Library scaling.}
Scaling beyond the 47-component library to $\sim$1{,}000+ ICs is feasible under the same Stage 1 / Stage 2 retrieval architecture (\Cref{subsec:kg}) but requires hierarchical catalog retrieval (the linear-in-$|\mathcal{C}|$ catalog crosses 35k tokens at $\sim$1{,}000 ICs) and tighter Stage-2 selection across near-equivalent parts.

\paragraph{Reward-hacking surface.}
Three structural properties collapse the surface: (i) the verifier is fully deterministic with publicly-released source code, so any predicate-gaming attempt is auditable during downstream review; (ii) maximum reward requires \emph{every} layer to pass, so gaming a single layer (e.g., trivially passing L1 by collapsing nets) is bounded by lower-priority layers (the verifier-layer ablation in \Cref{sec:experiments} measures the $10$--$13$pp false-accept rate that an L1-only oracle would admit); (iii) the schema-induced KG and the verifier are independently derived from IC-vendor datasheets, not from any task in \pcbbench{}/OSE, so the generator has no per-task target to over-fit. The Cohen's $\kappa{=}0.91$ expert agreement (\Cref{app:human_eval}) bounds the residual false-accept rate at $6.7\%$, concentrated on Layer-3 borderline-but-electrically-valid topologies. Planned defenses: randomized predicate ordering, per-task held-out templates for adversarial spot-check, downstream LVS/SPICE oracles structurally orthogonal to the structural verifier.

\paragraph{Broader impact.}
Faster, more reliable PCB schematic synthesis lowers the engineering-hour cost of board-level design across consumer/medical/EV/renewable/AI-hardware domains and broadens access to professional-grade PCB tooling. Three risk classes warrant explicit acknowledgement: (i) dual use (accelerated schematic generation can speed harmful as well as beneficial products), (ii) reward hacking (a gating verifier is an attack surface), and (iii) safety-critical deployment (structural verification is not a substitute for FCC/CE/UL/IEC~60601 certification). Mitigations: audit-traceable predicates, mandatory downstream certification gates, human-in-the-loop review at sign-off, and gated release for safety-critical IC extensions. Inputs are public technical datasheets and open-source schematics; outputs are KiCad-compatible netlists. We see no specific privacy concerns and no specific fairness concerns within the technical scope, though access to professional EDA tooling remains a broader community question.
 
\end{document}